%% file: ms.tex
\documentclass[twoside,11pt]{article}
\pdfoutput=1
\usepackage{jmlr2e}
\usepackage{graphicx}
\usepackage{subfigure}
\usepackage{bm}
\usepackage{amsmath}
\usepackage{rotating}
\usepackage{nicefrac}
\usepackage{url}

\newcommand{\ka}{\kappa}
\jmlrheading{}{}{}{}{}{}

\ShortHeadings{}{}
\firstpageno{1}

\begin{document}

\title{Sparse Multi-Output Gaussian Processes for \\
Medical Time Series Prediction}
\author{\name Li-Fang Cheng \email lifangc@princeton.edu \\
       \addr Department of Electrical Engineering\\
       Princeton University\\
       Princeton, NJ 08544, USA
       \AND
       \name Gregory Darnell \email gdarnell@princeton.edu \\
       \addr Lewis-Sigler Institute\\
       Princeton University\\
       Princeton, NJ 08544, USA
       \AND
       \name Bianca Dumitrascu \email biancad@princeton.edu \\
       \addr Lewis-Sigler Institute\\
       Princeton University\\
       Princeton, NJ 08544, USA
       \AND
       \name Corey Chivers \email corey.chivers@uphs.upenn.edu\\
	   \addr University of Pennsylvania Health System\\
	   Philadelphia, PA 19104, USA
       \AND
       \name Michael E Draugelis \email michael.draugelis@uphs.upenn.edu\\
	   \addr University of Pennsylvania Health System\\
	   Philadelphia, PA 19104, USA
       \AND
       \name Kai Li \email li@cs.princeton.edu \\
       \addr Department of Computer Science\\
       Princeton University\\
       Princeton, NJ 08540, USA
       \AND
       \name Barbara E Engelhardt \email bee@princeton.edu \\
       \addr Department of Computer Science\\
       Center for Statistics and Machine Learning\\
       Princeton University\\
       Princeton, NJ 08540, USA}
\editor{}
\maketitle

\begin{abstract}
In the scenario of real-time monitoring of hospital patients, high-quality inference of patients' health status using all information available from clinical covariates and lab tests is essential to enable successful medical interventions and improve patient outcomes. Developing a computational framework that can learn from observational large-scale electronic health records (EHRs) and make accurate real-time predictions is a critical step. In this work, we develop and explore a Bayesian nonparametric model based on Gaussian process (GP) regression for hospital patient monitoring. We propose MedGP, a statistical framework that incorporates 24 clinical and lab covariates and supports a rich reference data set from which relationships between observed covariates may be inferred and exploited for high-quality inference of patient state over time. To do this, we develop a highly structured sparse GP kernel to enable tractable computation over tens of thousands of time points while estimating correlations among clinical covariates, patients, and periodicity in patient observations. MedGP has a number of benefits over current methods, including (i) not requiring an alignment of the time series data, (ii) quantifying confidence regions in the predictions, (iii) exploiting a vast and rich database of patients, and (iv) inferring interpretable relationships among clinical covariates. We evaluate and compare results from MedGP on the task of online prediction for three patient subgroups from two medical data sets across 8,043 patients. We found MedGP improves online prediction over baseline methods for nearly all covariates across different disease subgroups and studies. The publicly available code is at \url{https://github.com/bee-hive/MedGP}.
\end{abstract}


\input{Introduction}
\input{Methods}
\input{Results}
\input{Discussion}

\input{AppendixA}
\input{AppendixB}
\input{AppendixC}
\input{AppendixD}
\input{AppendixE}

\newpage
\bibliography{MedGP}

\end{document}

%% file: Introduction.tex
\section{Introduction}
\label{sec:Introduction}
Large-scale collections of electronic health records (EHRs) are becoming useful for understanding disease progress, early diagnosis, and personalized treatments for many clinical diseases~\citep{BigData-to-Healthcare-JAMA, Hripcsak117-JAMA, Ghassemi2015-CC}. EHRs contain rich patient information---disease history, demographics, vital signs, and lab results---that clinicians use to diagnose and treat patients. In this work, we are interested in developing a statistical framework that leverages medical data from a set of reference patients to enable personalized, real-time monitoring of new hospital patients. In particular, we consider data from the Hospitals at the University of Pennsylvania (HUP) containing hospital information for over $260,000$ patients, and the public Multiparameter Intelligent Monitoring in Intensive Care (MIMIC-III) data set with more than $53,000$ admissions from $38,000$ patients in intensive care units (ICUs)~\citep{johnson2016mimic}.

One motivation for monitoring new patients is to characterize patient state to allow the early diagnosis of sepsis or septic shock. 
Sepsis is one of the leading causes of death in critically ill patients in the United States~\citep{NEJM-Sepsis-Review}. Each year an estimated 750,000 cases of sepsis or septic shock occur in the US. The mortality rate of septic patients ranges from 20\% to 30\%, and accounts for roughly 9.3\% of all US deaths~\citep{angus2001epidemiology, kumar2011nationwide}. Sepsis is usually developed during a patient's stay in the hospital. However, accurate diagnosis of sepsis is difficult due to heterogeneous symptoms across patients~\citep{pierrakos2010sepsis}.

\begin{figure*}
\centering
\includegraphics[width=16.0cm]{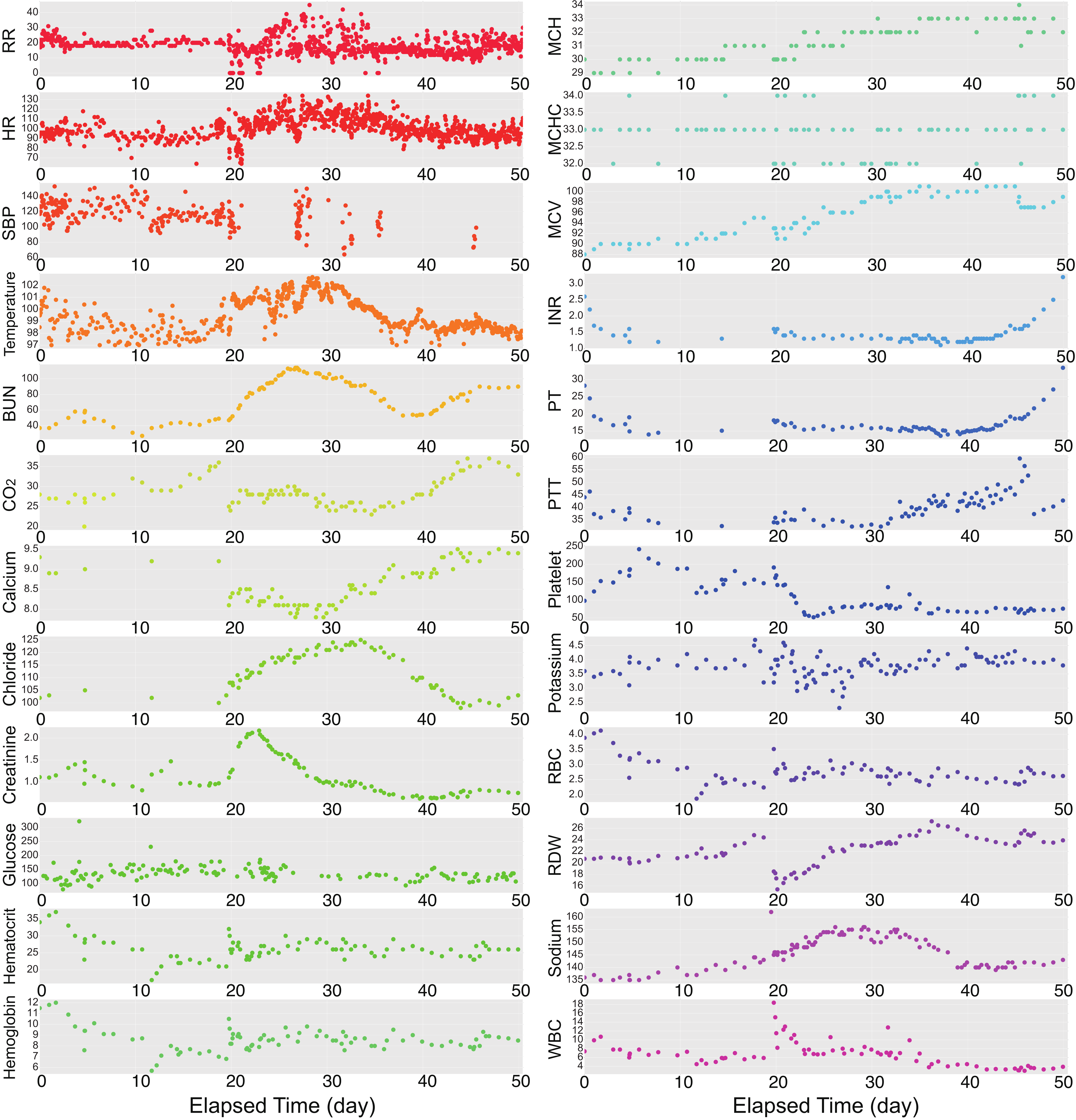}
\caption{\textbf{An example of time series data of 24 clinical covariates for a septic patient in the HUP data.} The 24 covariates include four vital signs---respiration rate (RR), heart rate (HR), systolic blood pressure (SBP), body temperature---and 20 lab results. The time series are aligned by the patient's admission time. The density of sampling varies widely over the 24 covariates. A full description of these covariates can be found in Table~\ref{table:covariate_list_24}.}
\label{fig:time_series_example}
\end{figure*}

One way to reduce the mortality rate of sepsis is to increase the accuracy of early diagnosis of sepsis. To do this, we might develop a model of patient state and fit this model to EHR data from previous hospital patients with sepsis. However, existing EHR data pose several challenges because they have been collected with traditional monitoring methods. Many of the covariates, lab results in particular, are sparsely sampled across patients. That is, there are only a small number of observations of any lab result per patient. For example, vital signs are generally taken once every three to four hours for inpatient data, and once every hour for patients in the intensive care unit (ICU). In contrast, blood tests requiring a blood draw are generally performed at most once a day. We see this sparsity in an example of 24 clinical covariates (Table~\ref{table:covariate_list_24}) measured across time for a single patient, including four densely sampled vital signs (respiration rate, heart rate, systolic blood pressure, and body temperature) and 20 sparsely sampled lab covariates (Figure~\ref{fig:time_series_example}). Data missingness is systematic and not at random~\citep{newgard2015missing}: a doctor will only order a test that will be informative in characterizing patient state relevant to diagnosis. Moreover, these time series data are not aligned across patients to a reference time point or disease onset; instead, patient intake is at time $0$ and release is hours or days later. The sparsity over patients and uncalibrated time series make the physiological progression of disease within patients or joint analysis of time series across patients challenging due to substantial uncertainty of patient state and rate of disease progression at any time.

\begin{table*}
\centering
{\footnotesize 
\begin{tabular}{llcccc}
\hline
Type & Covariate & Sepsis & Neoplasms & Heart Failure & MIMIC-III\\
\hline
Vital 	& Respiration rate (RR) & 87,076 & 493,964 & 147,445 & 291,466\\
Vital 	& Heart rate (HR) & 96,317 & 527,989 & 227,951 & 294,746\\
Vital 	& Systolic blood pressure (SBP) & 84,909 & 447,666 & 104,129 & 124,587\\
Vital 	& Body temperature (Temp) & 80,597 & 364,286 & 94,468 & 56,533\\
Lab 	& Blood urea nitrogen (BUN) & 12,528 & 71,825 & 21,751 & 25,102\\
Lab 	& Carbon dioxide (CO$_{2}$) & 12,672 & 72,784 & 21,844 & 20,979\\
Lab 	& Calcium level & 10,388 & 66,051 & 18,867 & 20,568\\
Lab 	& Chloride & 10,100 & 68,534 & 21,421 & 26,248\\
Lab 	& Creatinine & 12,689 & 72,928 & 21,889 & 25,237\\
Lab 	& Glucose point-of-care (Glucose POC) & 20,444 & 170,872 & 54,239 & 24,196\\
Lab 	& Hematocrit (Hct) & 12,752 & 74,060 & 22,035 & 24,810\\
Lab 	& Hemoglobin (Hgb) & 13,005 & 75,646 & 27,891 & 21,226\\
Lab 	& Mean cell hemoglobin (MCH) & 12,587 & 69,736 & 18,379 & 20,877\\
Lab 	& Mean cell hemoglobin concentration (MCHC) & 12,577 & 69,682 & 18,359 & 20,885\\
Lab 	& Mean cell volume (MCV) & 12,587 & 69,751 & 18,380 & 20,875\\
Lab 	& International normalization ratio (INR) & 5,733 & 38,810 & 17,005 & 15,735\\
Lab 	& Prothrombin time (PT) & 5,722 & 38,844 & 17,007 & 15,734\\
Lab 	& Partial thromboplastin time (PTT) & 5,872 & 41,894 & 19,596 & 17,185\\
Lab 	& Platelet & 12,586 & 69,945 & 18,367 & 21,395\\
Lab 	& Potassium level & 12,830 & 77,395 & 28,470 & 27,200\\
Lab 	& Red blood cell (RBC) & 12,600 & 69,776 & 18,387 & 20,876\\
Lab 	& Red cell distribution width (RDW) & 12,580 & 69,757 & 18,381 & 20,877\\
Lab 	& Sodium level & 12,848 & 78,617 & 28,597 & 26,383\\
Lab 	& White blood cell (WBC) & 12,581 & 69,950 & 18,384 & 20,960\\
\hline
\end{tabular}
}
\caption{{\bf The 24 clinical covariates modeled in MedGP.} This table includes the total number of observations for each covariate across patients in three disease groups---sepsis, neoplasms, and heart failure---in the HUP data, and the heart failure patients in the MIMIC-III data.}
\label{table:covariate_list_24}
\end{table*}

In this work, we build a statistical framework that uses sparse, heterogeneous EHR time series data to monitor and predict vital signs and lab results for each patient in an online way. To do this, we first designed a nonparametric model based on Gaussian process (GP) multivariate regression to explore the correlations both within each clinical covariate across time and across clinical covariates given rich EHR reference data. Our model includes a highly structured GP kernel regularized using sparsity-inducing priors to avoid overfitting, allow interpretability, and ensure computational tractability. Second, we propose a framework based on nonparametric density estimation to tailor the empirical model to a patient-specific model for each new patient. For real-time monitoring, we update the empirical distribution from reference patients with patient-specific observations as measurements are observed. We evaluate our method, MedGP, on over 6,000 patients from three disease groups with more than four million measurements from the HUP data and one disease group from the MIMIC-III data set. We compare results to state-of-the-art approaches for patient online monitoring and investigate differences in correlations among covariates across disease groups.

\section{Related Work}
\label{sec:RelatedWork}

Related work falls into three areas of medical time series analysis: (i) incorporating noisy, heterogeneous, irregular, and sparsely sampled time series data; (ii) combining information across multiple time series; and (iii) exploiting reference data in addition to observations about the current patient to enable patient-specific predictions for a new hospital patient. 

Most prior work has focused on modeling each clinical covariate separately. Due to the irregularity and temporal sparsity of medical data, conventional time series models, such as hidden Markov models (HMMs), autoregressive (AR) models, state-space models, and linear dynamical systems (LDS), are challenging to apply because of the assumption of regular measurement sampling in time. 
Recent work has focused on developing methods to compensate for the missing data in order to work with models that assume complete data. In~\cite{kim2010temporal}, missing data were imputed by averaging over a time window using a kernel support vector machine (SVM). Methods such as matrix factorization and $k$-nearest neighbor (KNN) clustering were used for missing data imputation, and improvements in septic shock prediction were reported~\citep{ho2014septic}. In other work, a hierarchical switching LDS model was used to monitor the physiological signals during neonatal sepsis; the model allows the latent state of a patient to change during periods with fewer observations~\citep{stanculescu2014hierarchical}. In an alternative approach, noisy and sparse time series data were smoothed temporally by putting Gaussian priors on the mean parameters of the Gaussian mixture model, which is related to a Gaussian process prior, although the distribution is over a finite-dimensional vector~\citep{marlin-2012-unsupervised}. 

Gaussian processes (GPs) are useful approaches for time series analysis because they can naturally capture irregular time series observations and estimate prediction uncertainties in a probabilistic framework~\citep{Roberts20110550}. For these reasons, GPs have been applied to the analysis of medical time series data. Previous work used a single-output GP regression model to smooth and impute each covariate independently~\citep{stegle2008gaussian,lasko2013computational}. The Probabilistic Subtyping Model (PSM) added patient-specific information for smoothing temporal trajectories of clinical covariates and clustering disease subtypes~\citep{suchi2015clustering}. PSM learns a mixture model based on a B-spline and GPs to impute the clinical measurements for patients with scleroderma. Demographic covariates, including gender, ethnicity, and clinical history, were also incorporated in the model. In an extension of PSM, the authors adapted patient-specific information to forecast specific clinical covariates~\citep{Suchi-NIPS2015_5873}; the time series for each covariate was still modeled independently.

The idea of capturing the joint dynamics between vital signs and lab tests has also been explored. Using high-frequency regularly sampled time series, the dynamics between heart rate (HR) and blood pressure (BP) were modeled using a mixture of an LDS model~\citep{discover-cardio-dynamics-2012} 
and a switching vector autoregressive model (SVAR)~\citep{dynamics-based-patient-monitoring-2015}. The joint dynamics estimated across covariates were reported to be associated with hospital mortality. In other work~\citep{rizopoulos2011bayesian}, a multivariate spline-based approach with linear mixed effects was used to predict multiple longitudinal outcomes and time-to-death of patients. Time series graphical models (TGMs)~\citep{dahlhaus2000graphical, tank2015bayesian} have also been studied and applied for analyzing multivariate medical time series of ICU patients~\citep{gather2002graphical}. TGMs model the partial correlations between each dimension of the multivariate time series as an undirected graph. However, both TGMs and SVAR models follow the assumptions of vector autoregressive (VAR) models, and thus assume the sampling interval of the time series is fixed across dimensions. In practice, this means missing data imputation needs to be done in advance~\citep{tank2015bayesian}.

Several multi-output GP frameworks have been proposed for other application areas. In the geostatistics literature, the linear model of coregionalization (LMC) characterizes correlations between outputs through a set of kernels and coregionalization matrices that estimate weights for pairwise outputs~\citep{journel1978mining, goovaerts1997geostatistics}. In the machine learning literature, related models include multi-task GPs~\citep{bonilla2007multi}, semiparametric latent factor models~\citep{TehSeeJor2005}, and multi-task kernel learning~\citep{NIPS2011_Spike_and_Slab_VI}. These can be viewed as variations of the LMC with different parameterizations and constraints. Convolution processes (CPs) have also been adapted to model multiple correlated outputs through the convolution of smooth kernels and latent processes~\citep{alvarez2011computationally}. This approach usually has fewer hyperparameters and more efficient computation as compared to LMC, but only squared exponential (SE) kernels have been shown to be computationally tractable. Applying a multi-task GP (MTGP) framework~\citep{bonilla2007multi} to clinical time series analysis has also been considered in two studies~\citep{ghassemi2015multivariate,Durichen2015MTGP}; both studies considered one patient as one task and used the remaining patients as reference training data. Other work adapted the LMC framework with one SE kernel to model three sparsely sampled clinical covariates (intracranial pressure, mean arterial blood pressure, and Pressure-Reactivity Index) jointly~\citep{ghassemi2015multivariate}. The MTGP was shown to outperform a single-task GP (STGP) in prediction error. Both MTGP and CP have also been used with an SE kernel to model three densely sampled vital signs (respiration rate, systolic blood pressure, and heart rate); both methods showed improvements as compared to a single-task GP~\citep{Durichen2015MTGP}.

Our work is distinct from previous research in several ways. First, we use the GP regression framework to model multiple irregularly sampled medical time series using a sparse structured multi-output kernel. In contrast to related work~\citep{ghassemi2015multivariate, Durichen2015MTGP}, our kernel uses a mixture of flexible spectral kernels~\citep{SMkernel-WilsonA13}, allowing periodic behavior and both short-term and long-term dependencies within and across the clinical covariates over time. Second, we use the LMC framework to enable an interpretable quantification of cross-correlation and sparsity between covariates. Third, we model many more clinical covariates (24) compared with previous studies (at most three); in the online medical setting, efficient and scalable computation in this multi-view model is essential. To the best of our knowledge, this is the first work that uses a sparse and low-rank formulation of the shared covariance matrix across clinical covariates to estimate and regularize the relationships between covariates in order to learn about covariate relationships specific to patient subgroups and to prevent overfitting.

In our methodology, MedGP, we trained a GP model on each reference patient separately, and used these models to estimate the empirical population-level model using nonparametric density estimation. This approach avoids training procedures that iterate through all reference patients, which is computationally intractable for an online system~\citep{ghassemi2015multivariate,Durichen2015MTGP}.  
To speed up training, we optimized the implementation in C++ using multithreading. 
Finally, in order to personalize the model for a new patient, we update the empirical population-level model on-the-fly to estimate patient specific parameters as measurements from the new patient are observed.

%% file: Methods.tex
\section{Methods}
\label{sec:Methods}
In this section, we describe our method, MedGP, for estimating the underlying dynamic processes jointly across a large number of sparsely sampled clinical covariates. We first describe the design of the Gaussian process kernel for capturing the temporal correlations within and between covariates. Next, we introduce the sparsity-inducing prior to regularize the LMC weight matrix. We then describe estimation of the parameters in the prior and the kernel. Next, we describe how to learn a patient-specific kernel by first building a population-level model from reference patients and then performing online updating of the parameters when observations about a new patient accumulate. Finally, we describe methods to perform computationally tractable online inference in these models, concluding with a discussion of computational complexity.

\subsection{Gaussian Processes (GPs)}
Gaussian processes (GPs) are distributions over arbitrary functions. By definition, a Gaussian process is a collection of random variables, any finite collection of which have a joint Gaussian distribution. Alternatively, a GP can be described as a distribution on an arbitrary function, defined as
\begin{equation}
f(\mathbf{x}) \sim \mathcal{GP}(m(\mathbf{x}), \ka(\mathbf{x}, \mathbf{x}')),
\end{equation}
where $m(\mathbf{x})$ is the \emph{mean function}:
\begin{equation}
m(\mathbf{x}) = \mathbb{E}[f(\mathbf{x})],
\end{equation}
and $\ka(\mathbf{x}, \mathbf{x}')$ is the \emph{covariance function} or \emph{kernel}:
\begin{equation}
\ka(\mathbf{x}, \mathbf{x}') = \mathbb{E}[(f(\mathbf{x})-m(\mathbf{x}))(f(\mathbf{x}')-m(\mathbf{x}'))].
\end{equation}
Any finite number of function values jointly have a multivariate Gaussian distribution with mean vector $\bm{\mu}$ and covariance matrix $\bm{K}$ between any pair of observations, defined by the kernel function,
\begin{equation}
\begin{array}{c}
[f(x_{1}), f(x_{2}), \cdots, f(x_{T})]^{\top} \sim \mathcal{N}(\bm{\mu}, \bm{K}),\\\\
\bm{\mu} = [m(x_{1}), m(x_{2}), \cdots, m(x_{T})]^{\top},\\\\
\bm{K}_{i,j} = \ka(x_{i}, x_{j}).
\end{array}
\end{equation}

Properties of the function $f(\mathbf{x})$ such as smoothness or periodicity are determined by the kernel function $\kappa(\mathbf{x}, \mathbf{x}')$. One of the most commonly used kernels is the squared exponential (SE) kernel
\begin{equation}
\kappa(\mathbf{x}, \mathbf{x}') = \sigma^{2}\exp{\left( -\frac{||\mathbf{x}-\mathbf{x}'||^{2}}{2\ell^{2}} \right)},
\end{equation}
which is parameterized by a length scale $\ell$ and a scale factor $\sigma$. The functions generated by a GP with an SE kernel are smooth because the kernel function is infinitely differentiable~\citep{Rasmussen2006}. The value of the length scale $\ell$ determines the distribution of changes over the function value with respect to changes in the input $\mathbf{x}$, encouraging a specific smoothness. Due to its simplicity, SE is used in many applications; however, the properties of the functions that it captures are fairly limited. Periodic functions, for example, are not well modeled by an SE kernel, but instead captured by a periodic kernel
\begin{equation}
\kappa(\mathbf{x}, \mathbf{x}') = \sigma^{2}\exp{\left[ -\frac{4 \sin^{2}{ \left(\frac{\pi ||\mathbf{x}-\mathbf{x}'||}{p} \right)}}{\ell^{2}} \right]},
\end{equation}
where $p$ is the period of the function. When modeling medical time series, the SE kernel or the periodic kernel are often used in combination to capture the unknown source-specific smoothness and periodicity of the trajectories of clinical covariates~\citep{stegle2008gaussian,Durichen2015MTGP}.

\subsection{Gaussian Process Regression with a Structured Multi-Output Kernel}
\label{subsec:GP_with_Structured_Multi-Output_Kernel}

Our first goal is to jointly model multiple clinical covariates---vital signs and lab tests---over time for each patient using GP regression. For the $i$th patient, we denote the time series of the $d$th covariate as a vector $\mathbf{x}_{i,d}$, representing the time points that the $d$th covariate was observed, and the corresponding observation vector $\mathbf{y}_{i,d}$:
\begin{equation}
\mathbf{x}_{i,d}^{\top} = \left[ x_{i,d,1}, x_{i,d,2}, \ldots x_{i,d,t} \ldots, x_{i,d,T_{i,d}} \right],
\end{equation}
\begin{equation}
\mathbf{y}_{i,d}^{\top} = \left[ y_{i,d,1}, y_{i,d,2}, \ldots y_{i,d,t} \ldots, y_{i,d,T_{i,d}} \right],
\end{equation}
where $t$ indexes time, and $T_{i,d}$ is the total number of observations for the $d$th covariate of the $i$th patient. 

To represent the time series data over all $D$ covariates, we define
\begin{eqnarray}
\mathbf{x}^{\top}_{i} &=& \left[ \mathbf{x}_{i,1}^{\top}, \mathbf{x}_{i,2}^{\top}, \ldots, \mathbf{x}_{i,D}^{\top} \right],\\
\mathbf{y}^{\top}_{i} &=& \left[ \mathbf{y}_{i,1}^{\top}, \mathbf{y}_{i,2}^{\top}, \ldots, \mathbf{y}_{i,D}^{\top} \right],
\end{eqnarray}
where $\mathbf{x}_{i}, \mathbf{y}_{i} \in \mathbb{R}^{T_{i} \times 1}$, $T_{i} = \left( \sum_{d=1}^{D}T_{i,d} \right)$. Let $\mathcal{F}_{i}$ be a multi-output function over time for the $i$th patient. We capture the relationship between time and clinical observations as a GP regression model:
\begin{equation}
\mathbf{y}_{i} = \mathcal{F}_{i}(\mathbf{x}_{i}) + \bm{\epsilon}_{i},
\end{equation}
where $\bm{\epsilon}_{i}$ is the residual noise vector. Marginally at the $t$th observation of the $d$th covariate, the residual noise is modeled as 
\begin{equation}
\epsilon_{i,d,t} \sim \mathcal{N}(0, \sigma_{i,d}^{2}),
\end{equation}
where $\sigma_{i, d}^{2}$ is the covariate-specific residual variance for each individual.

We assume that the function $\mathcal{F}_{i}$ is drawn from a patient-specific Gaussian process $\mathcal{GP}_{i}$ with mean function $\mu_{i}(\mathbf{x})$ and kernel $\kappa_{i}(\mathbf{x}, \mathbf{x}')$:
\begin{equation}
\mathcal{F}_{i} \sim \mathcal{GP}_{i}(\mu_{i}(\mathbf{x}), \kappa_{i}(\mathbf{x}, \mathbf{x}')).
\end{equation}
We set $\mu_{i}(\mathbf{x}) = \bm{0}$~\citep{Rasmussen2006}. 

We designed the kernel $\kappa_{i}(\mathbf{x}, \mathbf{x}')$ to capture predictive and generalizable covariance structure across medical time series data. Assuming the covariates are correlated across time, we adapted the linear model of coregionalization (LMC) framework~\citep{journel1978mining, goovaerts1997geostatistics}.
We used a set of $Q$ \emph{basis kernels} $\lbrace \kappa_{q}(\mathbf{x}, \mathbf{x}') \rbrace_{q=1}^{Q}$ to model $D$ covariates jointly. The kernel for the cross-covariance of any pair of covariate types is modeled by a weighted, structured linear mixture of the $Q$ basis kernels. The full joint kernel is written as a block structured function
\begin{equation}
\begin{array}{cl}
\kappa_{i}(\mathbf{x}_{i}, \mathbf{x}_{i}') &= 
\displaystyle \sum_{q=1}^{Q}
\left(
\begin{bmatrix}
b_{q, (1,1)}\kappa_{q}(\mathbf{x}_{i, 1}, \mathbf{x}'_{i, 1}) & \cdots & b_{q, (1,D)}\kappa_{q}(\mathbf{x}_{i, 1}, \mathbf{x}'_{i, D})\\
b_{q, (2,1)}\kappa_{q}(\mathbf{x}_{i, 2}, \mathbf{x}'_{i, 1}) & \cdots & \vdots\\
\vdots & \ddots & \vdots \\
b_{q, (D,1)}\kappa_{q}(\mathbf{x}_{i, D}, \mathbf{x}'_{i, 1}) & \cdots & b_{q, (D,D)}\kappa_{q}(\mathbf{x}_{i, D}, \mathbf{x}'_{i, D})\\
\end{bmatrix}
\right),
\end{array}
\label{eq:full_kernel}
\end{equation}
where $b_{q, (d, d')}$ scales the covariance (defined by the $q$th basis kernel) between covariates $d$ and $d'$, and $\kappa_{i}(\mathbf{x}_{i}, \mathbf{x}_{i}) \in \mathbb{R}^{T_{i} \times T_{i}}$. We collapsed $b_{q, (d, d')}$ into a set of weight matrices $\lbrace \bm{B}_{q} \rbrace_{q=1}^{Q}$, where each $\bm{B}_{q}$ is a symmetric positive definite matrix
\begin{equation}
\begin{array}{cl}
\bm{B}_{q} &= 
\begin{bmatrix}
b_{q, (1, 1)} & b_{q, (1, 2)} & \cdots & b_{q, (1, D)}\\
b_{q, (1, 1)} & \vdots & \ddots & \vdots\\
\vdots & \vdots & \ddots & \vdots \\
b_{q, (D, 1)} & b_{q, (D, 2)} & \cdots & b_{q, (D, D)}\\
\end{bmatrix}
\in \mathbb{R}^{D \times D}.
\end{array}
\end{equation}
If the inputs are the same for all covariates, we can further simplify Eq.~(\ref{eq:full_kernel}) with Kronecker product $\otimes$. That is, if $\mathbf{x}_{i,1}=\mathbf{x}_{i,2}=\cdots=\mathbf{x}_{i,D}\triangleq\mathbf{x}_{i,*}$ and $\mathbf{x}_{i,1}'=\mathbf{x}_{i,2}'=\cdots=\mathbf{x}_{i,D}'\triangleq\mathbf{x}_{i,*}'$ :
\begin{eqnarray}
\kappa_{i}(\mathbf{x}_{i}, \mathbf{x}_{i}') &=& \displaystyle \sum_{q=1}^{Q}{\bm{B}_{q} \otimes \kappa_{q}(\mathbf{x}_{i,*}, \mathbf{x}_{i,*}')},
\end{eqnarray}
although in practice we do not often see this situation in medical time series data.

The properties of the time series observations, such as periodicity and short term dependencies, are captured in the $Q$ basis kernels. For medical covariates, the properties of each patent's time series observations may vary. As a trivial example, when a patient is under age 18, their pulse will be well correlated with their age, height, and weight; above age 18, the correlation among pulse, age, height, and weight is more variable within age than across ages. Furthermore, only a few vital signs, such as heart rate, blood pressure, and body temperature, are known to be periodic with a 24-hour period (i.e., a circadian rhythm), but whether there is a similar period for specific lab results, such as white blood cell count or {pressure of carbon dioxide in the blood}, is unclear~\citep{HumanPhysiology}. 

To handle the heterogeneity of patterns within covariates and across patients, we selected the spectral mixture (SM) kernel as the basis kernel~\citep{SMkernel-WilsonA13}. The SM kernel is a general form of a variety of stationary kernels, including the squared exponential (SE) kernel and the periodic kernel, and has also shown good performance in modeling processes generated from more complex kernels through a mixture of kernels approach~\citep{SMkernel-WilsonA13}. The basis kernel $\kappa_{q}(x_{t}, x_{t'})$ is written as
\begin{equation}
\begin{array}{c}
\kappa_{q}(x_{t}, x_{t'}) = \exp{(-2\pi^2\tau^{2} v_{q})}\cos{(2\pi\tau\mu_{q})},\\
\tau = |x_{t} - x_{t'}| \quad \text{(absolute distance in time)}.
\end{array}
\end{equation}
In our work, the mixture weights for each basis kernel are encoded in $\bm{B}_{q}$.

To be used for GP regression, $\kappa_{i}(\mathbf{x}, \mathbf{x}')$ must be a valid Mercer kernel, i.e., the Gram matrix must be positive definite for all $\mathbf{x}$ and $\mathbf{x}'$. Since the matrix produced by each basis kernel $\kappa_{q}(\mathbf{x}, \mathbf{x}')$ is symmetric positive definite, we only need to ensure that every $\bm{B}_{q}$ is positive definite to produce a Mercer kernel. To do this, we parameterized $\bm{B}_{q}$ as
\begin{equation}
\begin{array}{cl}
 \bm{B}_{q}  &= \bm{A}_{q}\bm{A}_{q}^{\top} + 
 \begin{bmatrix}
 \lambda_{q, 1} & 0 & \cdots & 0 \\
 0 & \lambda_{q, 2} & \cdots & 0 \\
 \vdots & \vdots & \ddots & \vdots \\
 0 & 0 & \cdots & \lambda_{q, D}
 \end{bmatrix}
 = \bm{A}_{q}\bm{A}_{q}^{\top} + \text{diag}{(\bm{\lambda}_{q})},
\end{array}
\end{equation}

\begin{equation}
\bm{A}_{q} =
\begin{bmatrix}
a_{q, (1,1)} & \cdots & a_{q,(1, R_{q})} \\
\vdots & \ddots & \vdots \\
a_{q, (D,1)} & \cdots & a_{q,(D, R_{q})}. \\
\end{bmatrix}
\end{equation}
Here $\bm{A}_{q} \in \mathbb{R}^{D \times R_{q}}$, $\bm{\lambda}_{q} \in \mathbb{R}^{D \times 1}$. We let $R_{q}$ denote the number of non-zero columns in $\bm{A}_{q}$, or the rank for $\bm{B}_{q}$ when $\bm{\lambda}_{q} = \mathbf{0}$.

For any two observations from the same patient of different covariates at different times, denoted as $x_{i,d,t}$ and $x_{i,d',t'}$, the prior covariance from the GP kernel is
\begin{equation}
\kappa_{i}(x_{i,d,t}, x_{i,d',t'}) = \sum_{q=1}^{Q} b_{q, (d,d')} \kappa_{q}(x_{t}, x_{t'}).
\label{equation:element_kernel}
\end{equation}
We summarize the parameters and hyperparameters of our SM-LMC kernel in Table~\ref{tab:hyperparam}.

\begin{table}
\centering
\begin{tabular}{ccl}
\hline
notation & size & description\\
\hline
$v_{q}$ & $Q$ & squared exponential part of $q$th basis kernel\\
$\mu_{q}$ & $Q$ & periodicity of $q$th basis kernel\\
$a_{q,(d,r)}$ & $\sum_{q=1}^{Q} D \times R_{q}$ & weights of $(d, d')$ for $q$th basis kernel\\
$\lambda_{q, (d)}$ & $Q \times D$ & intra-covariate weights of the $d$th covariate for $q$th basis kernel\\
 & & $\bm{B}_{q}=\bm{A}_{q}\bm{A}_{q}^{\top} + \text{diag}(\bm{\lambda}_{q})$\\
\hline
\end{tabular}
\caption{The list of hyperparameters for modeling the $d = 1:D$ clinical variables and $q = 1:Q$ mixture kernels.}
\label{tab:hyperparam}
\end{table}

\subsection{Sparsity-Inducing Priors on Weight Matrix $\bm{B}_{q}$}
As the number of medical covariates included in the model increases, we need to increase the number of basis kernels $Q$ and corresponding $R_{q}$ to allow greater representational flexibility. However, too many basis kernels may lead to overfitting and will become computationally intractable. To avoid this, we regularized the elements of each weight matrix $\bm{B}_{q}$ by introducing structured sparsity-inducing priors on each $\bm{A}_{q}$ matrix as follows.

We included two layers of sparsity-inducing priors for flexible, data-adaptive shrinkage behavior, modified from previous work~\citep{Polson10shrinkglobally,Gao2014a}. First, we put column-wise sparsity-inducing priors to regularize each column in $\bm{A}_{q}$. This corresponds to regularizing the degree of freedom of the functions, or number of latent processes generated from each basis kernel in the LMC model~\citep{alvarez2011kernels}. Second, we put sparsity-inducing priors on each matrix element $a_{q, (d, r)}$ in $\bm{A}_{q}$ to produce element-wise sparsity. The effect of element-wise sparsity is to perform model selection on the number of basis kernels that each pair of covariates uses for covariance representation. Finally, we put sparsity-inducing priors on the elements of $\bm{\lambda}_{q}$ to shrink the covariance for observations from the same covariate.

In practice, we implemented each layer of the prior as a two-layer hierarchical gamma distribution. The generative model is written as 
\begin{equation}
\begin{aligned}
\tau_{q,(r)} & \sim \text{Gamma}(d, \eta),\\
\phi_{q,(r)} & \sim \text{Gamma}(\gamma, \tau_{q,(r)}),\\
\delta_{q,(d, r)} & \sim \text{Gamma}(\beta, \phi_{q,(r)}),\\
\psi_{q,(d,r)} & \sim \text{Gamma}(\alpha, \delta_{q,(d, r)}),\\
a_{q,(d, r)} & \sim \mathcal{N}(0, \psi_{q,(d,r)}),
\end{aligned}
\end{equation}
where each element $a_{q,(d, r)}$ has a Gaussian distribution. Parameters $\phi_{q,(r)}$ and $\tau_{q,(r)}$ control the column-specific shrinkage, while parameters $\psi_{q, (d, r)}$ and $\delta_{q,(d,r)}$ control the local shrinkage of each element in the $\bm{A}_{q}$ matrix. For vector $\bm{\lambda}_{q}$, we regularized each element with a local Laplace prior:
\begin{equation}
\lambda_{q, (d)} \sim \text{Laplace}(0, \beta_{\lambda}).
\end{equation}
For our results, we set $\alpha = \beta = \gamma = d = 0.5$ to recapitulate two layers of the horseshoe prior, using a statistically equivalent prior represented by a hierarchical gamma with four layers~\citep{Carvalho2010,armagan2011,Gao2014a,zhao2014bayesian-JMLR}. Parameters $\psi_{q,(d,r)}$, $\delta_{q,(d, r)}$, $\phi_{q,(r)}$, and $\tau_{q,(r)}$ were estimated during optimization. We set $\beta_{\lambda} = 0.01$ to regularize the diagonal terms $\lambda_{q, (d)}$.
The hyperparameter $\eta$ controls the overall shrinkage profile of the hierarchical gamma prior (see~Appendix A for more details). We chose $\eta$ over $\lbrace 0.01, 0.1, 1.0 \rbrace$ using cross-validation prediction error. Hyperparameter $\eta$ was chosen using grid search over the range $\lbrace 0.01, 0.1, 1.0 \rbrace$ using cross-validation prediction error as the objective.

\subsection{Parameter Learning}
\label{subsec:Parameter_Learning}

To estimate the parameters for the regularized kernel, we optimized the posterior probability. We denote all parameters that were estimated directly as $\bm{\theta}$ and hyperparameters in the sparsity-inducing prior as $\bm{\theta}_{f}$:
\begin{equation}
\begin{array}{c}
\bm{\theta} = \left\lbrace{\mu_{q}, v_{q}, a_{q, (d,r)}, \lambda_{q, (d)}, \psi_{q, (d,r)}, \delta_{q, (d,r)}, \phi_{q, (r)}, \tau_{q, (r)}}\right\rbrace,\\
\text{for } q = 1, \cdots, Q \quad d = 1, \cdots,D \quad r_{|q} = 1, \cdots, R_{q}
\end{array}
\end{equation}

\begin{equation}
\begin{array}{c}
\bm{\theta}_{f} = \left\lbrace{\alpha, \beta, \gamma, d, \eta, \beta_{\lambda}}\right\rbrace,\\
\alpha = \beta = \gamma = d = 0.5.
\end{array}
\end{equation}
The posterior density of our model is then
\begin{equation}
\begin{array}{ll}
p(\bm{\theta}|\mathbf{y}, \mathbf{x},\bm{\theta}_{f}) 
&\propto 
p(\mathbf{y}|\mathbf{x}, \bm{\theta})p(\bm{\theta}|\bm{\theta}_{f})\\

&\propto
p(\mathbf{y}|\mathbf{x}, \bm{\theta})
\displaystyle
\left[
\prod_{q=1}^{Q}\prod_{d=1}^{D}\prod_{r=1}^{R_{q}}
p(a_{q, (d, r)}|\psi_{q, (d,r)})
p(\psi_{q, (d,r)}|\alpha, \delta_{q, (d,r)})
p(\delta_{q, (d,r)}|\beta, \phi_{q, (r)})
\right]\\

& \times 
\displaystyle
\left[
\prod_{q=1}^{Q}\prod_{r=1}^{R_{q}}
p(\phi_{q,(r)}|\gamma, \tau_{q,(r)})p(\tau_{q,(r)}|d, \eta)
\right]
\left[
\prod_{q=1}^{Q}\prod_{d=1}^{D}
p(\lambda_{q, (d)}|\beta_{\lambda})
\right]
\left[
\prod_{q=1}^{Q}
p(v_{q})p(\mu_{q})
\right].
\\
\end{array}
\end{equation}
The term $p(\mathbf{y}|\mathbf{x}, \bm{\theta})$ is found by calculating the GP marginal likelihood given the values of $\bm{\theta}$~\citep{Rasmussen2006}, which is
\begin{equation}
\label{equation:GP_marginal_likelihood}
\log{p(\mathbf{y}|\mathbf{x}, \bm{\theta})} = -\frac{1}{2}\mathbf{y}^{\top}(K_{|\bm{\theta}}+\bm{\epsilon}I)^{-1}\mathbf{y} - \frac{1}{2}\log{|K_{|\bm{\theta}}+\mathbf{\epsilon}I|} - \left(\frac{\sum_{d=1}^{D}T_{i,d}}{2}\right)\log{(2\pi)}.
\end{equation}

We thus estimated $\bm{\theta}$ by solving the optimization problem:
\begin{equation}
\begin{array}{ll}
\displaystyle {
\arg\max_{\bm{\theta}}
} \quad
\log{p(\bm{\theta}|\mathbf{y}, \mathbf{x},\bm{\theta}_{f})} =&
\displaystyle {
\arg\max_{\bm{\theta}} \mathcal{Q}(\bm{\theta}) 
}.\\
\end{array}
\end{equation}
See Eq.~(\ref{equation:Q}) in Appendix B for the derivation of $\mathcal{Q}(\bm{\theta})$. 
 
Due to the conjugacy of the hierarchical gamma priors, we optimized parameters $\psi_{q,(d,r)}$, $\delta_{q,(d,r)}$, $\phi_{q,(r)}$, $\tau_{q,(r)}$ directly using maximum a posteriori (MAP) estimates of their posterior distribution (or mean when the mode does not exist). 
Our optimization procedure then consists of two parts. 
In the first part, we used the update equations to estimate $\psi_{q,(d,r)}$, $\delta_{q,(d,r)}$, $\phi_{q,(r)}$, and $\tau_{q,(r)}$ directly. In the second part, we estimated parameters $\mu_{q}$, $v_{q}$, $a_{q,(d, r)}$, and $\lambda_{q,(d,r)}$ using a scaled conjugate gradient method to find the local maximum, conditioned on $\hat{\psi}_{q,(d,r)}$, $\hat{\delta}_{q,(d,r)}$, $\hat{\phi}_{q,(r)}$, and $\hat{\tau}_{q,(r)}$. (Details can be found in Appendix B Eq.~(\ref{equation:prior_update_start})--(\ref{equation:prior_update_end}) and Eq.~(\ref{equation:GP_gradient_start})--(\ref{equation:GP_gradient_end}).) We iterated over the two steps until the change in $\mathcal{Q}(\bm{\theta})$ reached the convergence criterion ($<0.005$) or until the maximum number of iterations ($\geq 30$).

\subsection{Estimating the Population-Level Model and Online Updating}
The GP with the structured kernel described above lets us model the patient-specific joint dynamics between covariates within the same patient. We now describe how we built a population-level empirical prior from a set of mixture kernels estimated from all training patients, and how we apply this empirical prior to a new patient. 

\begin{figure}
    \centering
    \subfigure[]{\hfil \includegraphics[height=6cm]{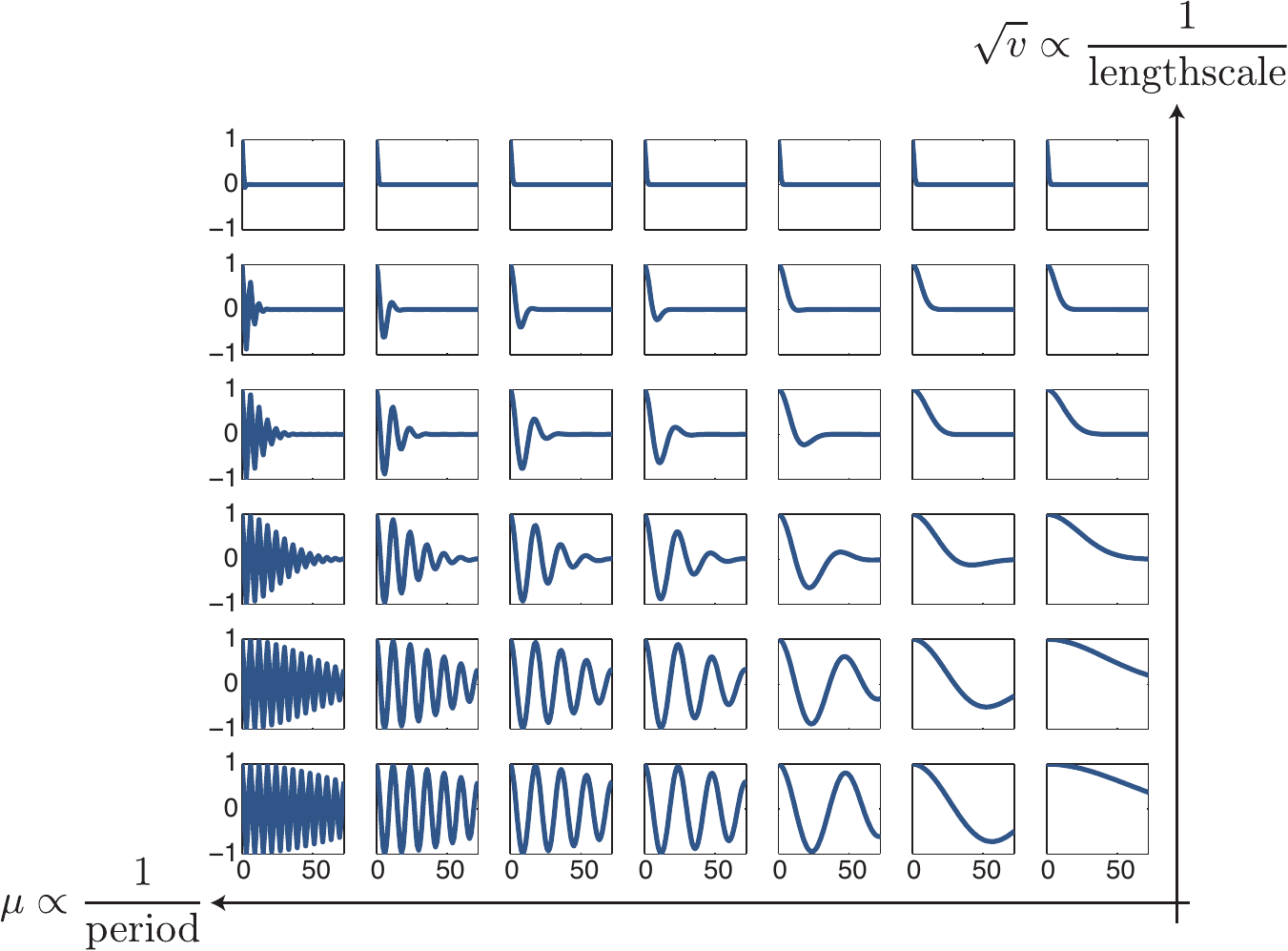}}
    \subfigure[]{\hfil \includegraphics[height=5cm]{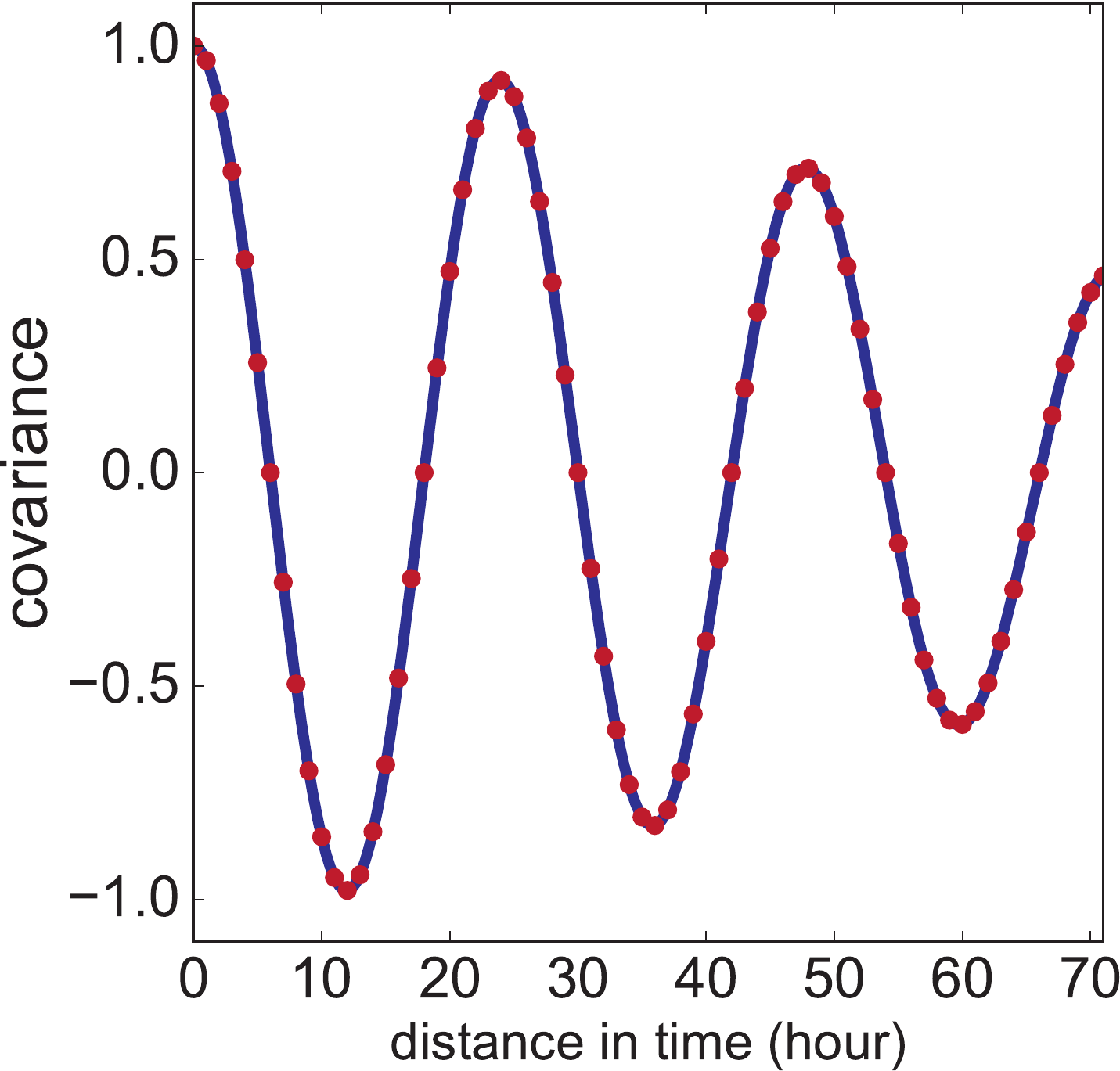}}
    \caption{\textbf{Illustrations of the basis kernels and the features for kernel clustering.} (a) An example of a discrete set of basis kernels with different $\mu$ and $v$ within a 72-hour window. (b) An example of the 72-dim temporal features (shown with red dots) taken from a kernel for GMM clustering.}
    \label{fig:SM_clustering}
\end{figure}

To estimate the empirical priors across reference patients, we trained one GP kernel for each patient separately, and then we clustered and extracted the distribution of the basis kernels (defined by hyperparameters $\mu_{q}$ and $v_{q}$). The idea here is that, when we observe a set of estimated mixture kernels, we would like to understand the high-level properties of these mixture kernels shared across covariates and patients in the same patient group, and then estimate the distributions of these hyperparameters through observations of basis kernels belonging to this cluster. For instance, a circadian rhythm (24-hour periodicity) may be observed in some covariates for some patients, but the period across patients could vary within a range. Across the space of $\mu$ and $v$ the spectral kernels vary substantially (Figure~\ref{fig:SM_clustering}a).  For each basis kernel that was estimated, the characteristic period is $\nicefrac{1}{\mu_{q}}$ and the length scale is $\nicefrac{1}{2\pi \sqrt{v_{q}}}$~\citep{SMkernel-WilsonA13}. There are different ways to define the features of a kernel. Here we used the temporal features of the learned kernels directly (Figure~\ref{fig:SM_clustering}b). The temporal spacing of two adjacent points is one hour, and we use the kernel values within a window of length 72 hours. 
We then used a Gaussian mixture model (GMM) model to perform clustering on the kernels, 
and we chose the best number of kernel clusters $Q'$ ($1 \leq Q' \leq Q$) based on Bayesian information criterion (BIC). For the MedGP implementation, we adapted the open source scikit-learn package~\citep{scikit-learn}. We used version 0.18.1, with ten random restarts, a maximum of 2,000 iterations, and allowing each mixture component to have its own covariance matrix.

For each identified kernel cluster, we estimated one set of parameters $\mu_{q}$ and $v_{q}$ for the basis kernel, and the weight coefficients---elements in $\bm{B}_{q}$ matrices, computed using the $\bm{A}_{q}$ matrices and $\bm{\lambda}_{q}$ vectors. We do this by building an empirical distribution using kernel density estimation (KDE) with a Gaussian kernel over the GP kernel hyperparameters assigned to that cluster. 
The bandwidth of the kernel density estimator was chosen based on Silverman's ``rule-of-thumb''~\citep{silverman1986density}. We estimated each new parameter using density weighted means with the density from the univariate KDE as the weights. Note that if there were multiple kernels in a patient cluster, the estimated $\bm{B}_{q}$ matrices were added based on the additive assumption of our kernel before aggregating to estimate the  population-level kernel for that cluster. To allow online updating, we estimated the elements of the new empirical $\bm{A}_{q}$ matrix and $\bm{\lambda}_{q}$ vector corresponding to each new $\bm{B}_{q}$ matrix using singular value decomposition (SVD). For the univariate GP regression, we did not use density weighted means because we found them to be unstable; instead we used a grid-based search to identify the hyperparameters with the highest kernel density estimates.

As the number of vital signs and lab measurements of a new patient accumulate, we update the hyperparameters to estimate a patient-specific kernel. Indeed, we update the kernel sequentially every time a new observation arrives. To do this in a computationally tractable way, we used the momentum method~\citep{rumelhart1988learning} with a 72-hour window of previous observations to update the kernel hyperparameters when predicting the value of next observation. For all experiments, we chose the momentum as $0.9$ and the learning rate as $10^{-5}$. For elements in the $\bm{A}_{q}$ matrices, we do not update the values if the elements were regularized to be zero so as to maintain the empirical sparsity structure.

\subsection{Efficient Inference in MedGP}
The main bottleneck of our method is in learning patient-specific kernel hyperparameters. Let $T_{i} = \sum_{d=1}^{D}{T_{i,d}}$ denote the total number of samples of the $i$th patient; the computational cost to compute the Gram matrix is $\mathcal{O}(QT_{i}^{2})$, which increases linearly with the chosen number of basis kernels. To find the MAP estimates of the parameters, we need to invert and compute the determinant of the Gram matrix $(K_{|\bm{\theta}}+\bm{\epsilon}I)$ in Eq.~(\ref{equation:GP_marginal_likelihood}). The computational complexity for the full matrix inversion is $\mathcal{O}(T_{i}^{3})$ using Cholesky decomposition. When calculating the gradients for optimizing the hyperparameters, the cost is dominated by $\mathcal{O}(QDRT_{i}^{2})$ after the inverse Gram matrix is pre-computed, which is linear with the total number of the kernel hyperparameters. In practice, the complexity of each iteration is either $\mathcal{O}(T_{i}^{3})$ or $\mathcal{O}(QDRT_{i}^{2})$. That is, the patient with the most measurements is the main bottleneck for training. In our implementation, we mitigate the bottleneck using optimized linear algebra functions in Intel MKL library with multithreading and computing the gradients of the hyperparameters in parallel.

\subsection{Medical Data Preprocessing}
The HUP medical time series data consist of electronic health records (EHRs) from more than 260,000 patients admitted to a University of Pennsylvania Hospital. For each patient, the data include many heterogeneous clinical covariates, including ICD-9 codes, patient demography, length-of-stay, vital signs, and lab results. We jointly modeled the 24 covariates with the greatest number of observations across patients (Table~\ref{table:covariate_list_24}). We selected three groups of discharged patients from these data: 1,365 septic patients, 952 patients with heart failure, and 4,723 patients with neoplasms. Each patient has at least one observation for each of the 24 covariates, and in total over four million observations were evaluated.

For each clinical covariate, we first removed obvious artifacts (e.g., values outside of the possible range in living humans). For the patients with neoplasms or heart failure, we used the full patient length-of-stay in training and testing. For septic patients, the disease progression varies substantially across patients, and the distribution of the covariates changes dramatically depending on the disease phase. To address this issue, we segmented the time series data into four disjoint partitions based on clinical status: \emph{no sepsis}, \emph{pre-sepsis}, \emph{sepsis}, and \emph{recovery}. To label each stage, we incorporated prior clinical domain knowledge. For instance, we identified sepsis stages using ICD-9 codes and positive blood culture results. Since our model assumes stationarity, to better estimate the temporal correlation across covariates, we chose the \emph{recovery} stage before the patients' discharge to test our method, since this is a relatively stable stage. We used the bed unit information to identify if the patient is in a stable state. That is, when a patient is transferred to step-down bed, we labeled the time series after the transfer as \emph{recovery}. The median length-of-stay after pre-processing is 140 hours for the sepsis group, 285 hours for the heart failure group, and 197 hours for the neoplasms group.

We applied similar preprocessing procedure to the MIMIC-III data. We selected patients with a heart failure diagnosis that eventually had a routine discharge. We removed artifacts such as out of bounds values for each covariate, and applied the criteria to each patient that at least five measurements were taken for all 24 selected covariates. We extracted 1,004 heart failure under these criteria and used 1,003 of them, excluding one patient with more than 50K measurements due to memory constraints.

\subsection{Experimental Setup}
\label{subsec:Experiment_Setup}
We applied MedGP to the three selected groups of patients separately, and evaluated characteristics and performance of MedGP under two different experimental settings. In the first analysis, we evaluated the model's ability to learn the covariance between a pair of highly correlated clinical covariates, and we measured the imputation performance in an online setting. In the second analysis, we follow the same online setting, but instead jointly model all 24 clinical covariates, including four vital signs and 20 lab covariates. In both settings, we evaluated our method using 10-fold cross-validation at the patient level. That is, for each fold we ran the kernel clustering step on the kernels from the training patients to estimate a set of population-level basis kernels and $\bm{B}_{q}$ matrices. This set of kernels was then applied to the held-out patients to predict the value of each covariate using observations from all other covariates measured at the same time as, or earlier than, the test observation (i.e., no future information included). After each prediction, we updated the patient-specific kernel parameters using the new observations from the test patient.

We compared our method to several univariate methods that modeled each covariate separately: (i) a naive one-lag prediction procedure, which predicts an observation equal to the last observation available from the same patient; (ii) an independent GP with squared exponential (SE) or spectral mixture (SM) kernels fitting each covariate separately (we tested with $Q=1$ for SM); (iii) the multi-resolution Probability Subtyping Model (PSM) combining linear regression, B-splines, and independent GPs~\citep{suchi2015clustering}. To estimate the spectral kernel parameters, for each patient we initialized $1,000$ random kernels by drawing uniformly from a length scale range (between 6 and 72 hours) and period range (between 24 and 72 hours). We computed the marginal likelihood of all random kernels for each patient, and then initialize optimization using the kernels with the highest marginal likelihood. The elements in the $\bm{A}_{q}$ matrices are initialized randomly between $-1.5$ and $1.5$.

We compared results from MedGP to these various methods using two metrics: (i) mean absolute error (MAE) of the predicted observations with the true observations, and (ii) 95\% coverage, the percentage of true observations that fell within the predictive 95\% confidence region. We quantified and reported the improvements with respect to both metrics compared to all three baselines (naive prediction, univariate GP, and PSM). To test if the differences in prediction results from different approaches were statistically significant, we performed paired t-tests for the results of each covariate and compared the $p$-values with a Bonferroni corrected threshold (dependent on the number of jointly modeled covariates in each experiment).

We note that the original PSM was designed to model scleroderma disease~\citep{suchi2015clustering}. Thus, to make it applicable to our different patient groups, several adjustments were made. First, we omitted the population and environmental factors selected for their relevance to scleroderma. Second, we chose the knots of the B-spline basis by sampling every hour for vital signs and every 24 hours for lab results between zero and the longest length-of-stay for patients in each disease group. Third, to make PSM training feasible on the scale of our data set, we limited the maximum number of subtypes to ten for the sepsis and heart failure groups, and 20 for the neoplasms group.

%% file: Results.tex
\section{Experimental Results}
\label{sec:Results}
We analyzed the performance of the method, MedGP---multi-output GP with a sparse SM-LMC kernel and online updating---by applying it to time series data from the Hospital of the University of Pennsylvania (HUP) and the public MIMIC-III data set~\citep{johnson2016mimic}. We ran two types of experiments---one on two correlated lab covariates and the other with 24 covariates jointly. The results were compared against the baseline methods for prediction accuracy and 95\% coverage calibration.

\subsection{Results of Two Lab Covariates}
\label{subsec:two_covariates}

\begin{figure*}[t]
\centering
    \includegraphics[width=15.5cm]{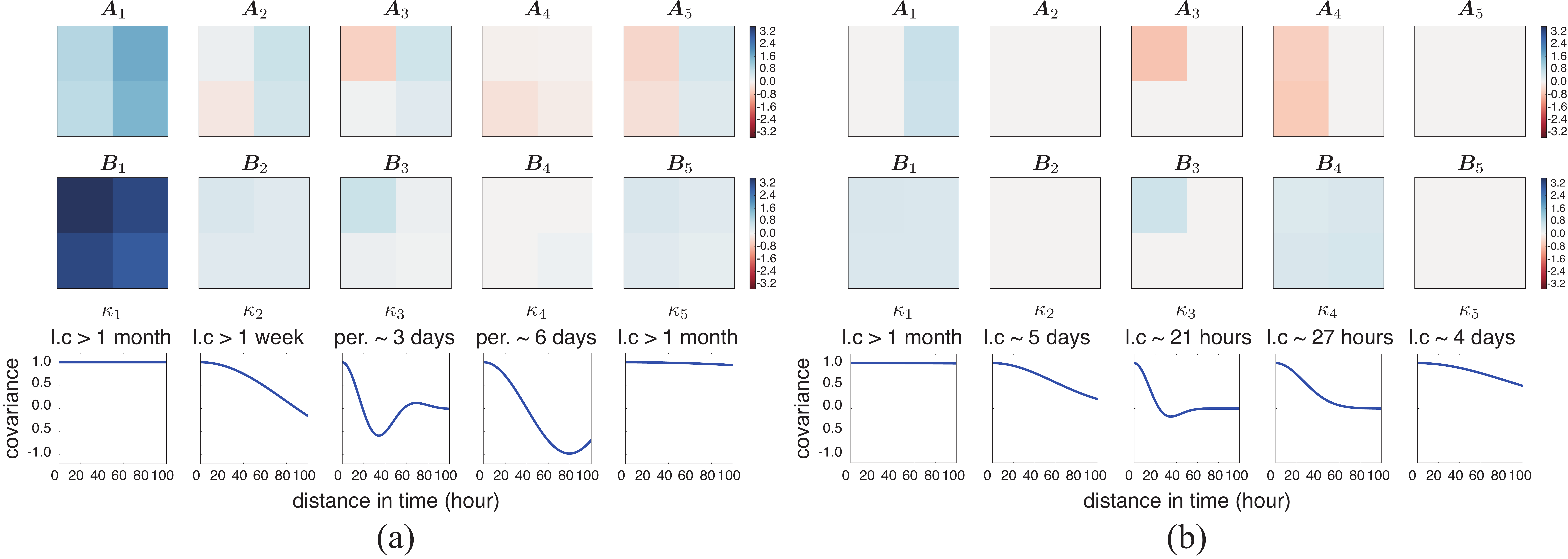}
    \caption{{\bf The trained kernel for one patient jointly modeling PT and INR.} For both the (a) SM-LMC kernel and (b) sparse SM-LMC kernel, the $\bm{A}_{q}$ matrices (upper row), $\bm{B}_{q}$ matrices (middle row), and the basis kernel $\kappa_{q}$ (bottom row) are illustrated. The zero elements are colored in light grey. Here \emph{l.c} denotes length scale for each basis kernel, and \emph{per.} denotes period. The length-of-stay for this patient was over 90 days.}
    \label{fig:single_case_kernel}
\end{figure*} 

As a proof of principle, we jointly modeled two well correlated lab covariates, prothrombin time (PT) and international normalization ratio (INR) on three HUP subgroups. PT measures the time it takes for the plasma in the blood to clot, and is often ordered to check bleeding problems. INR is an international standard for PT to account for possible variations across different labs. For the same patient, the two covariates usually have similar trajectories over time (Figure~\ref{fig:time_series_example}). 

We trained the kernels for one patient's INR and PT time series data both with and without the structured sparse prior (Figure~\ref{fig:single_case_kernel}). Both $\bm{A}_{q}$ and $\bm{B}_{q}$ matrices estimated using the sparse prior have higher levels of sparsity versus those estimated without using the sparse prior. We observed that for both methods, one of the estimated basis kernels $\kappa_{1}$ captures long-term (around one month) dependencies. However, with the sparse prior, the estimated weights associated with this long term kernel $\bm{A}_{1}$ are rank one instead of rank two. This means the trajectories of the two covariates are similar enough to be explained by one instead of two functions, and thus fewer hyperparameters. 
Moreover, two basis kernels were found with zeros weights $\bm{A}_{2}$ and $\bm{A}_{5}$ (Figure~\ref{fig:single_case_kernel}b), suggesting that the prespecified number of basis kernels may be reduced. We also found that the off-diagonal elements in the $\bm{B}_{q}$ matrices in both cases have nonzero values, suggesting a nonzero covariance between PT and INR observations. In particular, two basis kernels captured the covariance between PT and INR: one with a greater than one-month trend (Figure~\ref{fig:single_case_kernel}b, $\bm{B}_{1}$ and $\kappa_{1}$), and one with a 27-hour trend (Figure~\ref{fig:single_case_kernel}b, $\bm{B}_{4}$ and $\kappa_{4}$). Here, the sparse kernel has 18 non-zero hyperparameters, whereas there are 40 for the non-sparse kernel. 
We can compare the two fitted kernels using both log marginal likelihoods and model selection scores. The log marginal likelihoods of the two kernels are $-118.16$ (SM-LMC) and $-128.50$ (sparse SM-LMC), indicating a better fit for the SM-LMC model without sparsity. However, the Bayesian information criterion (BIC) values, which take into account the number of parameters in a model, were $353.63$ (SM-LMC) and $309.79$ (sparse SM-LMC), where values closer to zero reflect better models. Thus, using a sparse prior has the advantage of a more compact kernel representation.

\begin{figure*}[t]
\centering
\includegraphics[width=15.5cm]{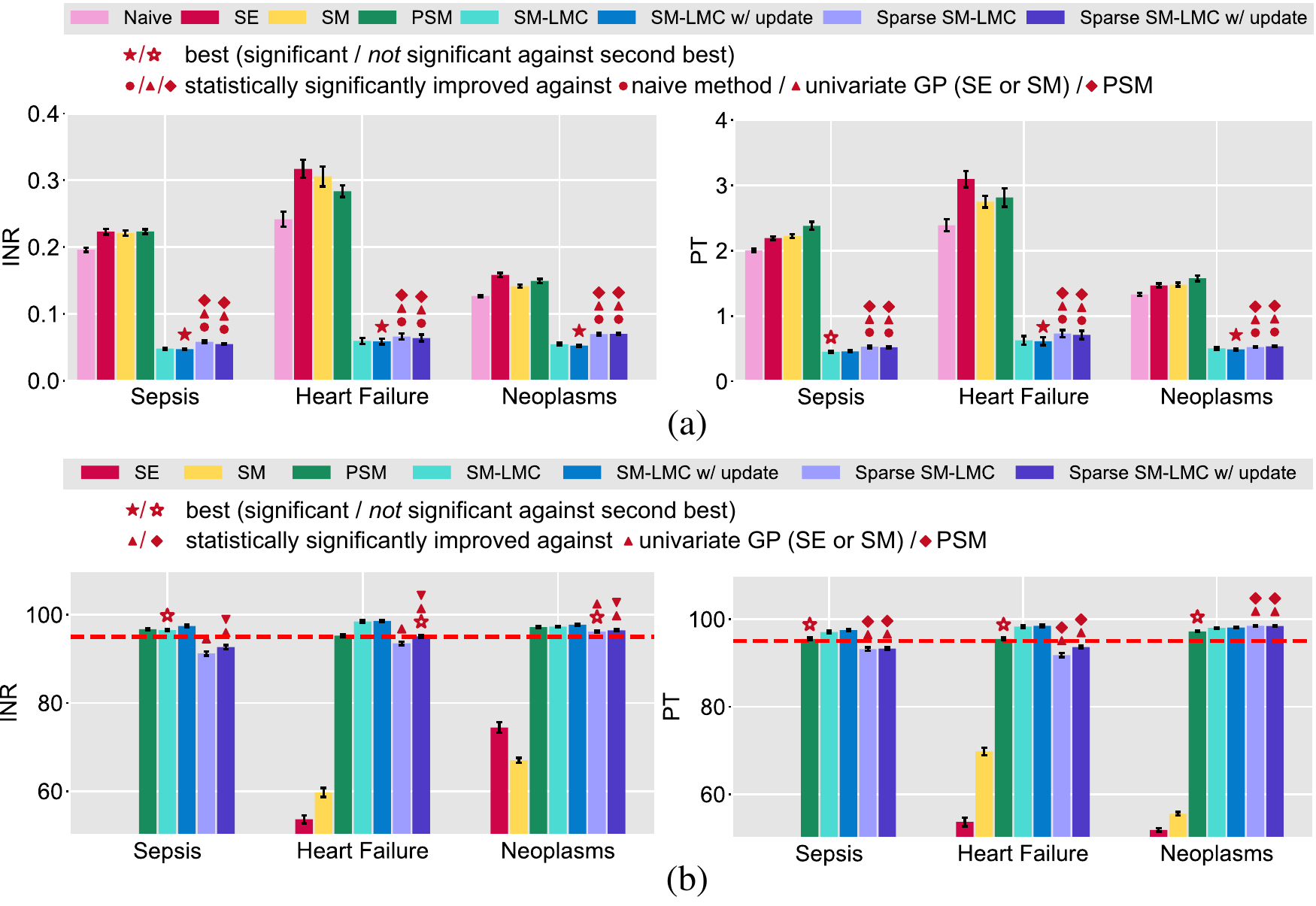}
\caption{{\bf The results of prediction when jointly modeling INR and PT.} The figure illustrates (a) mean absolute error (MAE), and (b) 95\% coverage (the dashed red line indicates 95\%). The error bars denote $\pm 1$ standard error.}
\label{fig:INR_PT_impute_online_one}
\end{figure*}

We then ran our model on all three disease groups separately, and compared our method with the univariate baselines described in Section~\ref{subsec:Experiment_Setup} under the scenario of online imputation of the same two well-correlated clinical covariates. For independent GPs, we used gradient descent to optimize the hyperparameters. For PSM, we performed grid search for the parameters of the B-spline and the independent GP kernel. For our method, we set $Q=5$ and $R_{q}=2$ for the $\bm{A}_{q}$ matrices for training. In the sepsis and heart failure groups, three nonzero basis kernel functions ($Q' = 3$) were found for the model using the SM-LMC kernel, while only two nonzero basis kernel functions ($Q' = 2$) were found using the sparse SM-LMC kernel; the number of nonzero hyperparameters were $18$ and $12$ respectively. In the neoplasms group, the number of nonzero basis kernels were the same as the pre-specified number ($Q' = Q = 5$). With 10-fold cross-validation, we found that results using the SM-LMC kernel showed smaller imputation error than those using the baselines for both PT and INR (Figure~\ref{fig:INR_PT_impute_online_one}). The mean absolute errors (MAEs) showed that the non-sparse SM-LMC kernels perform imputation the best among the related approaches. On the other hand, looking at the 95\% coverage, results using non-sparse or sparse SM-LMC kernel were well calibrated with respect to the confidence region compared with independent GPs, although sometimes slightly worse than PSM. Note that in this experiment we used a p-value threshold $p < 0.005$ to detect statistical significance, which reflects the Bonferroni correction.
The results indicate that the sparse prior finds models with sparse structure while maintaining prediction performance in this two covariate case.

\subsection{Results of a Joint Model Including 24 Vital Signs and Lab Covariates}
\label{subsec:results_cov24}
In the second experimental setting, we jointly modeled 24 vital signs and clinical covariates ($D=24$) for all three disease groups (Table~\ref{table:covariate_list_24}). We set the number of basis kernels $Q=5$ and the number of nonzero columns in $\bm{A}_{q}$ as $R_{q}=8$ in this experiment for the three HUP subsets. For the MIMIC-III heart failure subset, we set $Q=4$. More detailed results of the best setup as well as the results with different $Q$ could be found in Appendix C and Appendix D.

\subsubsection{Estimating Population-Level Kernels}
\begin{figure*}
 \centering
	\subfigure[]{\hfil \includegraphics[height=8.2cm]{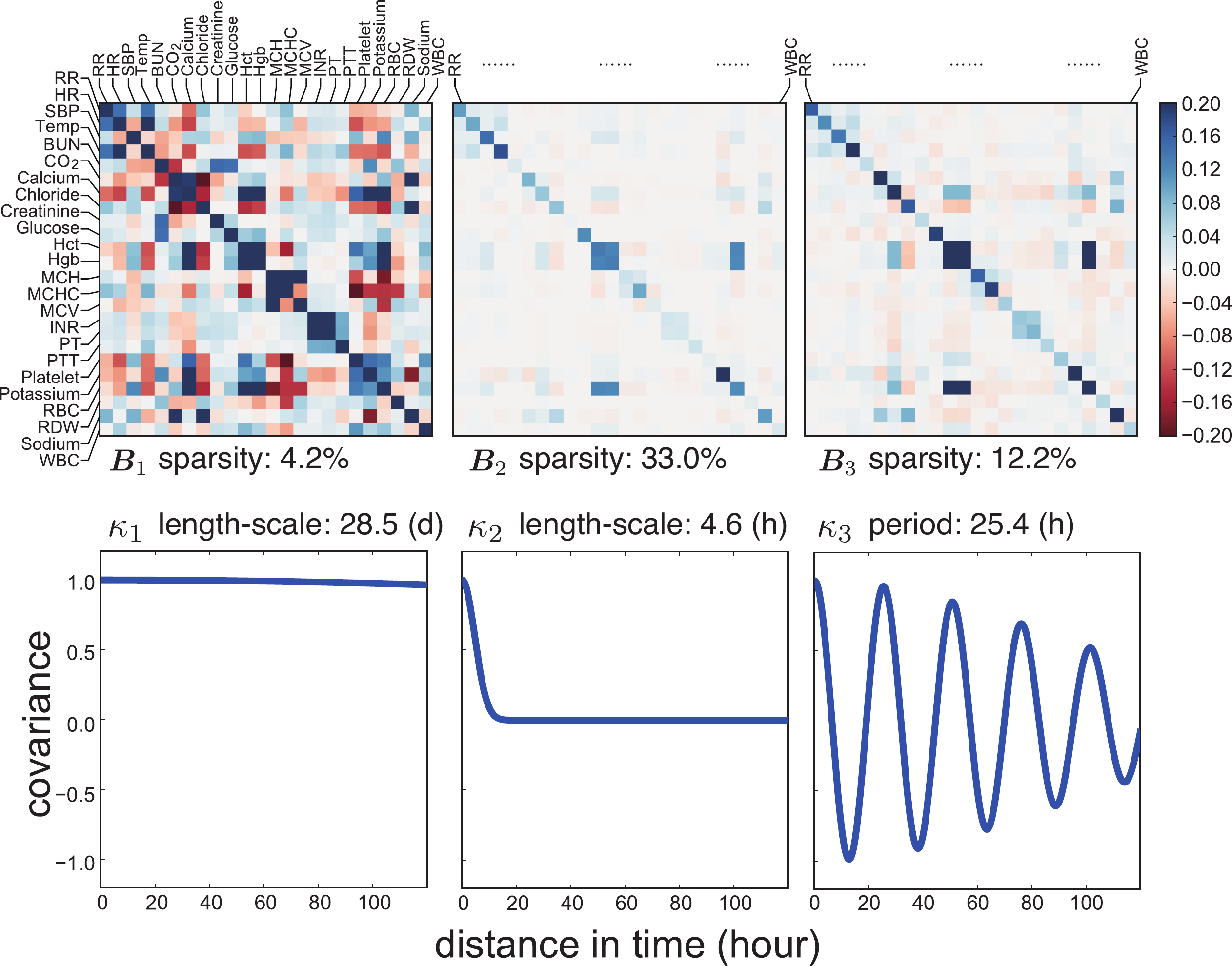}}
    \subfigure[]{\hfil \includegraphics[height=8.2cm]{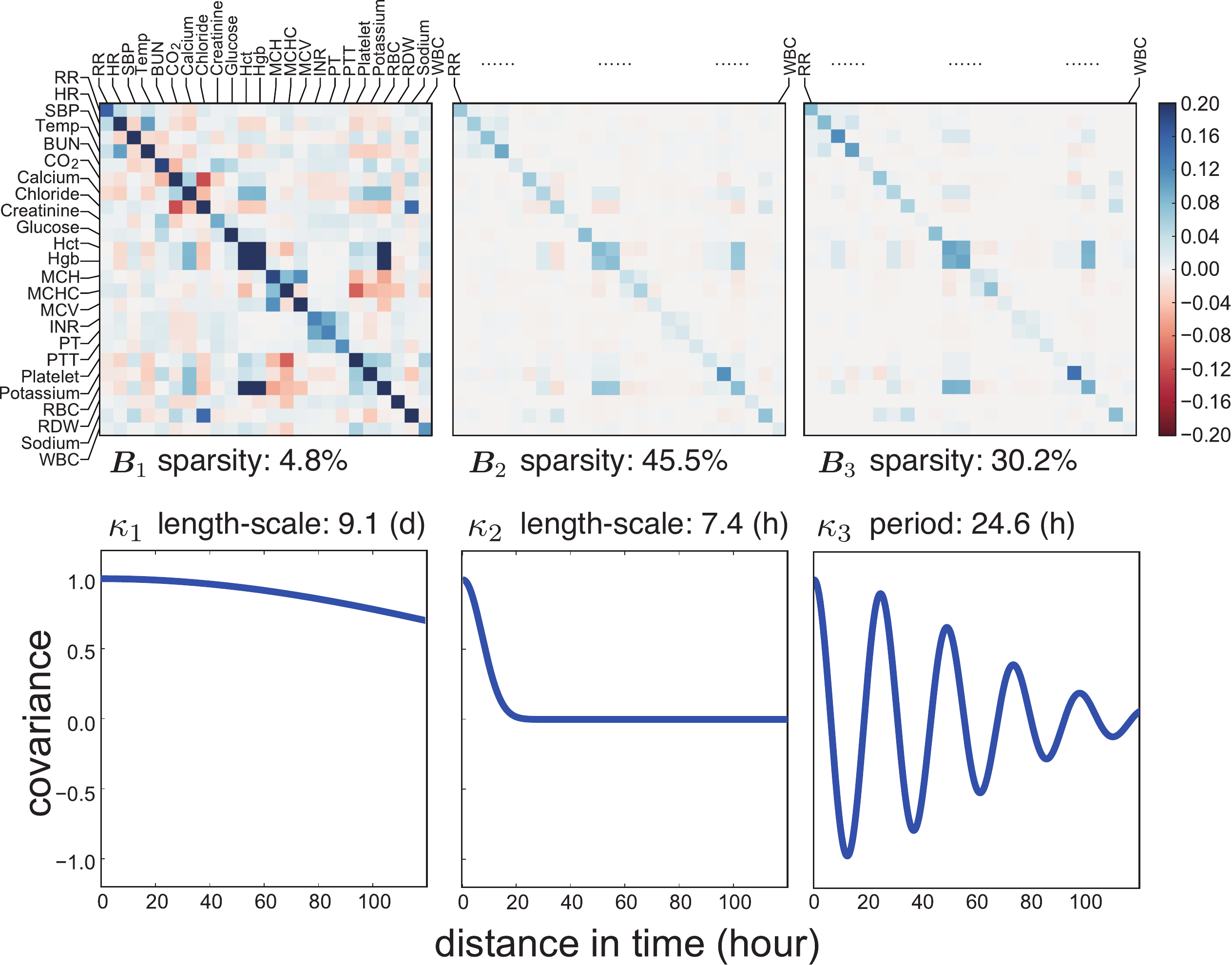}}
    \caption{\textbf{The estimated population-level basis kernels and corresponding $\bm{B}_{q}$ matrices for septic patients.} We show the kernels estimated (a) without a sparse prior ($Q'=3$) and (b) with a sparse prior ($Q'=3$). The sparsity of the $\bm{B}_{q}$ matrices is calculated as the percentage of nearly zero entries (i.e., values $\leq 10^{-3}$). The units for length scale or period are (d) for days and (h) for hours.}
    \label{fig:learned_kernel_24cov_sepsis}
\end{figure*}

\begin{figure*}
 \centering
	\subfigure[]{\hfil \includegraphics[height=8.2cm]{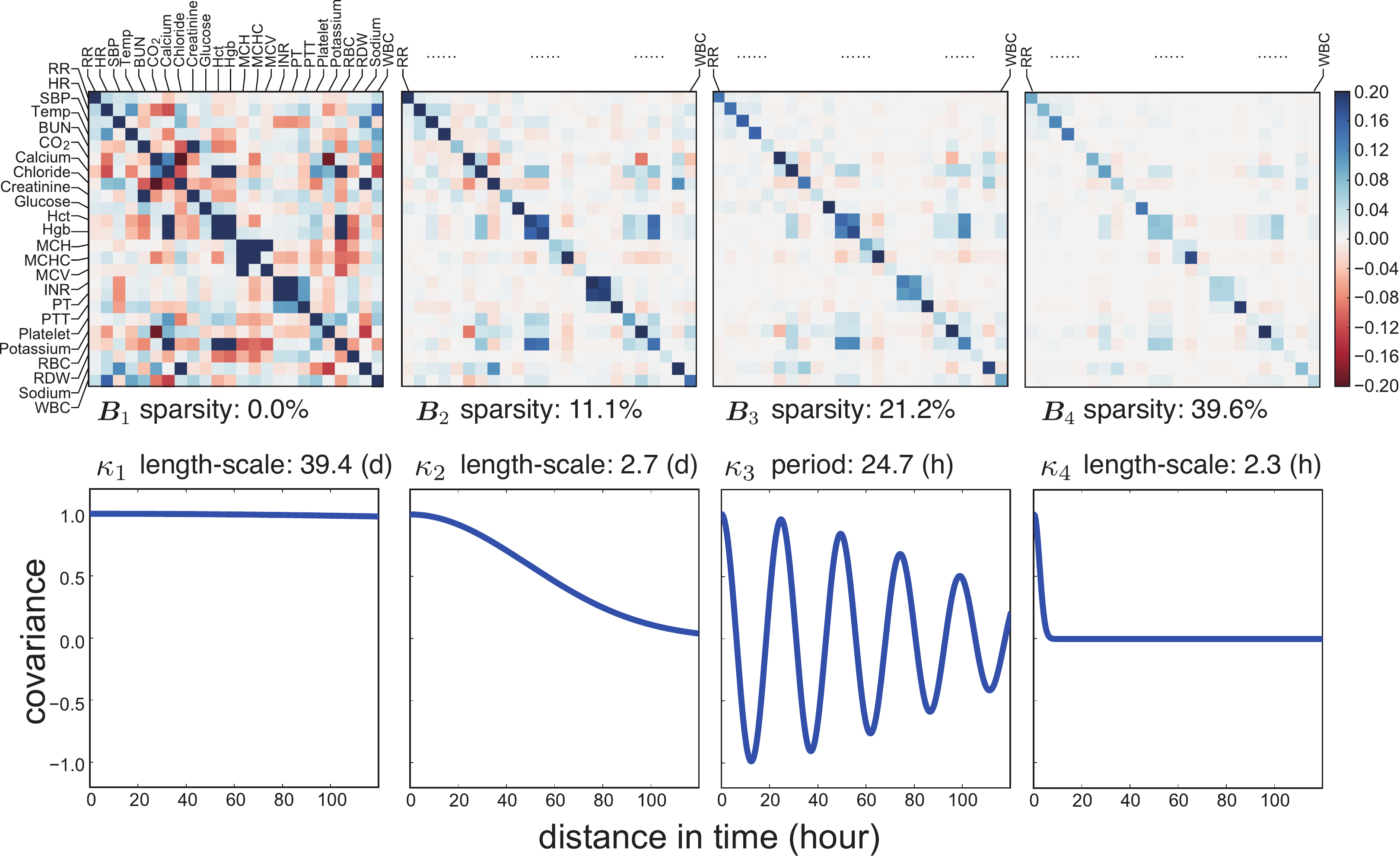}}
    \subfigure[]{\hfil \includegraphics[height=8.2cm]{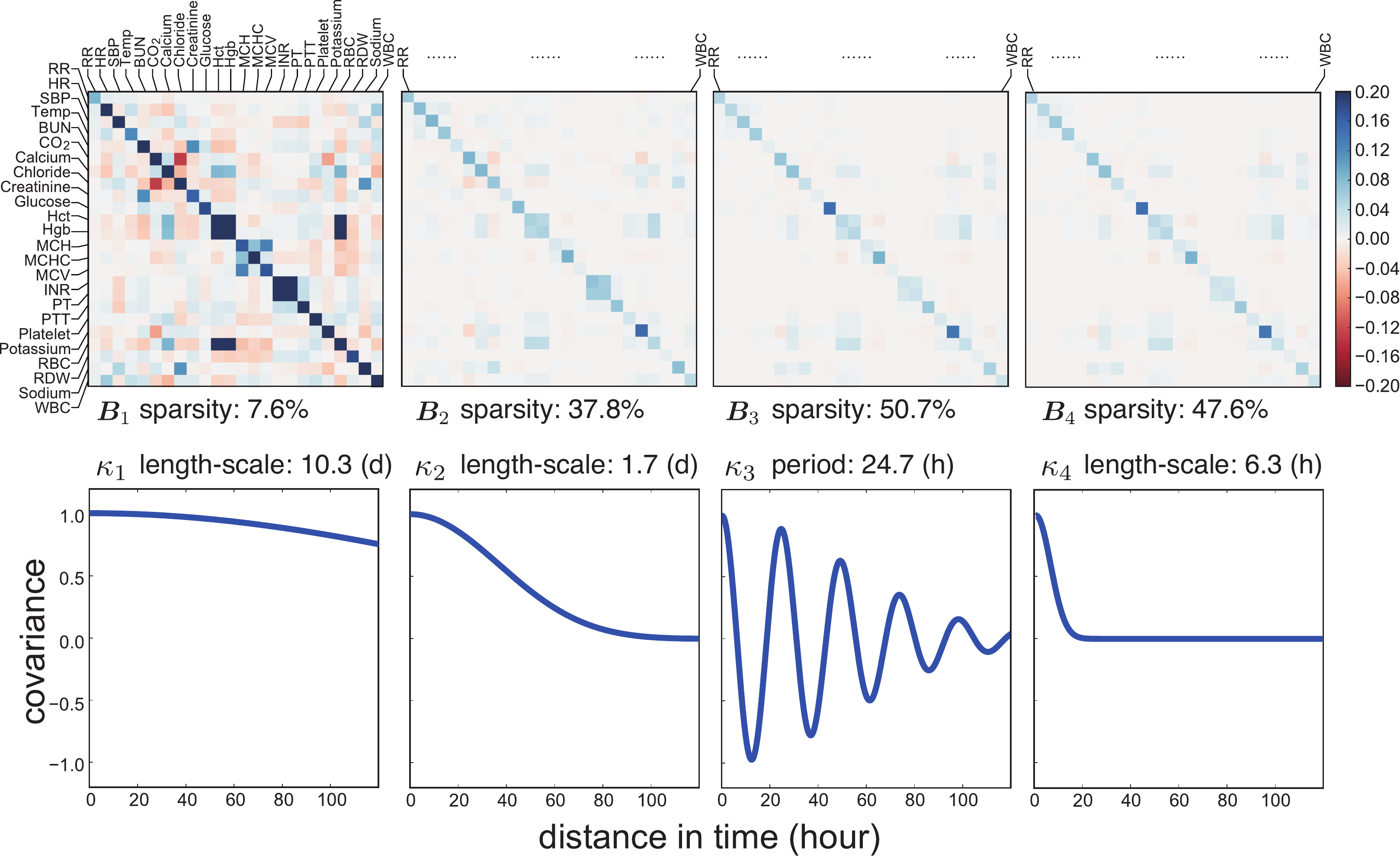}}
    \caption{\textbf{The estimated population-level basis kernels and corresponding $\bm{B}_{q}$ matrices for patients with heart failure.} We show the kernels estimated (a) without a sparse prior ($Q'=4$) and (b) with a sparse prior ($Q'=4$). The sparsity of the $\bm{B}_{q}$ matrices are calculated as the percentage of nearly zero entries (i.e., values $\leq 10^{-3}$). The units for length scale or period are (d) for days and (h) for hours.}
    \label{fig:learned_kernel_24cov_heart_failure}
\end{figure*}

\begin{figure*}
 \centering
	\subfigure[]{\hfil \includegraphics[height=8.2cm]{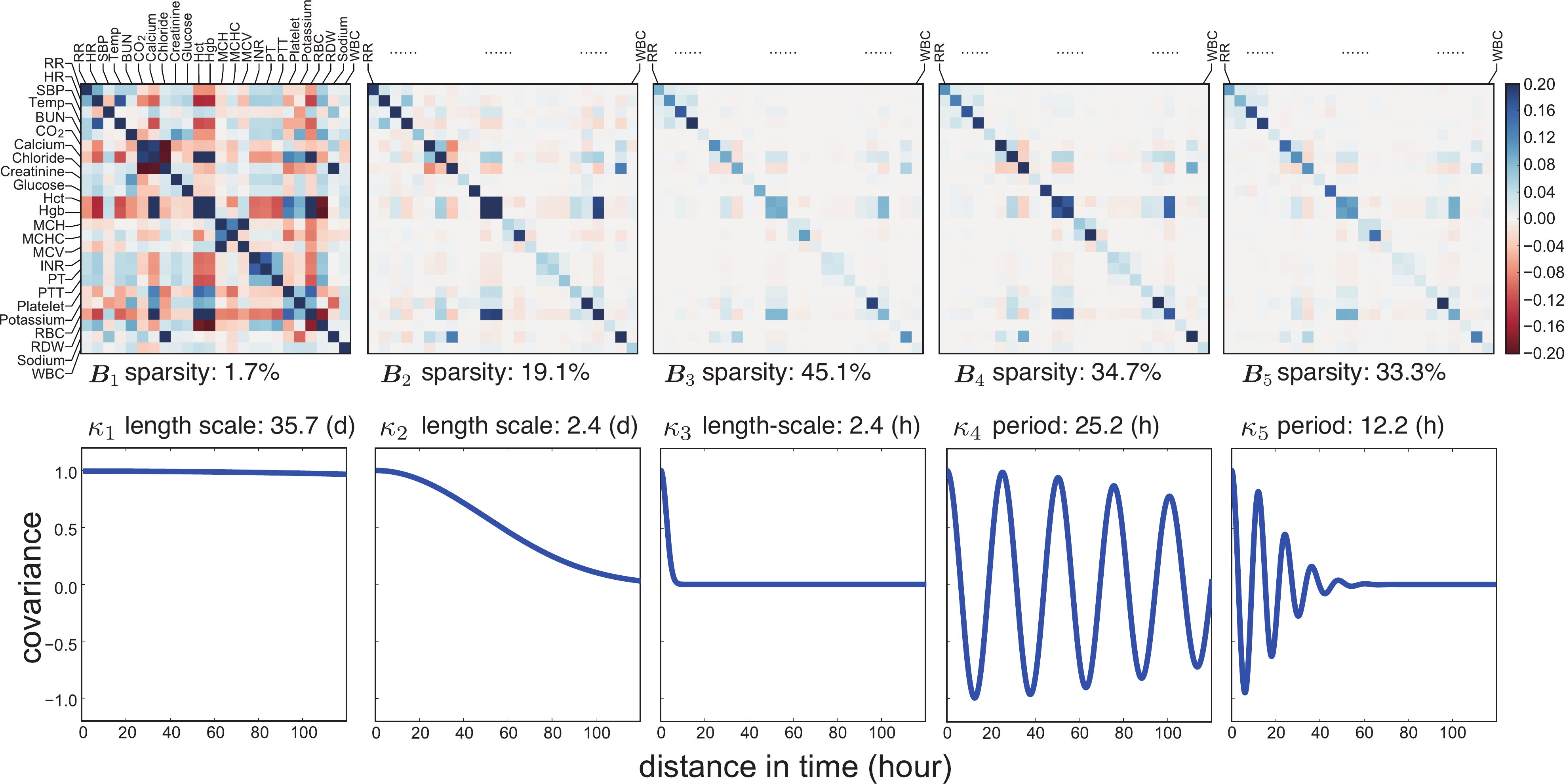}}
    \subfigure[]{\hfil \includegraphics[height=8.2cm]{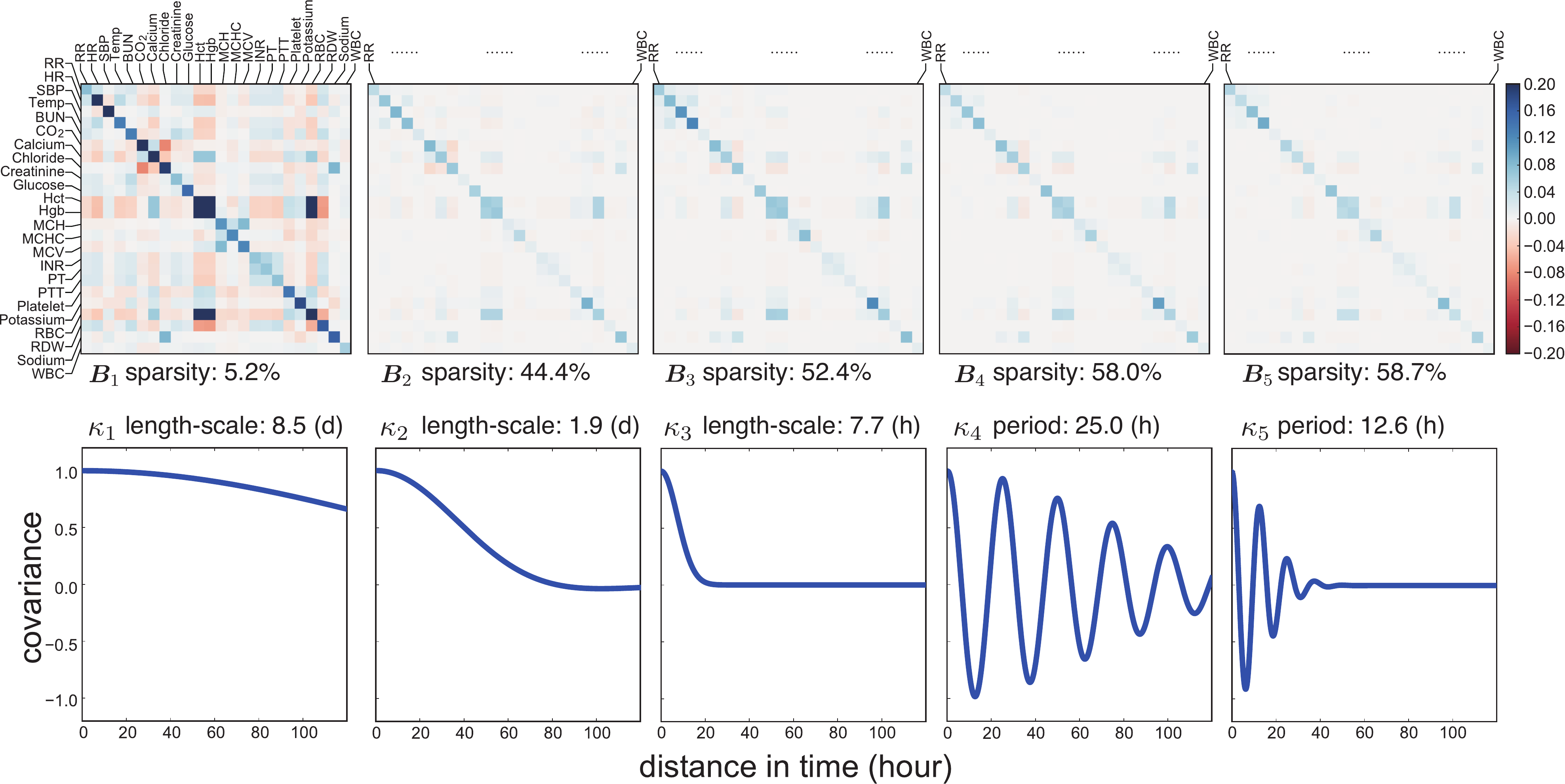}}
    \caption{\textbf{The estimated population-level basis kernels and corresponding $\bm{B}_{q}$ matrices for patients with neoplasms.} We show the kernels estimated (a) without a sparse prior ($Q'=5$) and (b) with a sparse prior ($Q'=5$). The sparsity of the $\bm{B}_{q}$ matrices are calculated as the percentage of nearly zero entries (i.e., values $\leq 10^{-3}$). The units for length scale or period are (d) for days and (h) for hours.}
    \label{fig:learned_kernel_24cov_neoplasms}
\end{figure*}

\begin{figure*}
 \centering
	\subfigure[]{\hfil \includegraphics[height=8.2cm]{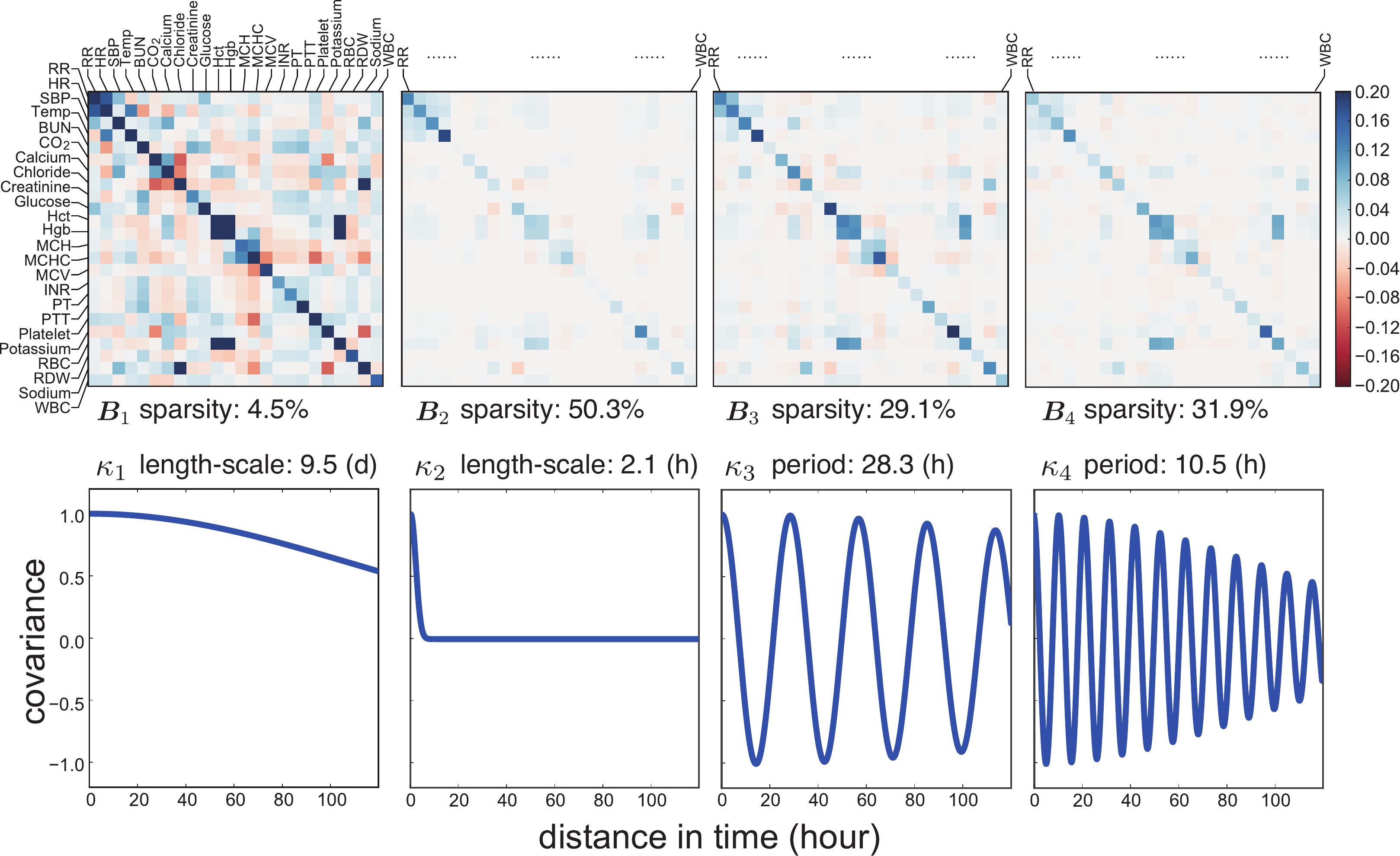}}
    \subfigure[]{\hfil \includegraphics[height=8.2cm]{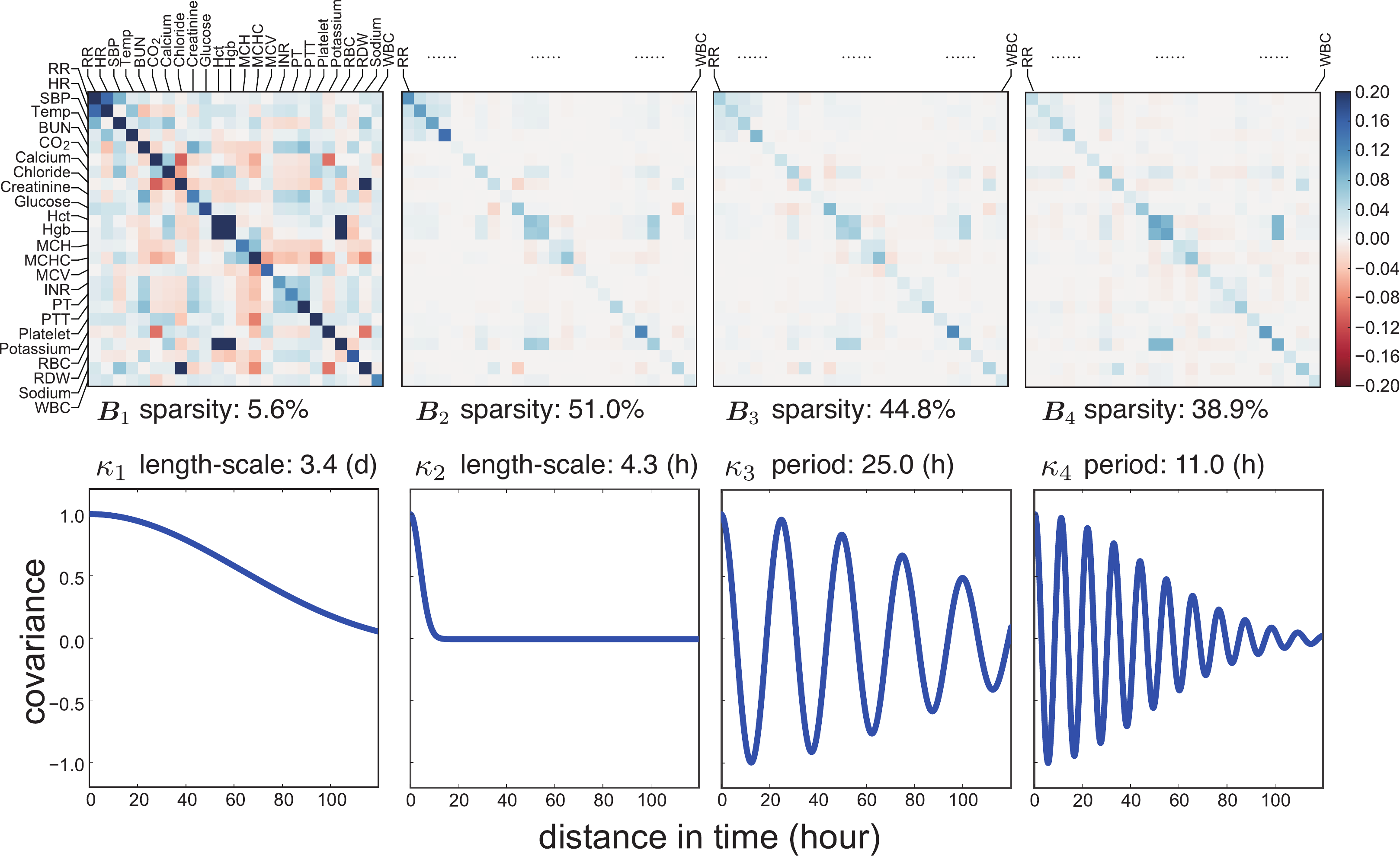}}
    \caption{\textbf{The estimated population-level basis kernels and corresponding $\bm{B}_{q}$ matrices for 1003 patients with heart failure in MIMIC-III data set.} We show the kernels estimated (a) without a sparse prior ($Q'=4$) and (b) with a sparse prior ($Q'=4$). The sparsity of the $\bm{B}_{q}$ matrices is calculated as the percentage of nearly zero entries (i.e., values $\leq 10^{-3}$). The units for length scale or period are (d) for days and (h) for hours.}
    \label{fig:learned_kernel_24cov_mimic}
\end{figure*}

We first visualized the population-level kernels estimated from the three patient groups of the HUP data (Figure~\ref{fig:learned_kernel_24cov_sepsis}--\ref{fig:learned_kernel_24cov_neoplasms}) and the MIMIC-III patient subgroup (Figure~\ref{fig:learned_kernel_24cov_mimic}). We observed shared patterns in the basis kernels $\kappa_{q}$ and the weight matrices $\bm{B}_{q}$ across all patient groups. Comparing the estimated population-level kernels, we found at least one long-term smoothing basis kernel with length scale longer than three days, and one 24- to 25-hour periodic basis kernel, which indicates the existence of circadian rhythms in specific covariates as expected. Furthermore, in the neoplasms group, which consists of more patients than the other two groups, we found additional short-term smoothing basis kernels and one 12- to 13-hour periodic basis kernel, which may correspond to known circasemidian rhythm of clinical covariates, such as body temperature. We also observed an 11-hour periodic kernel in the MIMIC-III subset.

In addition to the characteristics of the basis kernels, our model with the sparse prior also showed interpretable cross-covariate patterns (Figure~\ref{fig:learned_kernel_24cov_sepsis}b, Figure~\ref{fig:learned_kernel_24cov_heart_failure}b, and Figure~\ref{fig:learned_kernel_24cov_neoplasms}b and Figure~\ref{fig:learned_kernel_24cov_mimic}b). Based on the $\bm{B}_{q}$ matrices, we identified groups of well correlated covariates. For instance, lab covariates hematocrit (Hct), hemoglobin (Hgb), and red blood cell (RBC) count showed the highest levels of correlation. Since both Hct and Hgb are known to be proportional to the number of red blood cells, this positive correlation was encouraging~\citep{HumanPhysiology}. 
The pair of lab covariates studied in the previous section, INR and PT, also showed substantial positive correlation. We found that the four vital signs---respiration rate (RR), heart rate (HR), systolic blood pressure (SBP), and body temperature (Temp)---had substantial correlations with each other as well as weak correlations with some lab covariates. Another identifiable set of well-correlated covariates includes lab measurements of carbon dioxide (CO$_{2}$), calcium, chloride, potassium, and sodium. The three lab covariates related to the concentration of hemoglobin---mean cell hemoglobin (MCH), mean cell volume (MCV), and mean cell hemoglobin concentration (MCHC)---appeared to have substantial correlation (Figure~\ref{fig:learned_kernel_24cov_sepsis}). The correlations modeled in these covariance matrices are exploited for accurate prediction and imputation in the MedGP framework.

\begin{figure*}[t]
 \centering
    \includegraphics[width=13cm]{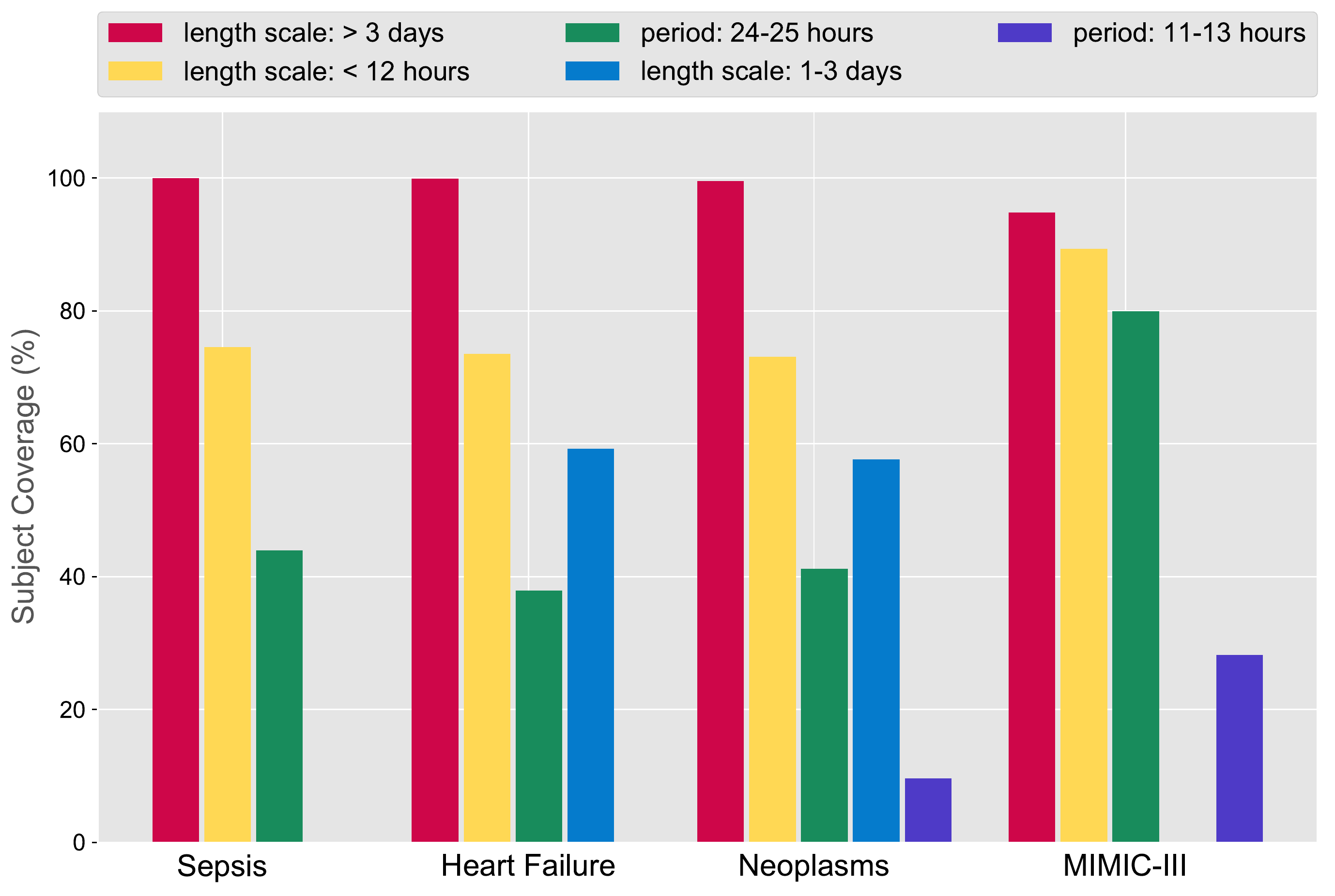}
    \caption{{\bf The coverage over subject for each discovered kernel.} We illustrated the proportion of subjects that have non-zero  $\bm{B}$ matrix of a kernel that is in the same cluster of the population-level kernel clusters.}
    \label{fig:all_group_sub_coverage}
\end{figure*}

To learn more about the importance of each kernel type across all subsets, we visualized the percent coverage of each type of kernel clusters found in all the subsets we have worked on (Figure~\ref{fig:all_group_sub_coverage}). The coverage of each kernel type is computed as the ratio of patients that have non-zero $\bm{B}_{q}$ matrix corresponding to it. We found that the kernel clusters of long-term (length scale $> 3$ days) and short-term (length scale $< 12$ hours) smooth dependencies have the highest coverage across four subsets. In the MIMIC-III subset, the coverages of the short-term kernel, and the 12-hour and 24-hour periodic kernels are higher than that of in the HUP subsets. We think this is because the higher sampling frequency in the MIMIC-III subset enables more accurate estimation of the short-term and periodic dependencies.

\subsubsection{Results for Online Imputation}
\begin{figure*}
\centering
\includegraphics[width=15.0cm]{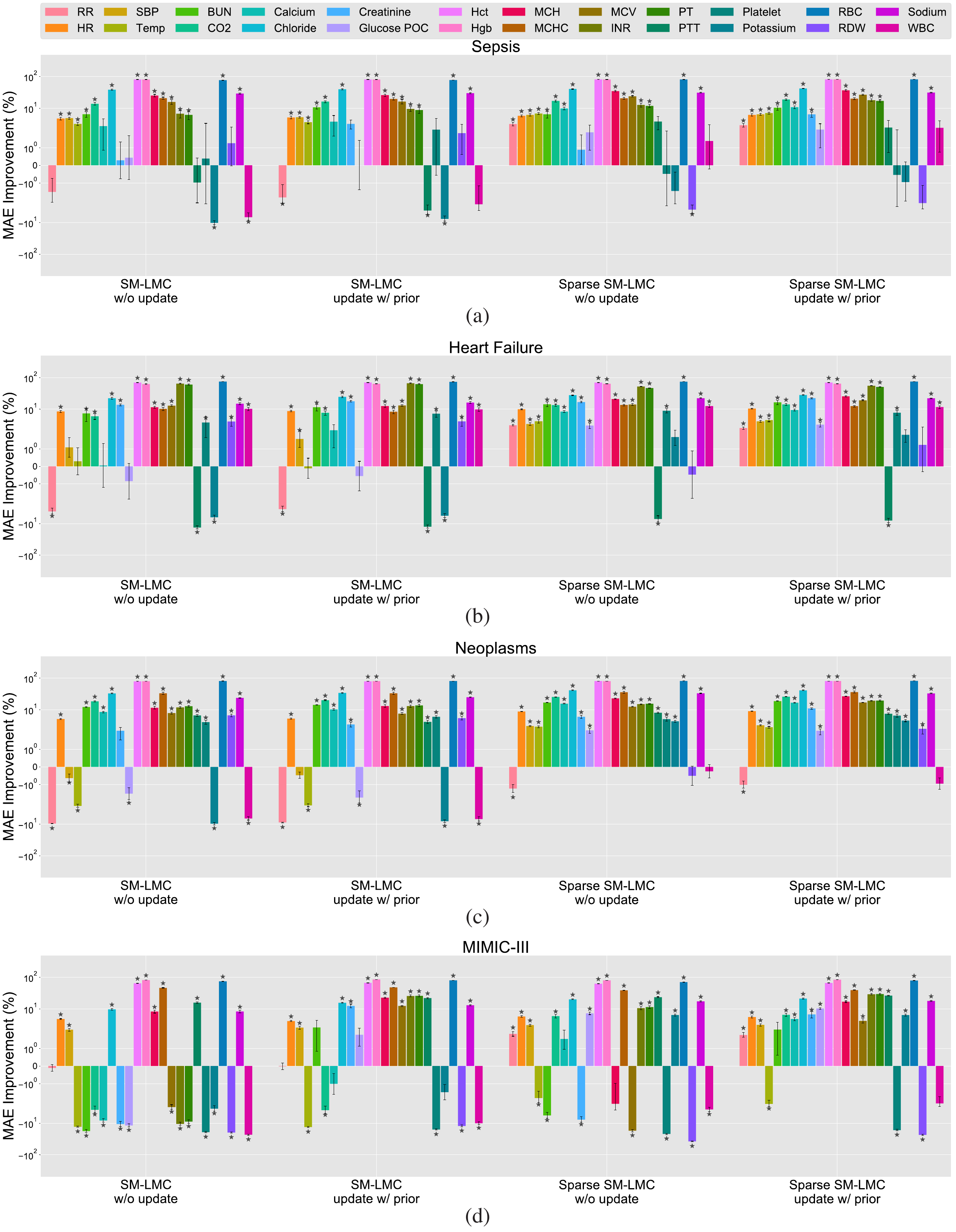}
\caption{\textbf{The percent improvement using MedGP for online imputation compared to independent (unvariate) GPs.} The figures depicts the results of 24 covariates for the (a) sepsis, (b) heart failure, and (c) neoplasms and (d) MIMIC-III heart failure subgroups. The $y$-axis is on log scale. The error bars denote $\pm 1$ standard error. The $\star$ indicates statistical significance.}
\label{fig:improvements_24cov_impute_vs_GP}
\end{figure*}

Next, we used the trained kernels to perform online imputation for each patient subgroup, where the goal is to predict the next observation for each covariate given the observations at previous time points. Across these methods, we used the percentage of improvement in MAE over three types of baselines---naive prediction, univariate GP (with SE or SM kernel), and PSM---to compare results for each of the 24 clinical covariates; we visualized the results separately (Figure~\ref{fig:improvements_24cov_impute_vs_GP}--\ref{fig:improvements_24cov_impute_vs_PSM}; Figure~\ref{fig:all_mae_p1}--\ref{fig:all_ci_p2} in Appendix C; Figure~\ref{fig:new_mae_p1}--\ref{fig:new_ci_p8} in Appendix D). We also show the results of variations of our method for comparison (with or without the proposed sparse prior; with or without online updating). We performed paired t-tests on predictions from MedGP and each baseline to quantify the improvements, and statistical significance was evaluated using Bonferroni-corrected $p < 4.17 \times 10^{-4}$.

\begin{figure*}
\centering
\includegraphics[width=15.0cm]{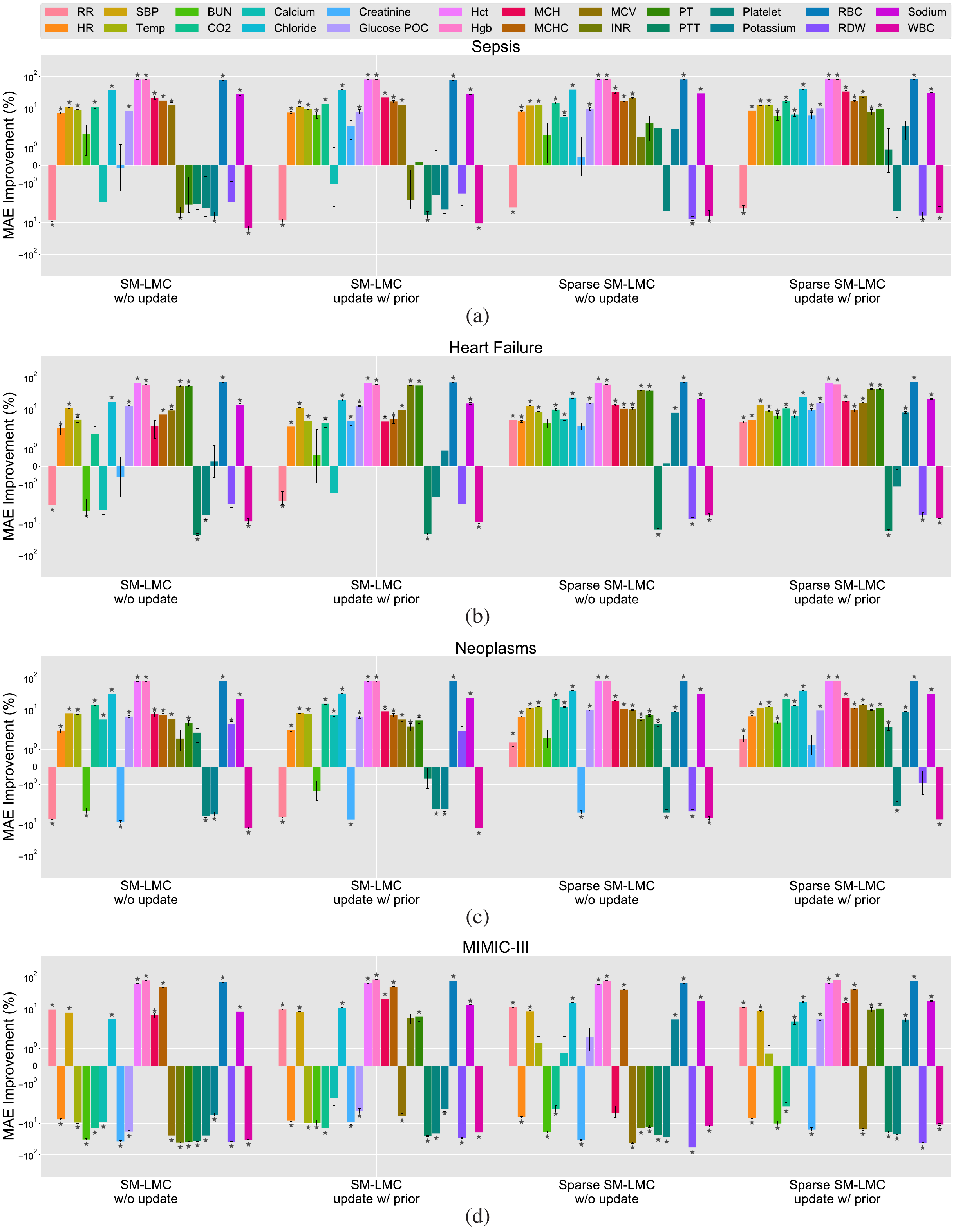}
\caption{\textbf{The percent improvement using MedGP for online imputation compared to the naive method.} The figures depicts the results of 24 covariates for the (a) sepsis, (b) heart failure, and (c) neoplasms and (d) MIMIC-III heart failure subgroups. The $y$-axis is on log scale. The error bars denote $\pm 1$ standard error. The $\star$ indicates statistical significance.}
\label{fig:improvements_24cov_impute_vs_Naive}
\end{figure*}

\begin{figure*}
\centering
\includegraphics[width=15.0cm]{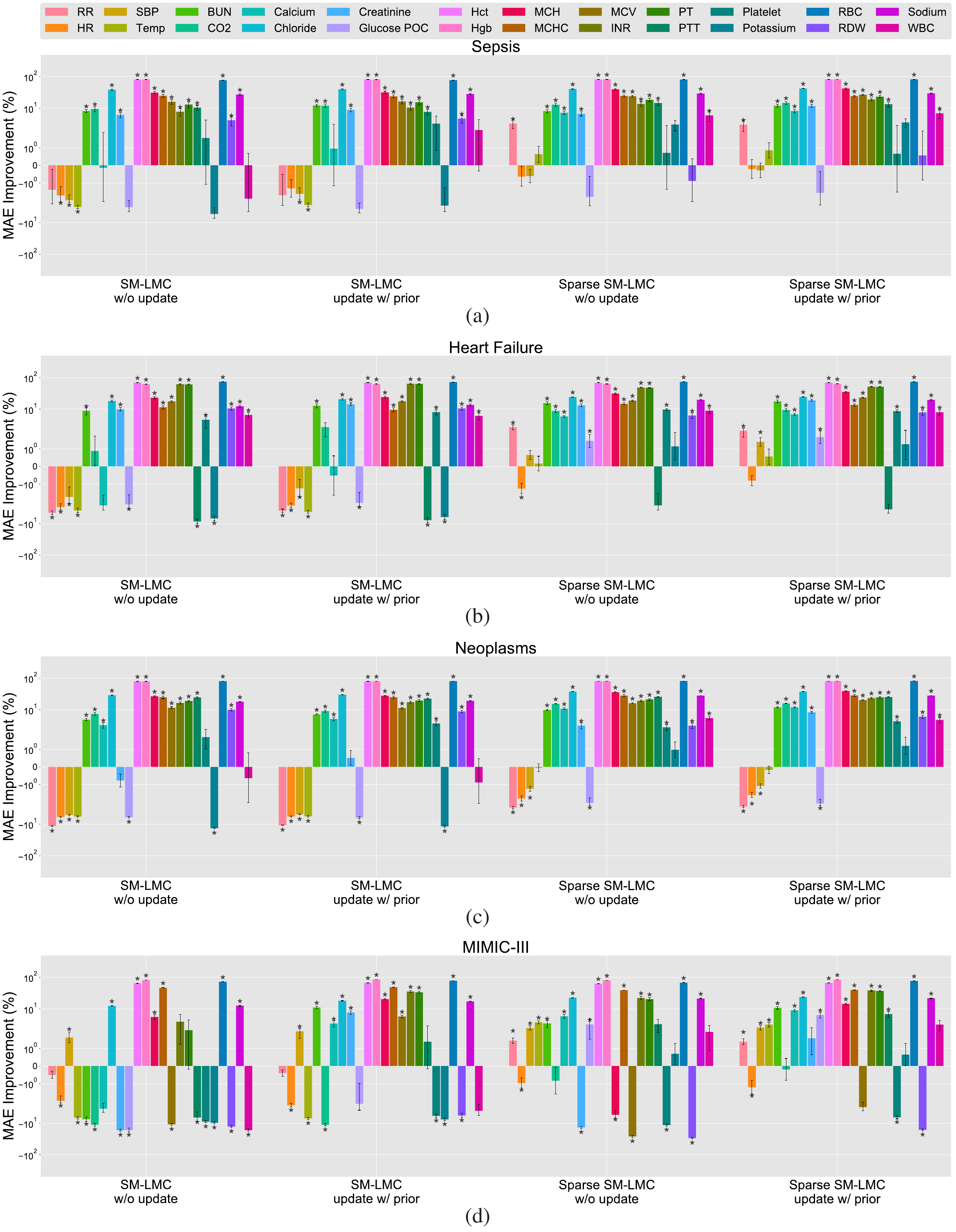}
\caption{\textbf{The percent improvement using MedGP for online imputation compared to PSM.} The figure depicts the results of 24 covariates for the (a) sepsis, (b) heart failure, and (c) neoplasms and (d) MIMIC-III heart failure subgroups. The $y$-axis is on log scale. The error bars denote $\pm 1$ standard error. The $\star$ indicates statistical significance.}
\label{fig:improvements_24cov_impute_vs_PSM}
\end{figure*}

Comparing results with the independent GP model---specifically, selecting the best results from the SE or SM kernel, we found that MedGP, and in particular sparse SM-LMC with online updating, outperformed the independent GP model on the online imputation task for most covariates across the four patient groups (Figure~\ref{fig:improvements_24cov_impute_vs_GP}). In the HUP data, we found 18, 21, 22 covariates significantly improved by MedGP in the sepsis, heart failure, and neoplasms subgroups respectively. In the MIMIC-III subset, we found 19 covariates were improved. For all four groups, the number of covariates that were improved significantly by MedGP is greater than using SM-LMC kernels without the sparse prior.
We found that the covariates that were well correlated in $\bm{B}_{q}$ usually showed significant positive improvements over independent GPs; Hct, Hgb, and RBC are notable examples. Similar observations could be made for INR and PT, the pair of lab covariates studied previously (Figure~\ref{fig:INR_PT_impute_online_one}). Across 24 covariates, the MAEs for INR and PT were slightly worse compared with only modeling these two covariates. However, we also observed that using the sparse prior with the SM-LMC kernel led to better performance as compared to not using the sparse prior, indicating that sparse regularization is helpful when jointly modeling heterogeneous covariates. Finally, there were some covariates for which MedGP did not improvement over univariate GPs in two or more disease groups, including red cell distribution width (RDW), white blood cell count (WBC) and platelets.

When the baseline method is the naive one-lag method, for all three disease groups, we found fewer covariates with significant improvements compared with improvements over univariate GPs (Figure~\ref{fig:improvements_24cov_impute_vs_Naive}). In particular, the covariates for which the naive method had an advantage were lab covariates that have piece-wise linear behavior, such as mean cell hemoglobin (MCH) and mean cell hemoglobin concentration (MCHC) (Figure~\ref{fig:time_series_example}). In the case of piece-wise linear behavior, our kernel does not improve the performance compared with the naive approach since the time series are neither smooth nor periodic.
Moreover, we also found that the naive method performed better in respiration rate, PTT, platelet, RDW, and white blood cell (WBC) count.
Overall, however, our method improved online prediction results for 18, 20, 20 of the 24 covariates in sepsis, heart failure, and neoplasms groups, respectively. In the MIMIC-III subset, we found 14 covariates were improved significantly over the naive method.

When the baseline method is PSM~\citep{suchi2015clustering}, we found that our method outperformed PSM for most of the lab covariates, but PSM outperformed MedGP in imputation of vital signs and two lab covariates: glucose point-of-care (Glucose POC) and potassium (Figure~\ref{fig:improvements_24cov_impute_vs_PSM}). 
For vital signs and glucose level, PSM has an advantage because of a higher sampling rate in those covariates and the highly structured mean function in PSM in the HUP subsets. The sampling rates are usually every 4 hours for vital signs and every 8 hours for glucose, which is more frequent than other lab covariates. 
Since PSM uses a B-spline basis function to capture the empirical mean, it may tolerate non-stationarity better. 
However, in the MIMIC-III subset, we observed that our method improved in imputing glucose and three vital signs (RR, SBP, temperature) over PSM significantly. We think this reflects the higher sampling rate of the covariates that allows better estimation of the short-term temporal dependencies. Overall, MedGP significantly improved the imputation of 17, 20, 18 covariates in sepsis, heart failure, neoplasms subsets respectively in the HUP data set, and 16 covariates in the MIMIC-III subset when compared with PSM. We contrast the PSM approach of structuring the mean function with our approach of structuring the kernel function, which leads to different types of gains in this problem.

Next, we looked at the calibration of the 95\% coverage (Figure~\ref{fig:all_ci_p1} and Figure~\ref{fig:all_ci_p2} in Appendix C; Figure~\ref{fig:new_ci_p1}--\ref{fig:new_ci_p8} in Appendix D). We found that MedGP outperformed independent GPs in terms of calibration of the 95\% confidence region for all covariates. For this evaluation, the values closer to 95\% are better. We observed that the coverage using the non-sparse SM-LMC kernel was usually higher than the coverage using the sparse SM-LMC kernel in the three HUP subgroups, indicating that MedGP may slightly underestimate covariate-specific noise. In contrast, in the MIMIC-III subset, we observed that MedGP gave consistently more accurate 95\% coverage than without regularization in most covariates. We also found that, in all patient subsets, online updating significantly improves the accuracy of the 95\% coverage.
Among all tested methods, PSM tended to overestimate the 95\% confidence region. We think this is because PSM assumes that the input time series are aligned by patient status, and this alignment is not the case in our data. With unaligned data, PSM learned large marginal variance parameters due to high empirical variance of the observations across patients at the same elapsed time. In contrast, the estimation of marginal covariance parameters in MedGP is not affected by alignment because estimates are patient-specific. We also observed that for either MedGP or PSM, the coverage was lower for some covariates in the MIMIC-III subset than in HUP subsets, such as temperature, CO$_{2}$, and PTT. This potentially reflects greater non-stationarity in the MIMIC-III subset, whose records were from intensive care units (ICUs) instead of regular hospital beds.

Finally, we compared the prediction performance of MedGP compared with the version without patient-specific online updating. We observed that online updating significantly improves the imputation errors of at least 12 out of 24 covariates in sepsis, heart failure, neoplasms, and the MIMIC-III subset (Figure~\ref{fig:all_mae_p1} and Figure~\ref{fig:all_mae_p2} in Appendix C; Figure~\ref{fig:new_mae_p1}--\ref{fig:new_mae_p8} in Appendix D). Similarly, evaluating the 95\% coverage, all 24 covariates were improved by the online updating across the three diseases groups in HUP, and 18 covariates were improved in the MIMIC-III subset (Figure~\ref{fig:all_ci_p1} and Figure~\ref{fig:all_ci_p2} in Appendix C; Figure~\ref{fig:new_ci_p1}--\ref{fig:new_ci_p8} in Appendix D). This improvement highlights the importance of updating the empirical priors with patient-specific observations for this problem.

\subsection{Computational Efficiency and Scalability}
\label{subsec:computation}
In this section, we compare computational speed between different implementations of our method. For patients with only a few observations, an existing implementation using conventional GP inference is sufficient for computationally tractable online inference. However, since our data include a large number of patients with potentially thousands of observations each, we implemented an exact inference algorithm in C++ and optimized it through Intel MKL libraries and customized multithreading blocks. In the experimental setting of $Q=5$, $D=24$, and $R_{q}=8$, there are 1114 hyperparameters to estimate. We summarized the runtime under different implementations for one patient with 2,028 unique time points and 6,679 observations (Table~\ref{tab:computation_time}); 
the tests were performed using a machine with Intel(R) Xeon(R) CPUs running at 2.40GHz. Using our optimized implementation, for patients with large number of observations ($T_{i} \geq 5000$), we accelerated training by a factor of 10 to 25 on average as compared with the sequential approach. We also compared our implementation with the standard GPy~\citep{gpy2014} implementation under different sample sizes and $Q$, and reached empirically at least three times speed up. We provide these results in Appendix E.

\begin{table}
\centering
\begin{tabular}{lcc}
\hline
Implementation & Sequential & Multithreading\\
\hline
Computing Gram matrix & 11 & 2 \\
Inverting Gram matrix & 13 & 3 \\
Computing gradients   & 2497 & 97 \\
Total per iteration   & 2521 & 102 \\
\hline
\end{tabular}
\caption{\textbf{Training time (in seconds) for a single iteration under different implementations of MedGP.} The total number of observations across time for this patient is 6,679. The sequential test used a single CPU, while the multithreading test used 35 CPUs---one thread per CPU.}
\label{tab:computation_time}
\end{table}

The proposed framework can be parallelized at the patient level and is suitable for analysis when patient data are observed in a streaming form. For each reference patient, we distributed the optimized training process on a computing cluster to estimate the patient-specific hyperparameters in parallel. In addition, the population-level kernels could be updated sequentially; the computationally expensive GP training procedure does not need to be applied to patient data in bulk. That is, when we receive more data from new patients, we only need to update the kernel density estimators. 
Our framework provides better computational efficiency compared to models designed for smaller collections of observations (e.g., approximately two hundred observations for each patient) as in most previous work. Those approaches are computationally intractable when working on a set of rich patient observations of the magnitude of the HUP data due to large matrix inversions and summing marginal likelihoods across patients at each iteration.

%% file: Discussion.tex
\section{Discussion}
\label{sec:Discussion}
In this paper, we propose a flexible and efficient framework for estimating the temporal dependencies across multiple sparse and irregularly sampled medical time series data. We developed a model with multi-output Gaussian process regression with a highly structured kernel. We fit this model using an optimized implementation of exact GP inference to three different disease groups in the HUP medical data set and the MIMIC-III ICU data set. We showed that our method, MedGP, improves performance for online prediction of 24 clinical covariates as compared with independent univariate GPs, a naive method of propagating the previous observation, and an earlier state-of-the-art approach, PSM~\citep{suchi2015clustering}. We found that, for well-correlated covariates, our method improves online imputation performance substantially over the related methods in most tested covariates. The improvements over the naive one-lag prediction and univariate GPs were significant in both vital signs and lab covariates. We found that PSM was, in general, better at predicting vital signs with more densely sampled observations. However, our approach does not require patient time series alignment and shows better calibration of the 95\% confidence region as compared to PSM.

There are several directions that will be explored using the MedGP framework motivated by the present results. The first direction is to allow time-varying covariances by specifically modeling non-stationarity. 
Some possible approaches to explore include incorporating state-space models or change point detection~\citep{adams2007bayesian,saatcci2010gaussian}, and extending those methods to work on multivariate scenarios. 
Another direction of interest is to consider latent subpopulation-level structured kernels through multivariate medical time series. We expect that our results could be further improved through incorporating hierarchical methods with proper features or metrics to represent the differences between patients within the same disease group and across disease groups more carefully. For instance, the original PSM used three levels of hierarchy based on the subpopulations of patients with scleroderma, including population level, subpopulation level, and individual level. Our model may benefit from such an approach, but more efficient inference procedures are needed to train on our large data set~\citep{feinberg2017large}. We should point out that this is possible through, for instance, deriving corresponding stochastic variation inference (SVI) algorithm. For example, previous work develops the SVI algorithm for semiparametric latent factor model (SLFM) with $R_{q}=1$~\citep{nguyen2014collaborative}, which could be generalized to apply to MedGP.

For future applications, we will use the framework to monitor the health status of patients in a hospital setting and identify those patients at high risk for acute diseases in order to assist with decision making in treatment plans. Specifically, MedGP can impute latent state in patients at any time point, including confidence region around those estimates; this latent state can be used for a number of downstream analyses which require complete knowledge of patient state at specific time points. For instance, the changes of dynamics and temporal correlations between two vital signs have been found to be useful for disease detection given high-frequency regularly sampled time series~\citep{discover-cardio-dynamics-2012,dynamics-based-patient-monitoring-2015}. We demonstrated that MedGP accurately estimates the temporal correlations in the presence of sparse, unaligned time series data for up to 24 covariates, and we would expect to further associate the cross-covariate dynamics to more complicated diseases, such as septic shock~\citep{Henry299ra122}, where the interactions of multiple covariates are jointly taken into consideration for diagnosis.

%% file: AppendixA.tex
\newpage
\appendix
\section*{Appendix A. Details of the Hierarchical Gamma Prior}

In this appendix, we provide more background and visualization for the hierarchical gamma prior we used for regularization. For the convenience, we use $\Gamma(\cdot)$ to denote the gamma function, and $\mathcal{G}(a, b)$ to represent a gamma distribution with shape parameter $a$ and rate parameter $b$.

Following Proposition 1 in~\cite{armagan2011}, for a random variable $x$ drawn from a normal distribution with two-layered gamma priors on variance
\begin{equation}
x \sim \mathcal{N}(0, \psi_{1}), \quad
\psi_{1} \sim \mathcal{G}(\alpha, \delta), \quad
\delta \sim \mathcal{G}(\beta, \nu),
\end{equation}
is equivalent to the hierararchy
\begin{equation}\label{equ:TPBN}
x \sim \mathcal{N}(0, \nicefrac{1}{\rho}-1), \quad \rho \sim \mathcal{TPB}(\alpha, \beta, \nu),
\end{equation}
where $\mathcal{TPB}(\alpha, \beta, \nu)$ denotes the three-parameter beta distribution. The probability density function of $\rho$ is given as
\begin{equation}
f(\rho; \alpha, \beta, \nu) = \displaystyle\frac{\Gamma(\alpha+\beta)}{\Gamma(\alpha)\Gamma(\beta)}\nu^{\beta}\rho^{\beta-1}(1-\rho)^{\alpha-1}[1 + (\nu-1)\rho]^{-(\alpha+\beta)}.
\end{equation}

\begin{figure*}[h]
 \centering
    \includegraphics[width=10cm]{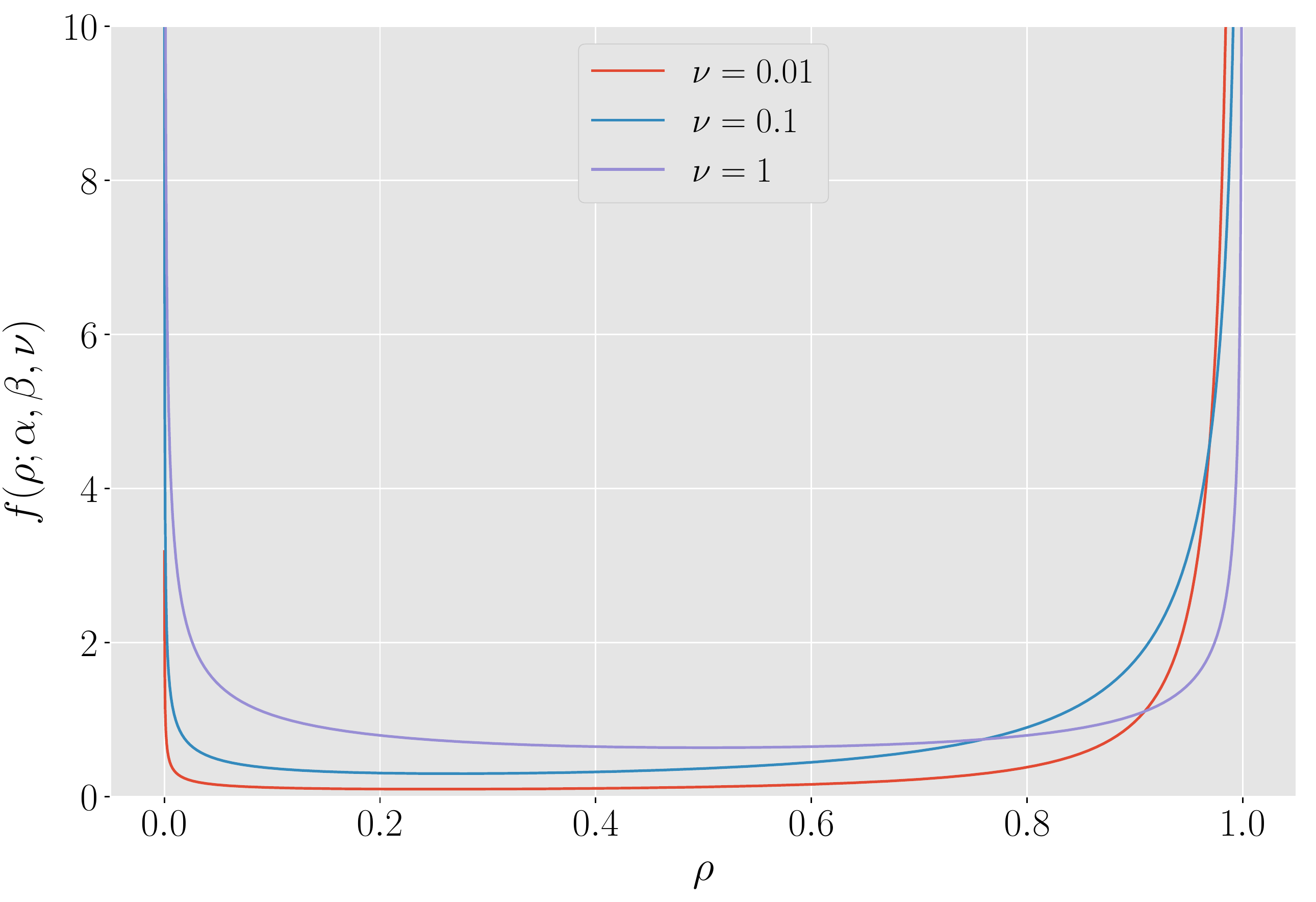}
    \caption{{\bf The density of $\rho$ drawn from a three parameter beta prior with different values of $\nu$.} For all values of $\nu$, we set $\alpha=\beta=0.5$.}
    \label{fig:tpb_density}
\end{figure*}

In Figure~\ref{fig:tpb_density}, we visualized the density of $\rho$ in Equation (\ref{equ:TPBN}) for $\alpha=\beta=0.5$, and under different values of $\nu$~\citep{armagan2011}. In this case, the prior distribution of $x$ is equivalent to a horseshoe prior, and $\rho$ can be interpreted as the shrinkage coefficient~\citep{Carvalho2010}. 

Specifically, for the case with four layers of gamma prior used in our work,
\[
x \sim \mathcal{N}(0, \psi_{2}), \quad
\psi_{2} \sim \mathcal{G}(\alpha, \delta), \quad
\delta \sim \mathcal{G}(\beta, \phi), \quad
\phi \sim \mathcal{G}(\gamma, \tau), \quad
\tau \sim \mathcal{G}(d, \eta),
\]
is equivalent to
\[
x \sim \mathcal{N}(0, \nicefrac{1}{\rho}-1), \quad \rho \sim \mathcal{TPB}(\alpha, \beta, \nicefrac{1}{\zeta}-1), \quad \zeta \sim \mathcal{TPB}(\gamma, d, \eta).
\]
In our case, we set $\alpha = \beta = \gamma = d = 0.5$ so both $\rho$ and $\zeta$ recapitulate horseshoe priors~\citep{armagan2011,Gao2014a,zhao2014bayesian-JMLR}.

%% file: AppendixB.tex
\newpage
\appendix
\section*{Appendix B. Details of Gradient Computation and Update Equations}

In this appendix, the equations for the objective function during optimization, update equations for the parameters in the sparsity inducing prior and the gradients for the hyperparameters of the GP kernel are listed as reference.

The objective function to optimize for training one patient, $\mathcal{Q}(\bm{\theta})$, is 
\begin{equation}
\label{equation:Q}
\begin{array}{ll}
\mathcal{Q}(\bm{\theta}) 
\propto& \displaystyle{\left[ -\frac{1}{2}\mathbf{y}^{\top}(K_{|\bm{\theta}}+\bm{\epsilon}I)^{-1}\mathbf{y} - \frac{1}{2}\log{|K_{|\bm{\theta}}+\mathbf{\epsilon}I|} - \left(\frac{\sum_{d=1}^{D}T_{i,d}}{2}\right)\log{2\pi}
\right]} \\
&+ \displaystyle{
\sum_{q=1}^{Q}\sum_{d=1}^{D}\sum_{r=1}^{R_{q}}
\left(
-\frac{1}{2}\log{\psi_{q,(d,r)}}-\frac{a_{q,(d,r)}^{2}}{2\psi_{q,(d,r)}}
\right) } \\
&+ \displaystyle{
\sum_{q=1}^{Q}\sum_{d=1}^{D}\sum_{r=1}^{R_{q}}
\left[
\alpha\log{\delta_{q,(d,r)}} + (\alpha-1)\log{\psi_{q,(d,r)}}-\delta_{q,(d,r)}\psi_{q,(d,r)}
\right] } \\
&+ \displaystyle{
\sum_{q=1}^{Q}\sum_{d=1}^{D}\sum_{r=1}^{R_{q}}
\left[
\beta\log{\phi_{q,(r)}} + (\beta-1)\log{\delta_{q,(d,r)}}-\phi_{q,(r)}\delta_{q,(d,r)}
\right] } \\
&+ \displaystyle{
\sum_{q=1}^{Q}\sum_{r=1}^{R_{q}}
\left[
\gamma\log{\tau_{q,(r)}} + (\gamma-1)\log{\phi_{q,(r)}}-\tau_{q,(r)}\phi_{q,(r)}
\right] } \\
&+ \displaystyle{
\sum_{q=1}^{Q}\sum_{r=1}^{R_{q}}
\left(
d\log{\eta} + (d-1)\log{\tau_{q,(r)}}-\eta\tau_{q,(r)}
\right) } \\
&+ \displaystyle{
\sum_{q=1}^{Q}\sum_{d=1}^{D}\left( 
-\log{2\beta_{\lambda}}-\frac{|\lambda_{q,(d)}|}{\beta_{\lambda}}
\right)
}.\\
\end{array}
\end{equation}

For update equations, we quoted from \cite{zhao2014bayesian-JMLR}:
\begin{equation}
\label{equation:prior_update_start}
\hat{\psi}_{q,(d,r)} = \displaystyle{ \frac{(2\alpha-3)+\sqrt{(2\alpha-3)^{2} + 8 a_{q,(d,r)}^{2} \delta_{q,(d,r)}}}{4\delta_{q,(d,r)}} }
\end{equation}

\begin{equation}
\hat{\delta}_{q,(d,r)} = \displaystyle \frac{\alpha + \beta}{\psi_{q,(d,r)} + \phi_{q,(r)}}
\end{equation}

\begin{equation}
\hat{\phi}_{q,(r)} = \displaystyle{ \frac{D\beta+\gamma-1}{\sum_{d=1}^{D}\delta_{q,(d,r)}+\tau_{q,(r)}} }
\end{equation}

\begin{equation}
\label{equation:prior_update_end}
\hat{\tau}_{q,(r)} = \displaystyle \frac{\gamma + d}{\phi_{q,(r)}+\eta}
\end{equation}

\begin{equation}
\label{equation:GP_gradient_start}
\frac{\partial}{\partial \theta_{j}}
\log{p(\textbf{y}|
\textbf{x}, \bm{\theta})} 
= \frac{1}{2}\text{tr}\left( 
\left( \bm{\alpha\alpha}^{\top} - K^{-1}_{|\bm{\theta}}\right)\frac{\partial K_{|\bm{\theta}}}{\partial \theta_{j}}
\right) 
\quad \text{where } \bm{\alpha}=K^{-1}_{|\bm{\theta}}\mathbf{y}, {\theta_{j}} \in \bm{\theta}
\end{equation}

\begin{equation}
\begin{array}{rl}
\displaystyle \frac{\partial \mathcal{Q}(\bm{\theta})}{\partial a_{q, (d, r)}} &= \displaystyle \frac{1}{2}\text{tr}\left( 
\left( \bm{\alpha\alpha}^{\top} - K^{-1}_{|\bm{\theta}}\right)\frac{\partial K_{|\bm{\theta}}}{\partial a_{q,(d,r)}}
\right) - \frac{a_{q,(d,r)}}{\psi_{q,(d,r)}},\\\\
\text{where} \quad \displaystyle \frac{\partial K_{|\bm{\theta}}}{\partial a_{q, (d, r)}} &=
B_{q}' \otimes k_{q}(\mathbf{x}, \mathbf{x}'),\\
B_{q, (i, j)}' &= \left \lbrace 
\begin{array}{ll}
2 a_{q, (d, r)} &, \text{for } i=j=d,\\
a_{q, (j, r)} &, \text{for } i=d, j \neq d,\\
a_{q, (i, r)} &, \text{for } i \neq d, j=d,\\
0 &,  \text{otherwise}.\\
\end{array}
\right.
\end{array}
\end{equation}

For partial gradients used for optimization:
\begin{equation}
\begin{array}{rl}
\displaystyle \frac{\partial \mathcal{Q}(\bm{\theta})}{\partial \lambda_{q, (d)}} &= \displaystyle
\frac{1}{2}\text{tr}\left( 
\left( \bm{\alpha\alpha}^{\top} - K^{-1}_{|\bm{\theta}}\right)\frac{\partial K_{|\bm{\theta}}}{\partial \lambda_{q, (d)}}
\right) 
- \frac{\text{sign}(\lambda_{q, (d)})}{\beta_{\lambda}},\\\\
\text{where} \quad \displaystyle \frac{\partial K_{|\bm{\theta}}}{\partial \lambda_{q, (d)}} &= \text{diag}\left(\bm{\lambda}_{q}'\right) \otimes k_{q}(\mathbf{x}, \mathbf{x}'),\\
\lambda_{q, (i)}' &= \left \lbrace
\begin{array}{ll}
1 &, \text{for } i = d,\\
0 &, \text{otherwise}.
\end{array}
\right.
\end{array}
\end{equation}

\begin{equation}
\begin{array}{rl}
\displaystyle \frac{\partial \mathcal{Q}(\bm{\theta})}{\partial v_{q}} 
&= \displaystyle
\frac{1}{2}\text{tr}\left( 
\left( \bm{\alpha\alpha}^{\top} - K^{-1}_{|\bm{\theta}}\right)\frac{\partial K_{|\bm{\theta}}}{\partial v_{q}}
\right),\\\\
\text{where} \quad \displaystyle \frac{\partial K_{|\bm{\theta}}}{\partial v_{q}} &= B_{q} \otimes k_{qv}(\mathbf{x}, \mathbf{x}'),\\\\
k_{qv}(\mathbf{x}, \mathbf{x}') &= -2\pi^{2}\tau^{2}\exp(-2\pi^{2}\tau^{2}v_{q})\cos(2\pi\tau\mu_{q}).
\end{array}
\end{equation}

\begin{equation}
\label{equation:GP_gradient_end}
\begin{array}{rl}
\displaystyle \frac{\partial \mathcal{Q}(\bm{\theta})}{\partial \mu_{q}} 
&= \displaystyle
\frac{1}{2}\text{tr}\left( 
\left( \bm{\alpha\alpha}^{\top} - K^{-1}_{|\bm{\theta}}\right)\frac{\partial K_{|\bm{\theta}}}{\partial \mu_{q}}
\right),\\\\
\text{where} \quad \displaystyle \frac{\partial K_{|\bm{\theta}}}{\partial \mu_{q}} &= B_{q} \otimes k_{q\mu}(\mathbf{x}, \mathbf{x}'),\\\\
k_{q\mu}(\mathbf{x}, \mathbf{x}') &= -2\pi\tau \exp{(-2\pi^{2}\tau^{2} v_{q})}\sin{(2\pi\tau\mu_{q})}.
\end{array}
\end{equation}

%% file: AppendixC.tex
\newpage
\appendix
\section*{Appendix C. Detailed Results of Imputation Error and 95\% Coverage}

\begin{figure*}
\centering
\includegraphics[width=15.5cm]{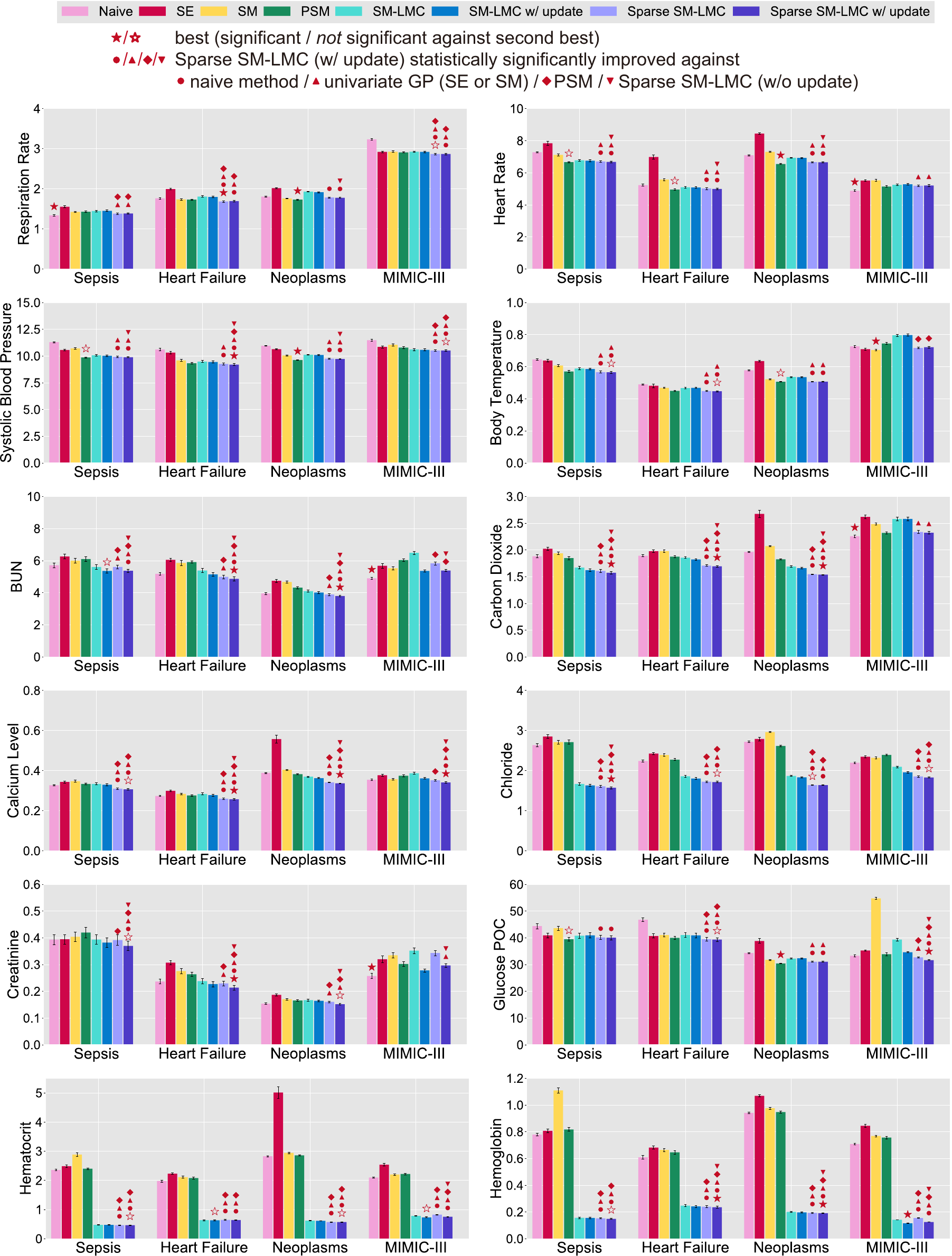}
\caption{\textbf{Mean absolute error (MAE) for 12 out of 24 covariates tested.} The error bars denote $\pm 1$ standard error.}
\label{fig:all_mae_p1}
\end{figure*}

\begin{figure*}
\centering
\includegraphics[width=15.5cm]{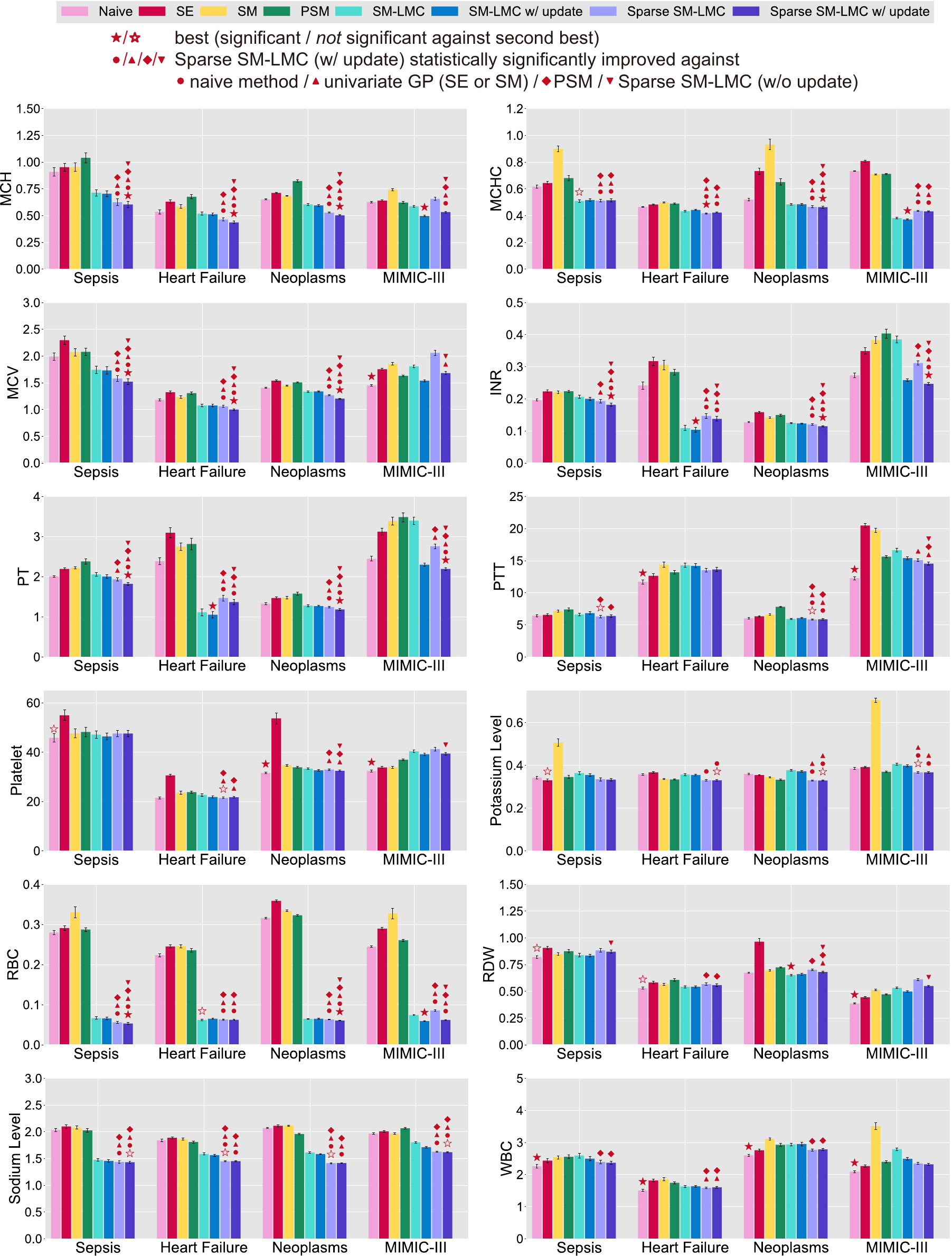}
\caption{\textbf{Mean absolute error (MAE) for 12 out of 24 covariates tested.} The error bars denote $\pm 1$ standard error.}
\label{fig:all_mae_p2}
\end{figure*}

\begin{figure*}
\centering
\includegraphics[width=15.5cm]{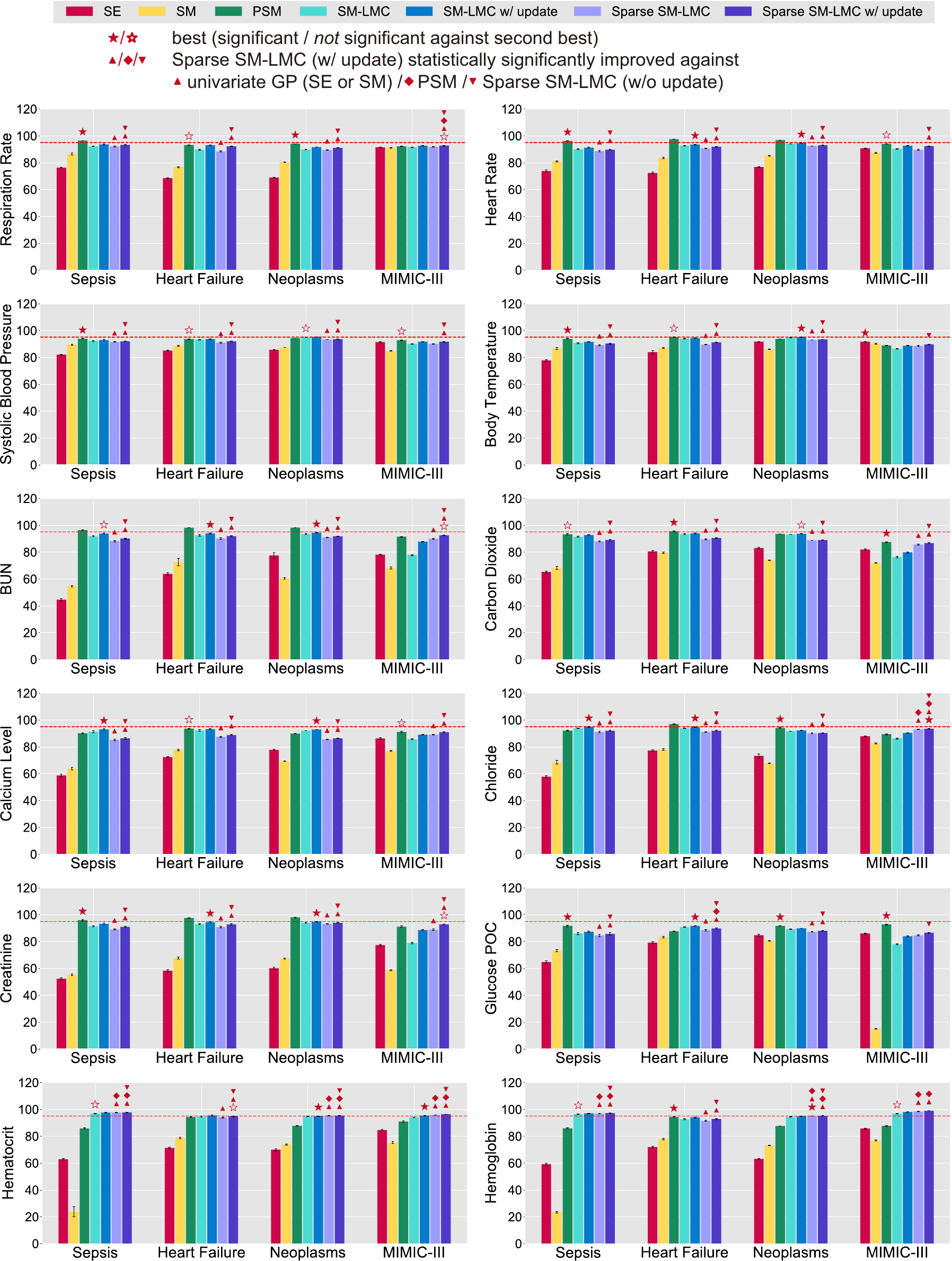}
\caption{\textbf{The 95\% coverage for 12 out of 24 covariates tested.} The error bars denote $\pm 1$ standard error. The red dashed line indicates 95\%.}
\label{fig:all_ci_p1}
\end{figure*}

\begin{figure*}
\centering
\includegraphics[width=15.5cm]{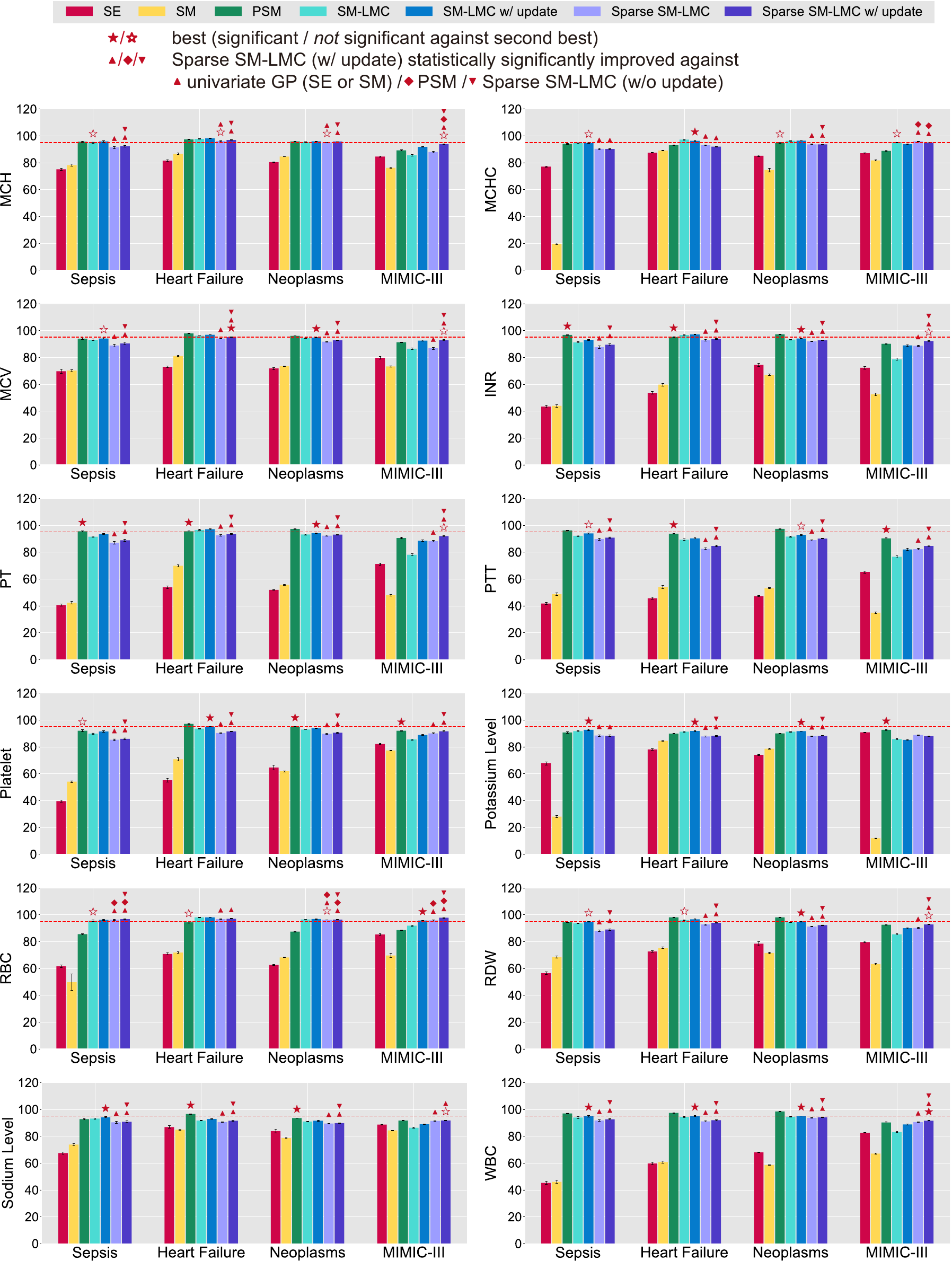}
\caption{\textbf{The 95\% coverage for 12 out of 24 covariates tested.} The error bars denote $\pm 1$ standard error. The red dashed line indicates 95\%.}
\label{fig:all_ci_p2}
\end{figure*}

We organized the detailed results of online imputation on all 24 covariates under the best number of basis kernel ($Q=5$ for HUP subsets and $Q=4$ for the MIMIC-III subset) in Figure~\ref{fig:all_mae_p1} to Figure~\ref{fig:all_ci_p2}. For Figure~\ref{fig:all_mae_p1} and Figure~\ref{fig:all_mae_p2}, the mean absolute errors (MAEs) for each covariate is shown (in the original unit of measure). In Figure~\ref{fig:all_ci_p1} and Figure~\ref{fig:all_ci_p2}, we showed the percentage for the prediction lied within the 95\% confidence region (i.e. 95\% coverage). We put markers in the figures to indicate the best among all methods, and the comparison of MedGP (sparse SM-LMC with online updating) against other methods. The statistical significance were tested using paired t-tests on patient-level results.

%% file: AppendixD.tex
\newpage
\appendix
\section*{Appendix D. Results under Different Number of Basis Kernels}

\begin{figure*}
\centering
\subfigure{
\includegraphics[width=16.0cm]{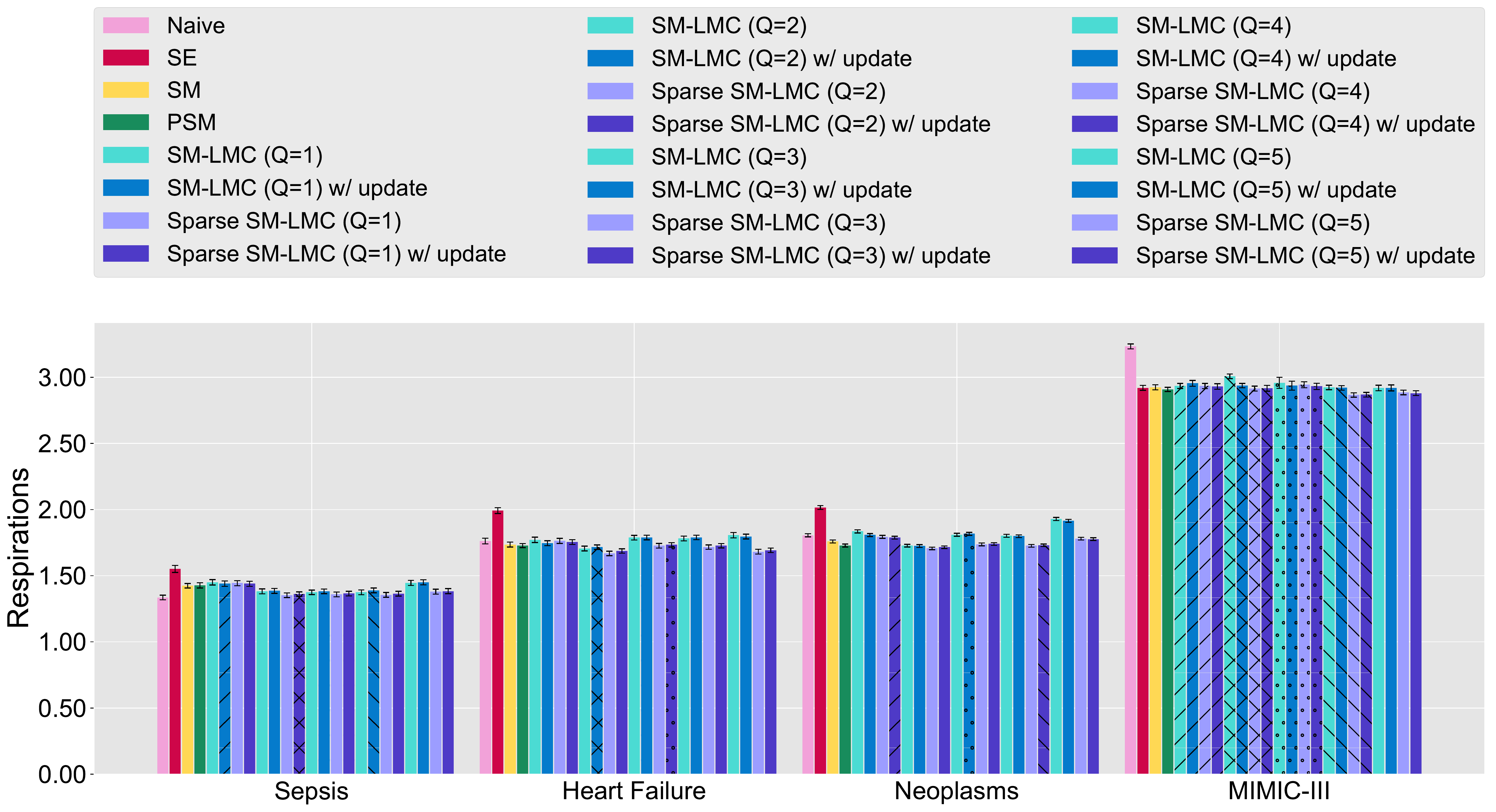}
}
\subfigure{
\includegraphics[width=16.0cm]{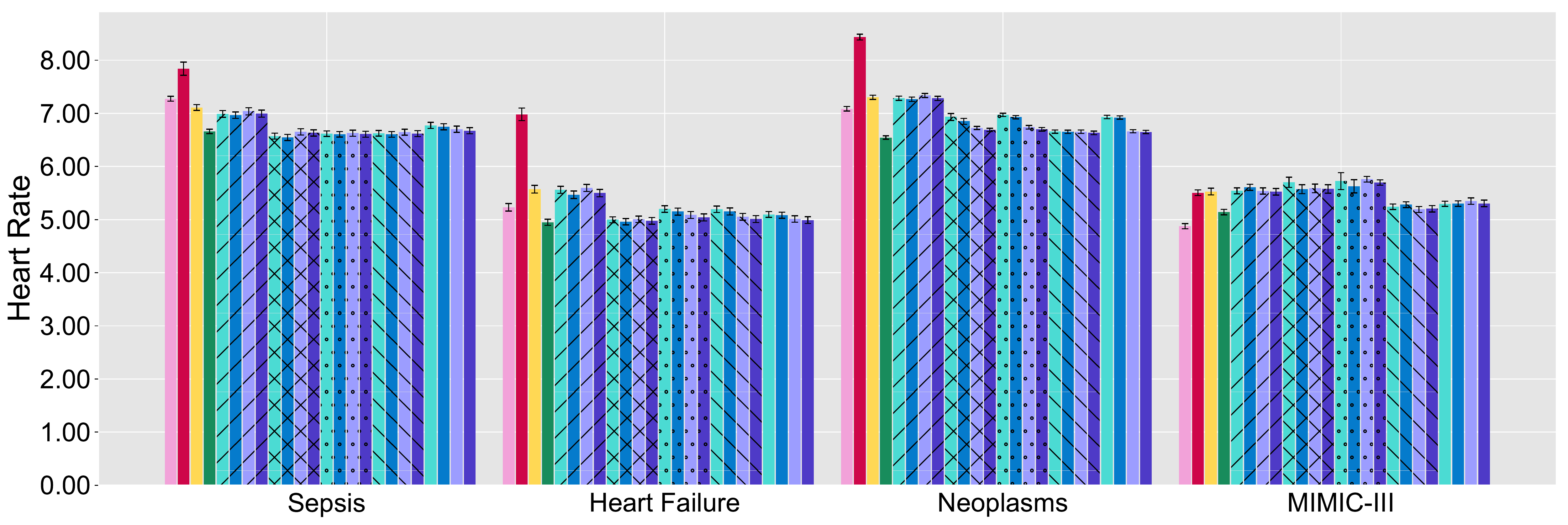}
}
\subfigure{
\includegraphics[width=16.0cm]{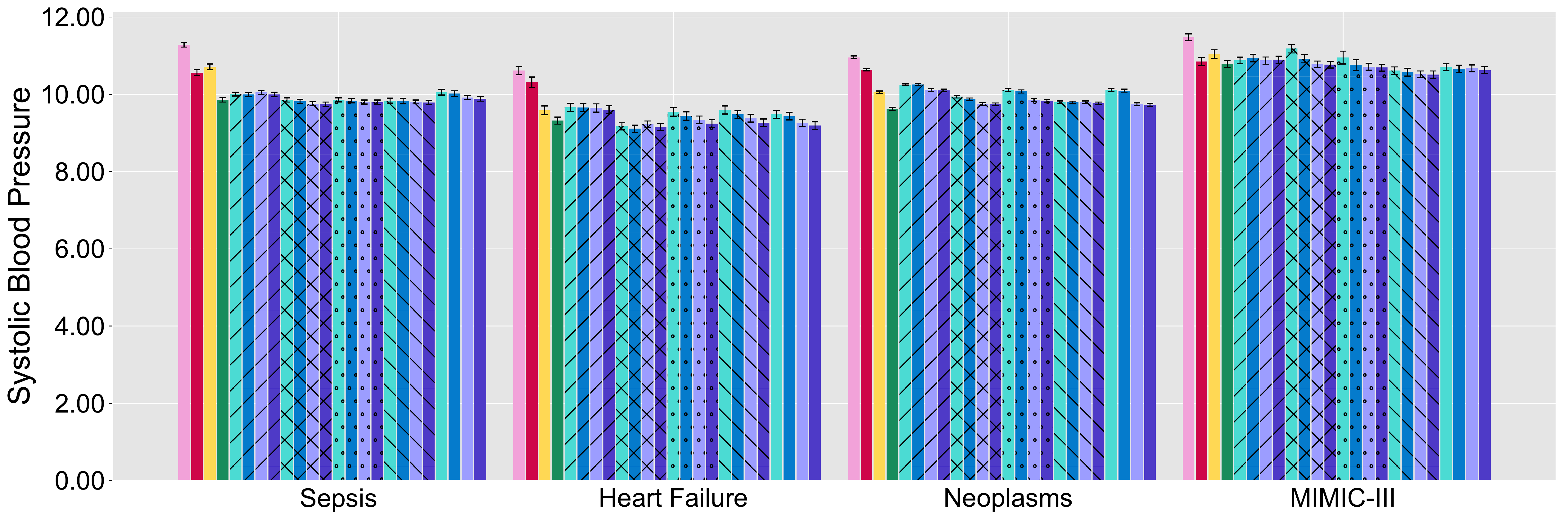}
}
\caption{\textbf{The mean absolute error (MAE) of online imputation under different $Q$ for all cohorts.} The error bars denote $\pm 1$ standard error.}
\label{fig:new_mae_p1}
\end{figure*}

\begin{figure*}
\centering
\subfigure{
\includegraphics[width=16.0cm]{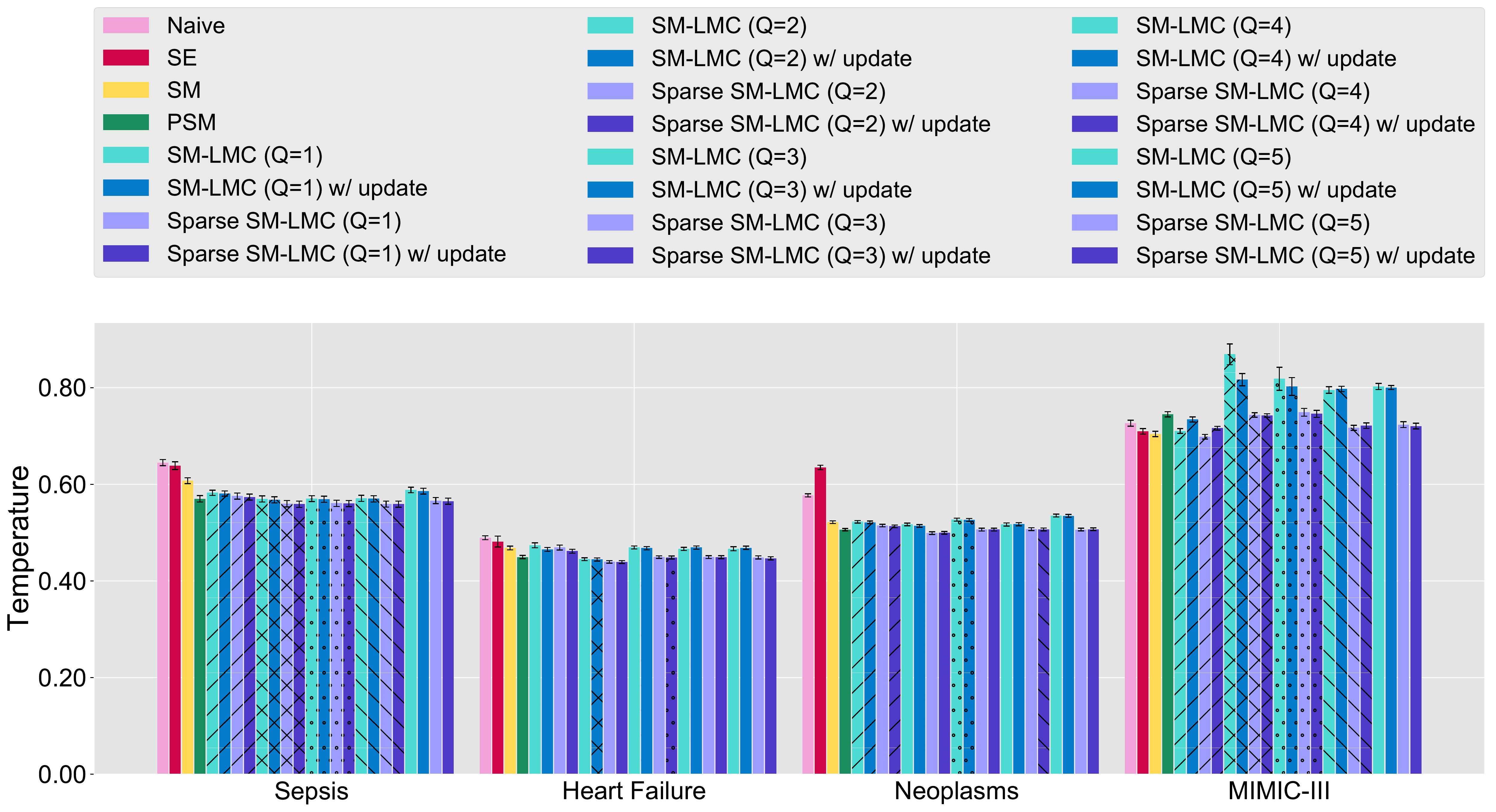}
}
\subfigure{
\includegraphics[width=16.0cm]{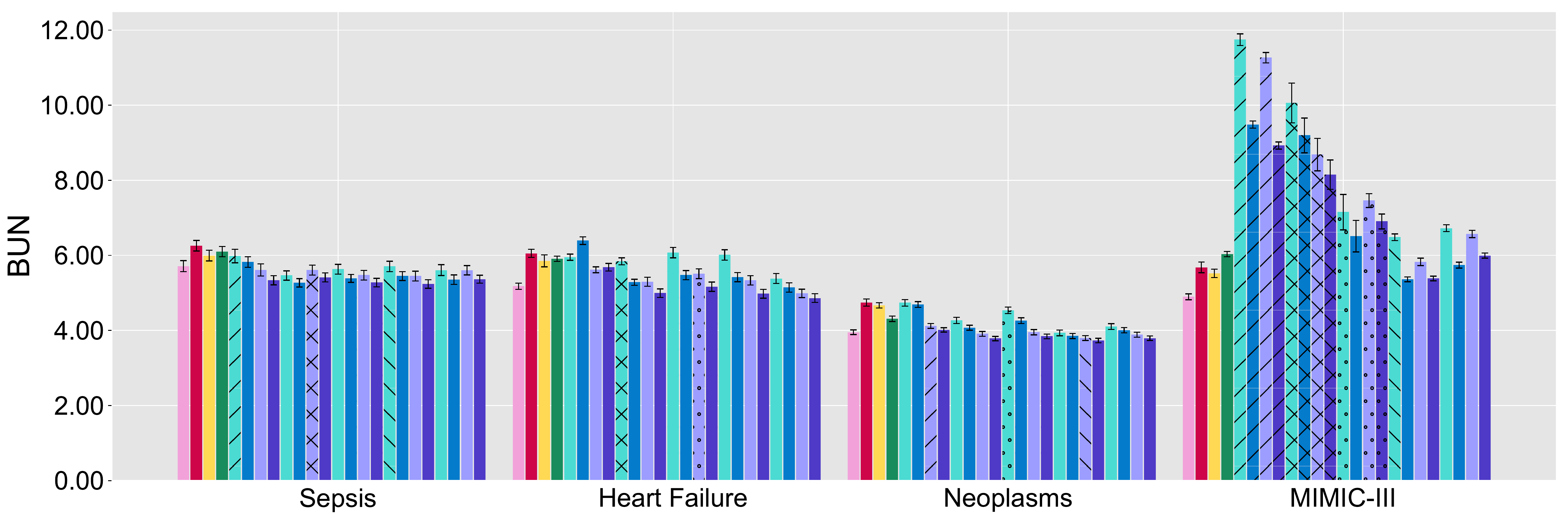}
}
\subfigure{
\includegraphics[width=16.0cm]{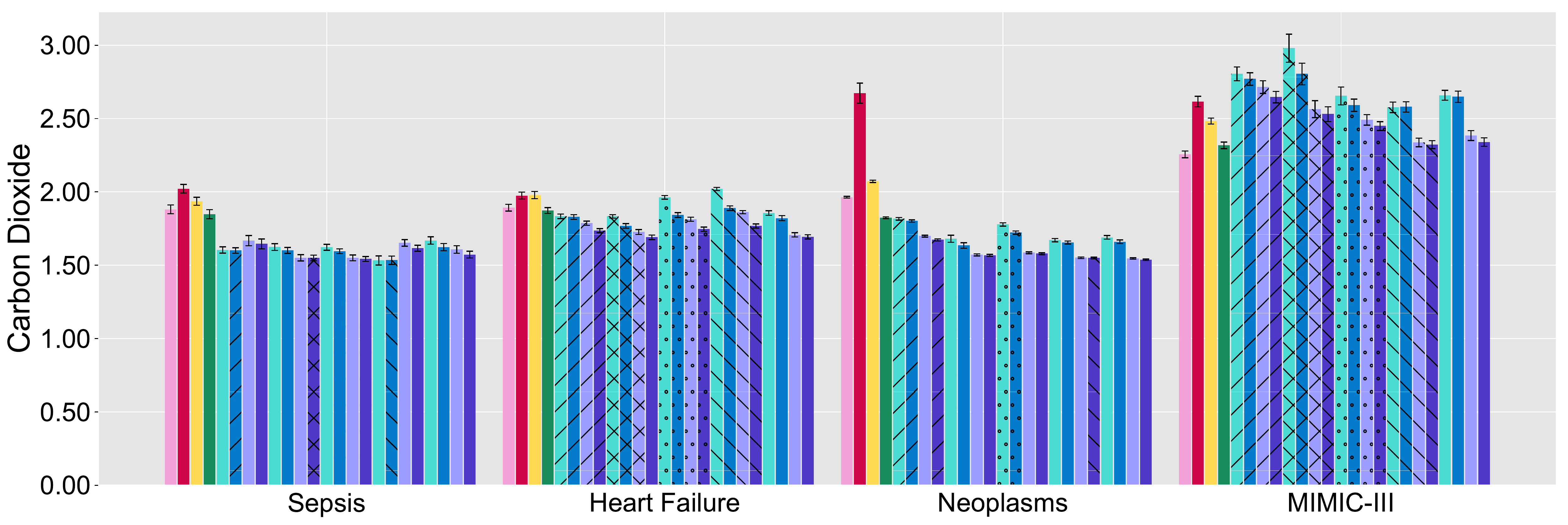}
}
\caption{\textbf{The mean absolute error (MAE) of online imputation under different $Q$ for all cohorts.} The error bars denote $\pm 1$ standard error.}
\label{fig:new_mae_p2}
\end{figure*}

\begin{figure*}
\centering
\subfigure{
\includegraphics[width=16.0cm]{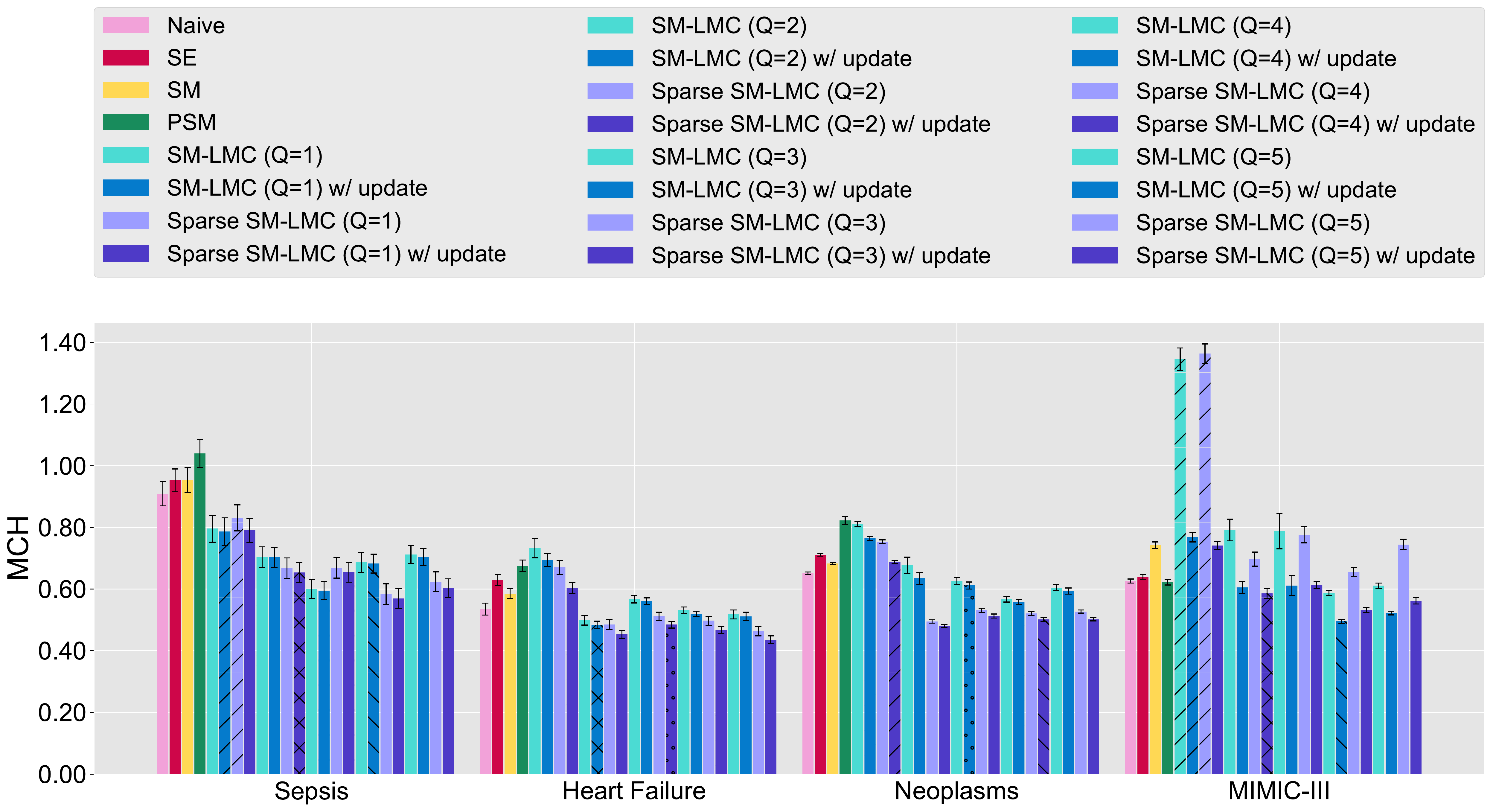}
}
\subfigure{
\includegraphics[width=16.0cm]{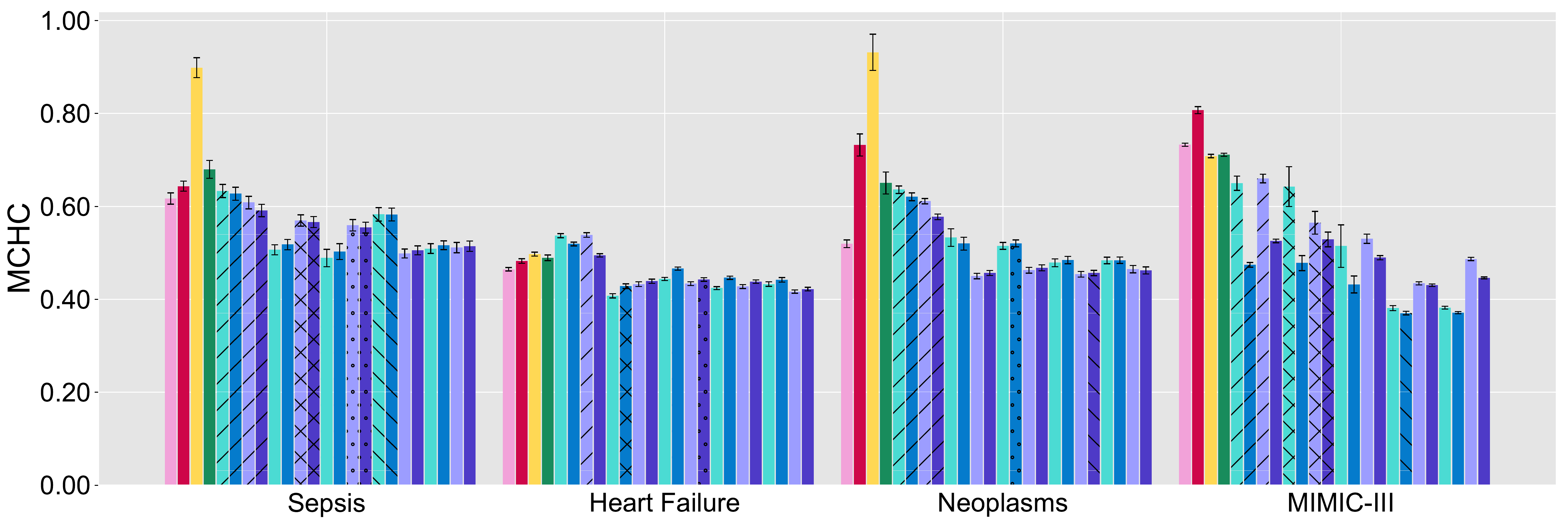}
}
\subfigure{
\includegraphics[width=16.0cm]{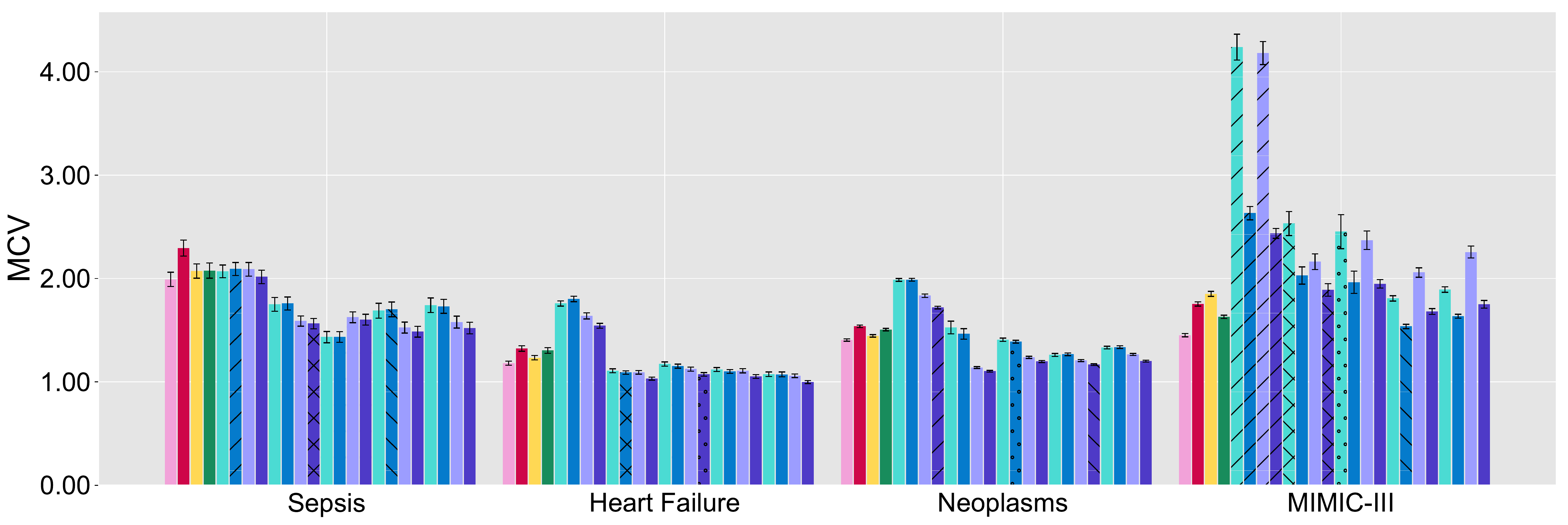}
}
\caption{\textbf{The mean absolute error (MAE) of online imputation under different $Q$ for all cohorts.} The error bars denote $\pm 1$ standard error.}
\label{fig:new_mae_p5}
\end{figure*}

\begin{figure*}
\centering
\subfigure{
\includegraphics[width=16.0cm]{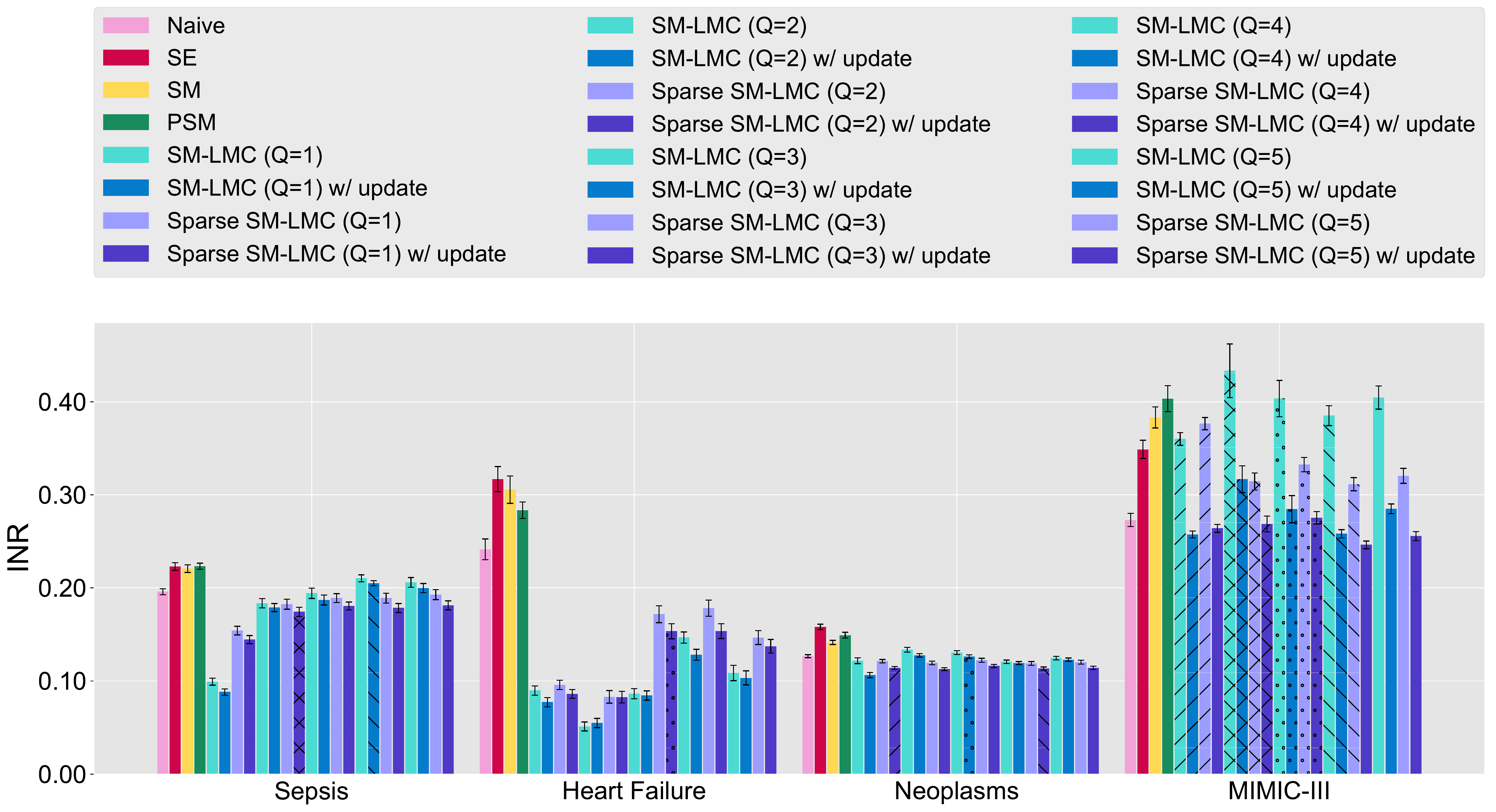}
}
\subfigure{
\includegraphics[width=16.0cm]{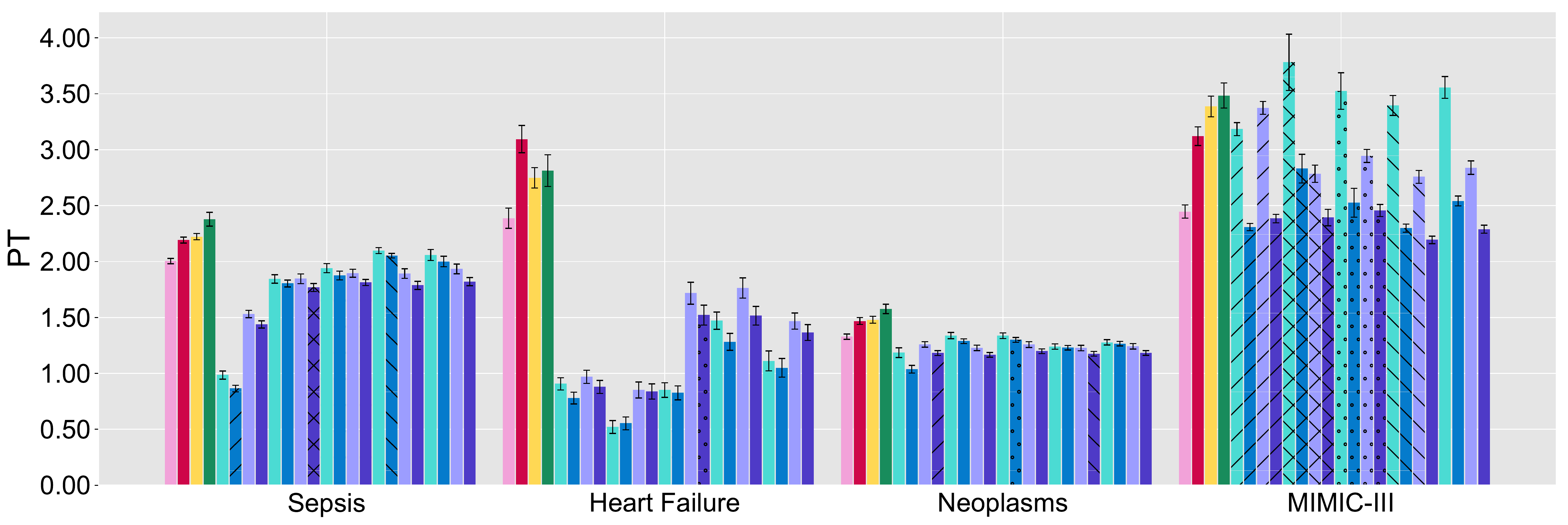}
}
\subfigure{
\includegraphics[width=16.0cm]{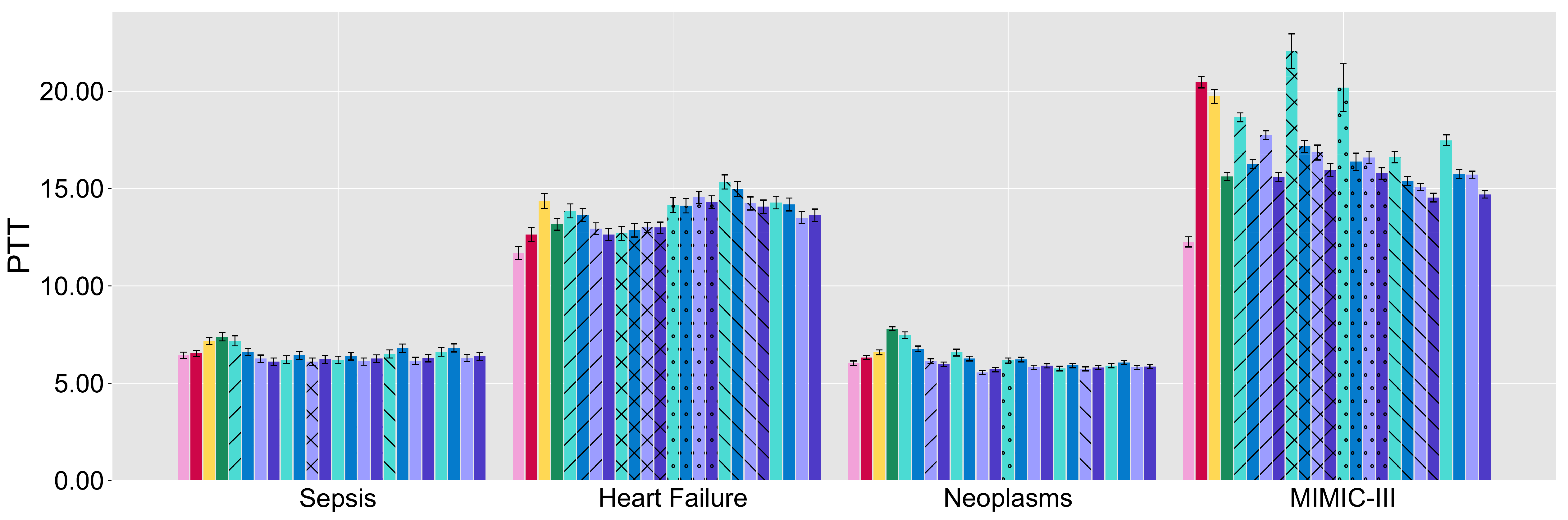}
}
\caption{\textbf{The mean absolute error (MAE) of online imputation under different $Q$ for all cohorts.} The error bars denote $\pm 1$ standard error.}
\label{fig:new_mae_p6}
\end{figure*}

\begin{figure*}
\centering
\subfigure{
\includegraphics[width=16.0cm]{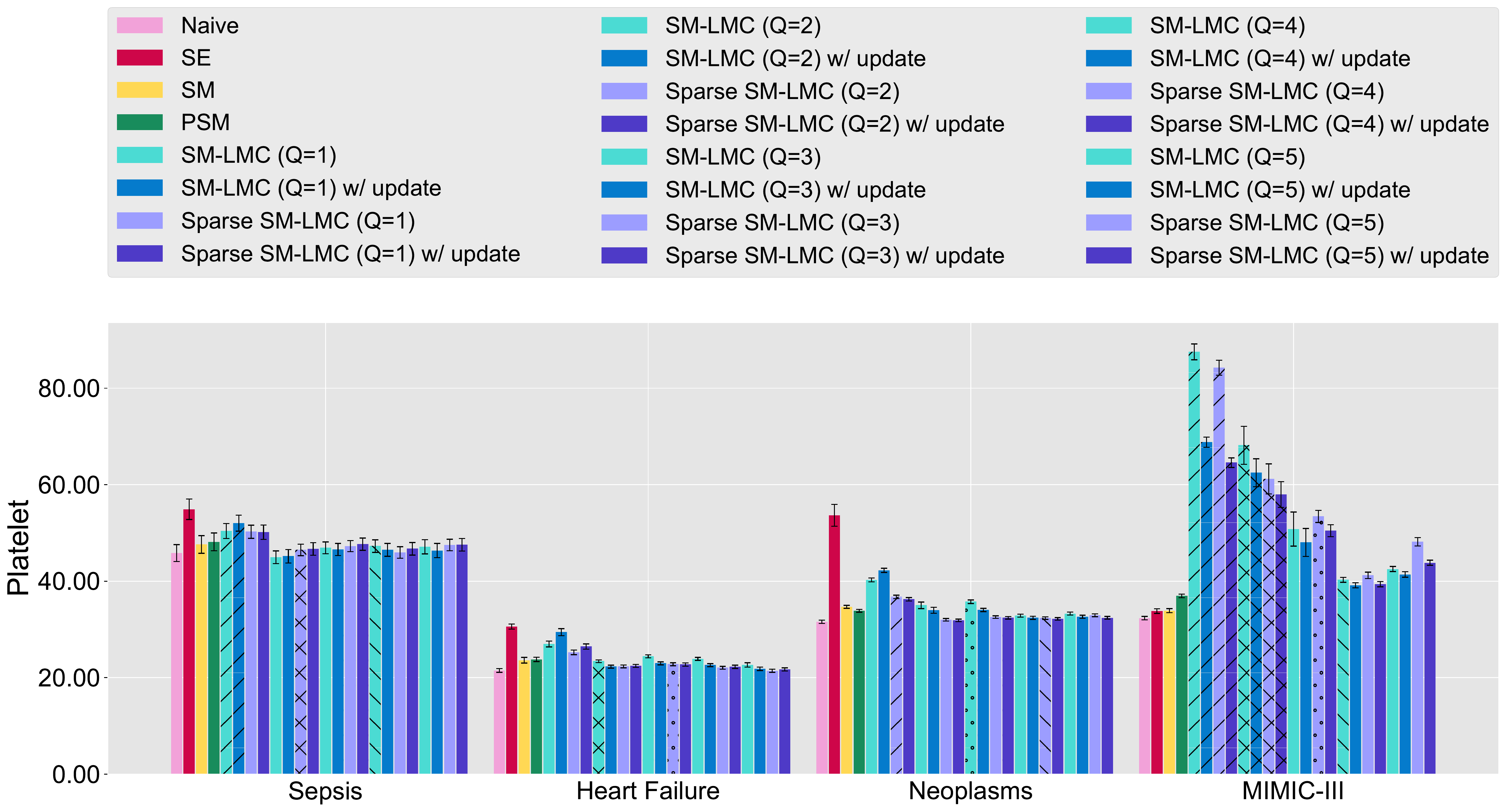}
}
\subfigure{
\includegraphics[width=16.0cm]{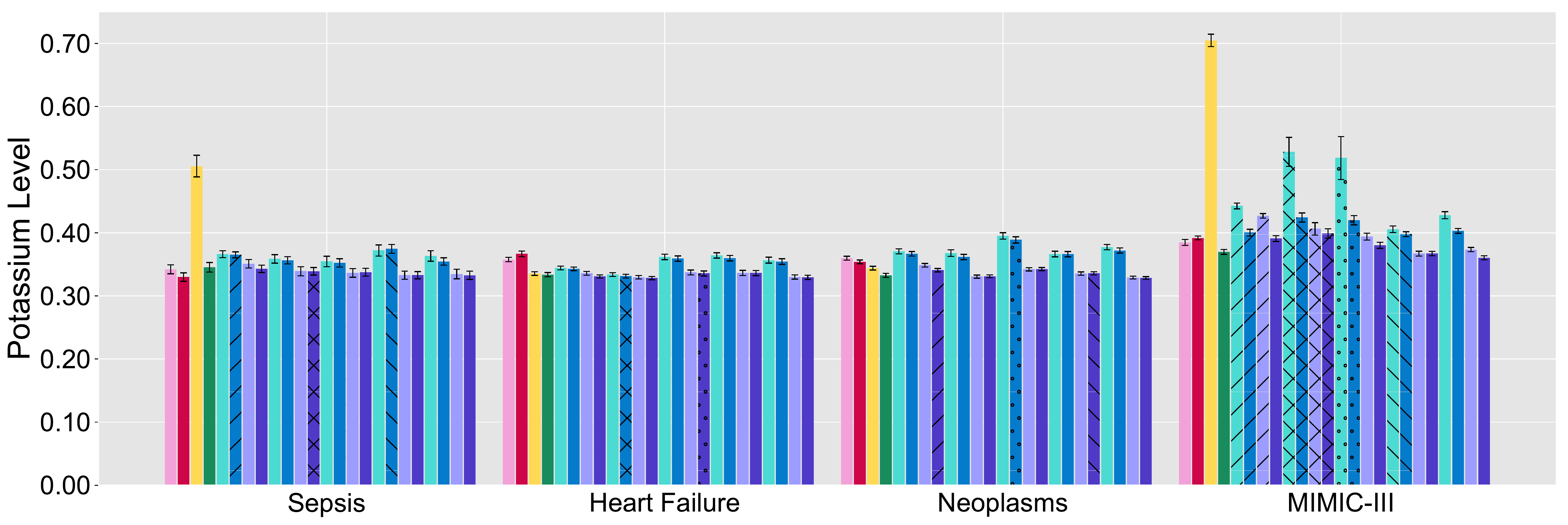}
}
\subfigure{
\includegraphics[width=16.0cm]{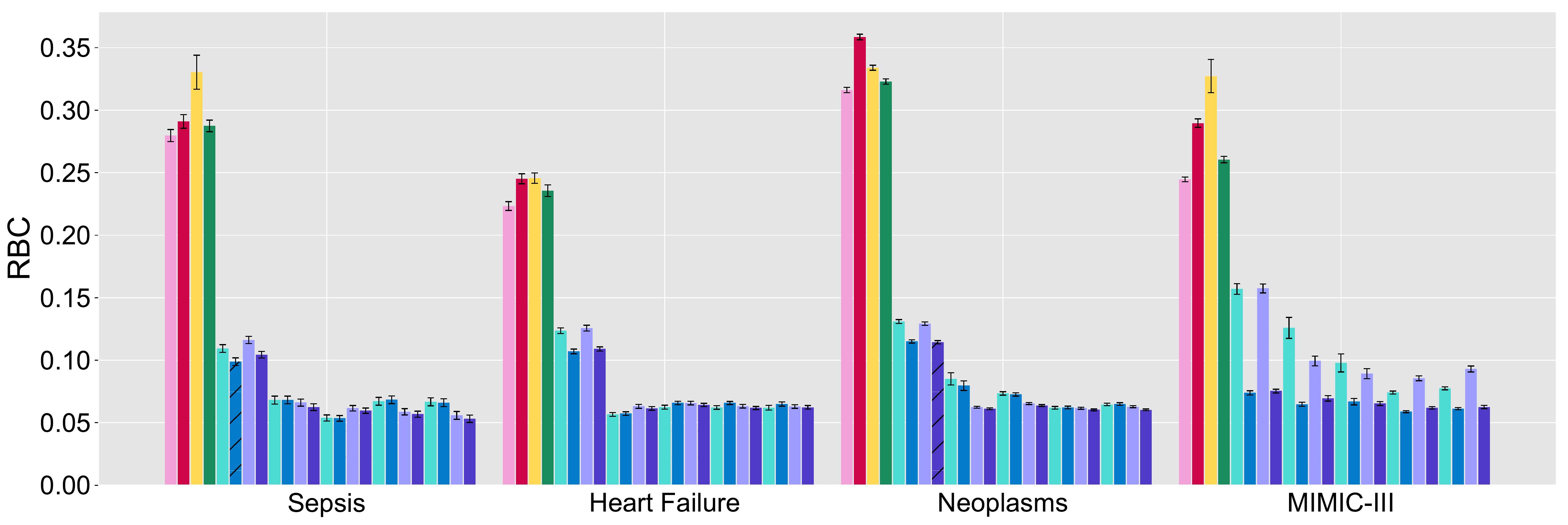}
}
\caption{\textbf{The mean absolute error (MAE) of online imputation under different $Q$ for all cohorts.} The error bars denote $\pm 1$ standard error.}
\label{fig:new_mae_p7}
\end{figure*}

\begin{figure*}
\centering
\subfigure{
\includegraphics[width=16.0cm]{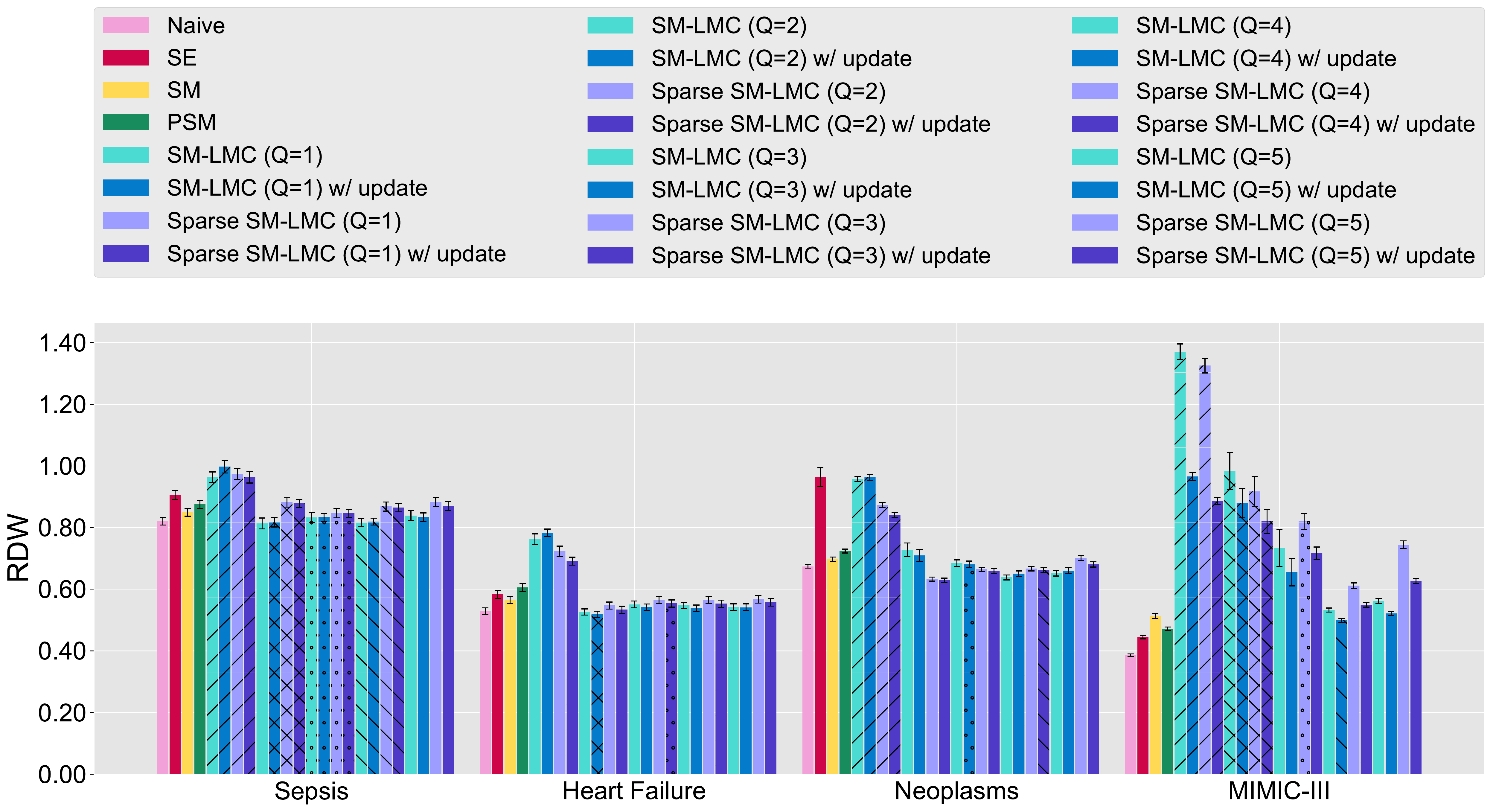}
}
\subfigure{
\includegraphics[width=16.0cm]{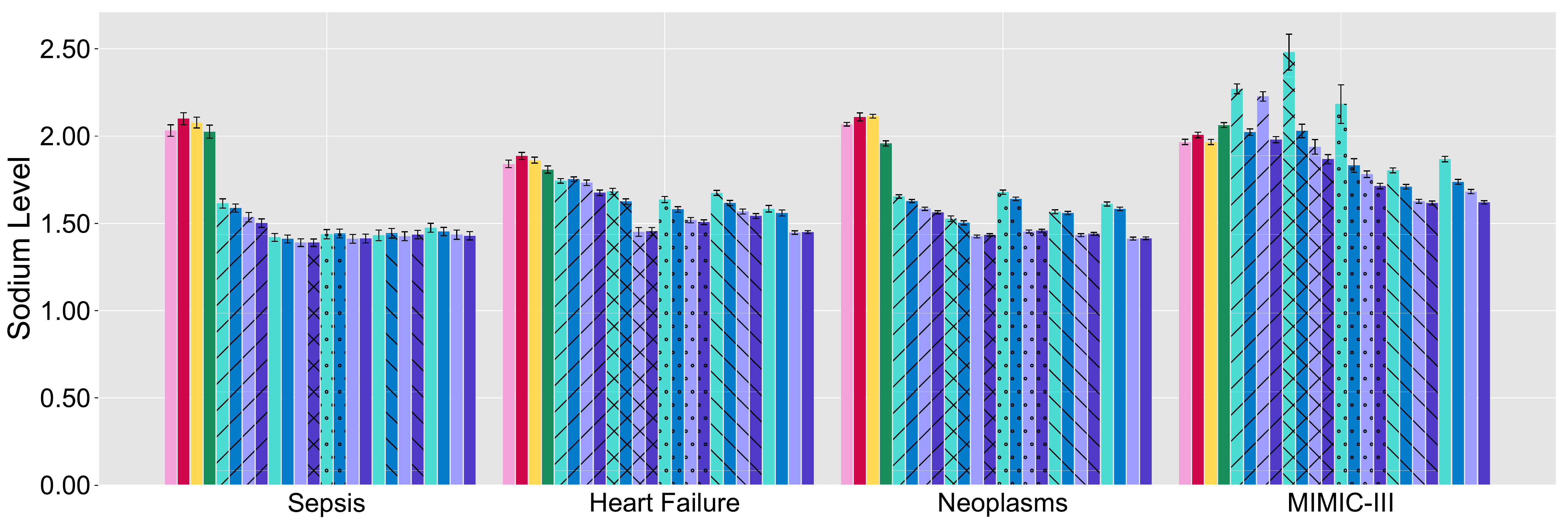}
}
\subfigure{
\includegraphics[width=16.0cm]{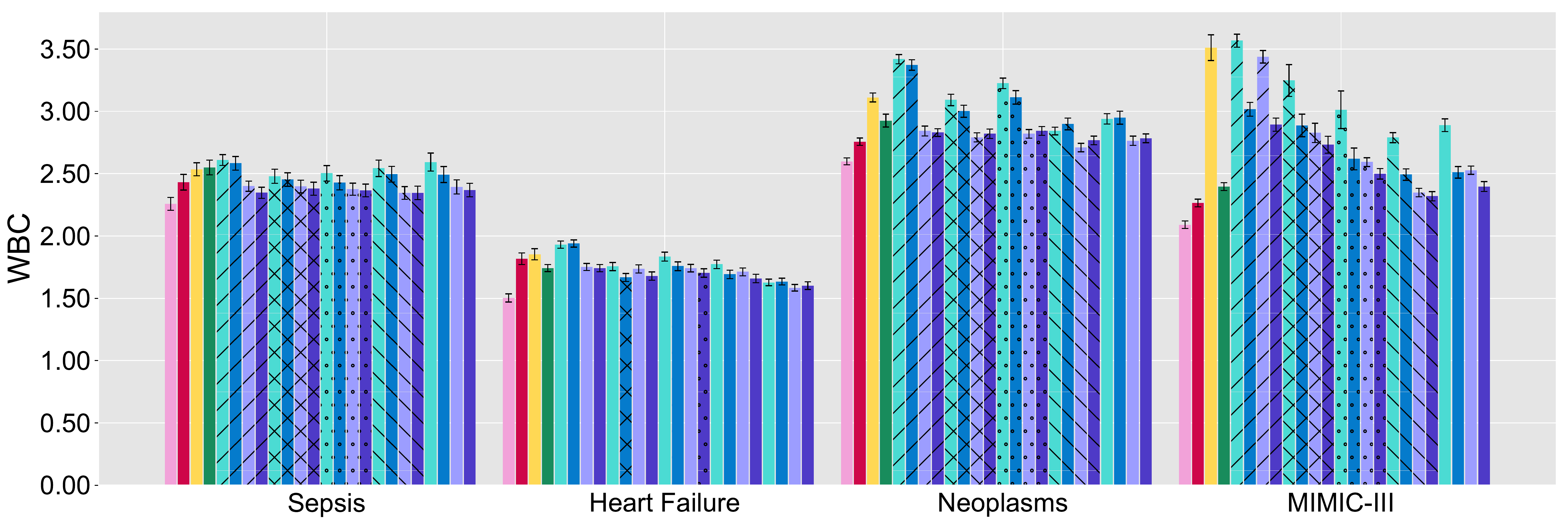}
}
\caption{\textbf{The mean absolute error (MAE) of online imputation under different $Q$ for all cohorts.} The error bars denote $\pm 1$ standard error.}
\label{fig:new_mae_p8}
\end{figure*}

\begin{figure*}
\centering
\subfigure{
\includegraphics[width=16.0cm]{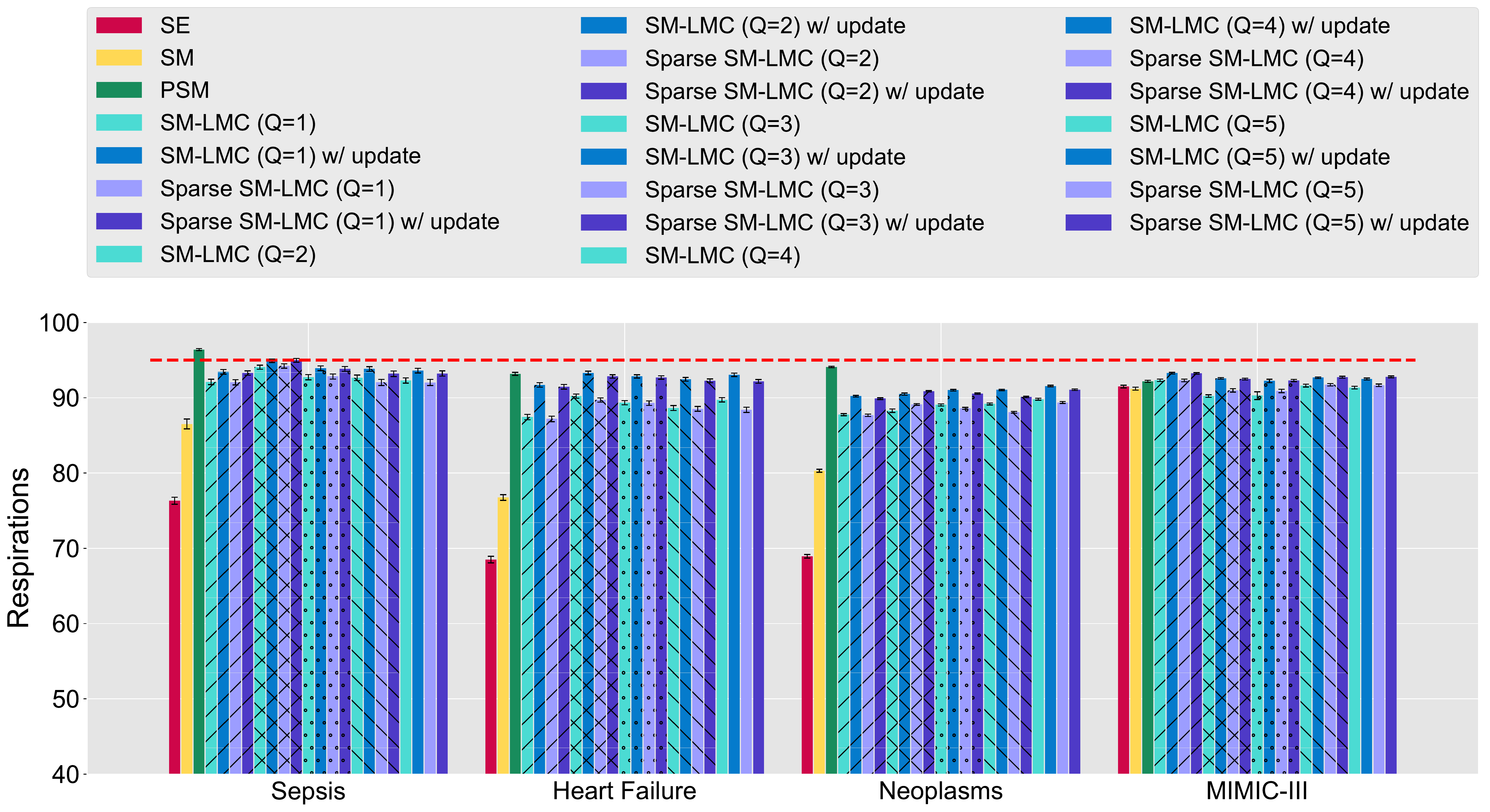}
}
\subfigure{
\includegraphics[width=16.0cm]{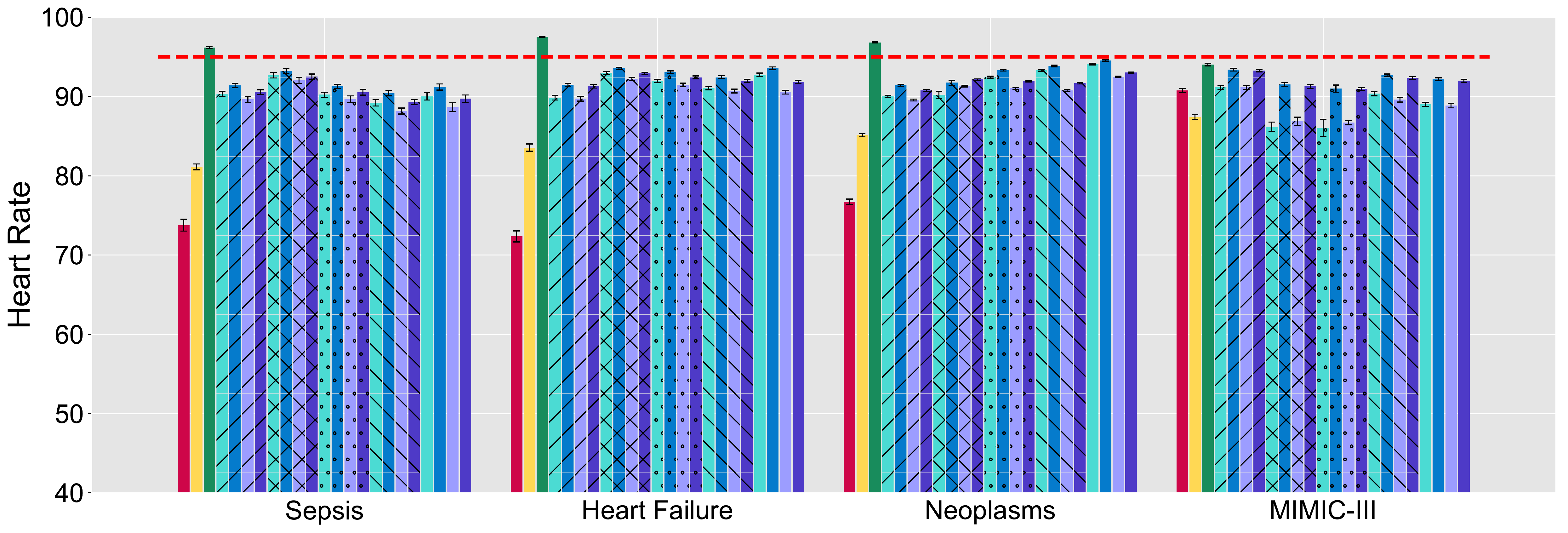}
}
\subfigure{
\includegraphics[width=16.0cm]{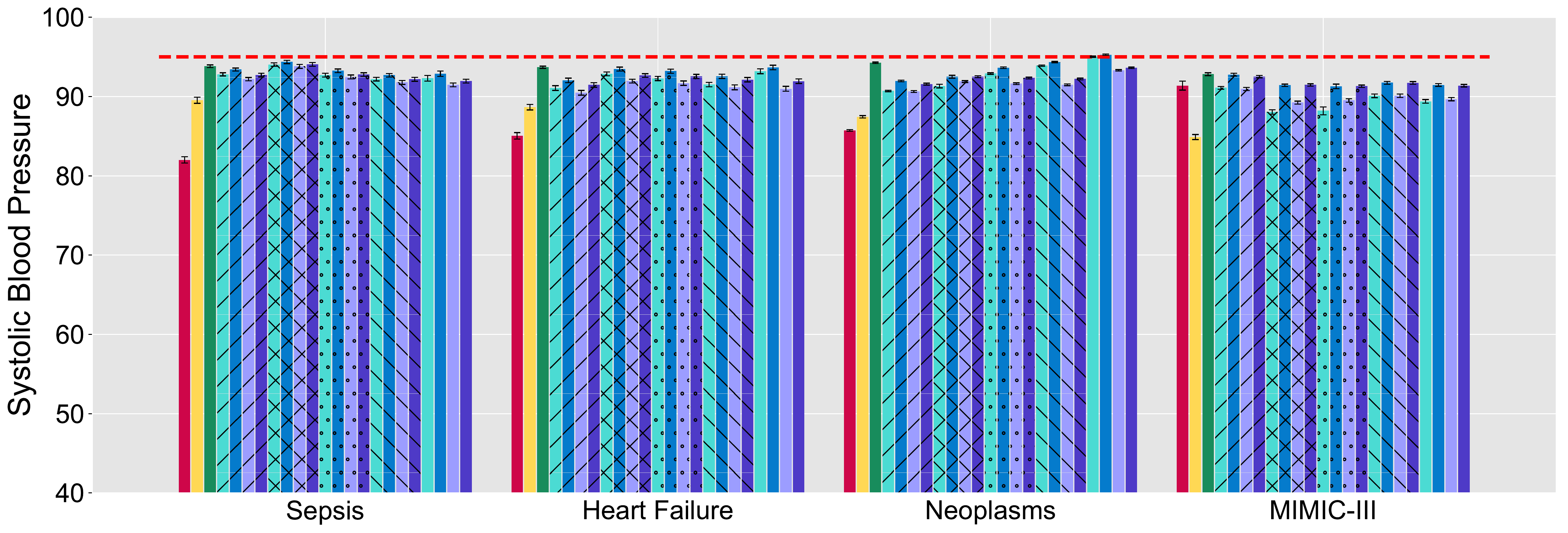}
}
\caption{\textbf{The 95\% coverage (in percentage) of online imputation under different $Q$ for all cohorts.} The error bars denote $\pm 1$ standard error. The red dashed line indicates 95\%.}
\label{fig:new_ci_p1}
\end{figure*}

\begin{figure*}
\centering
\subfigure{
\includegraphics[width=16.0cm]{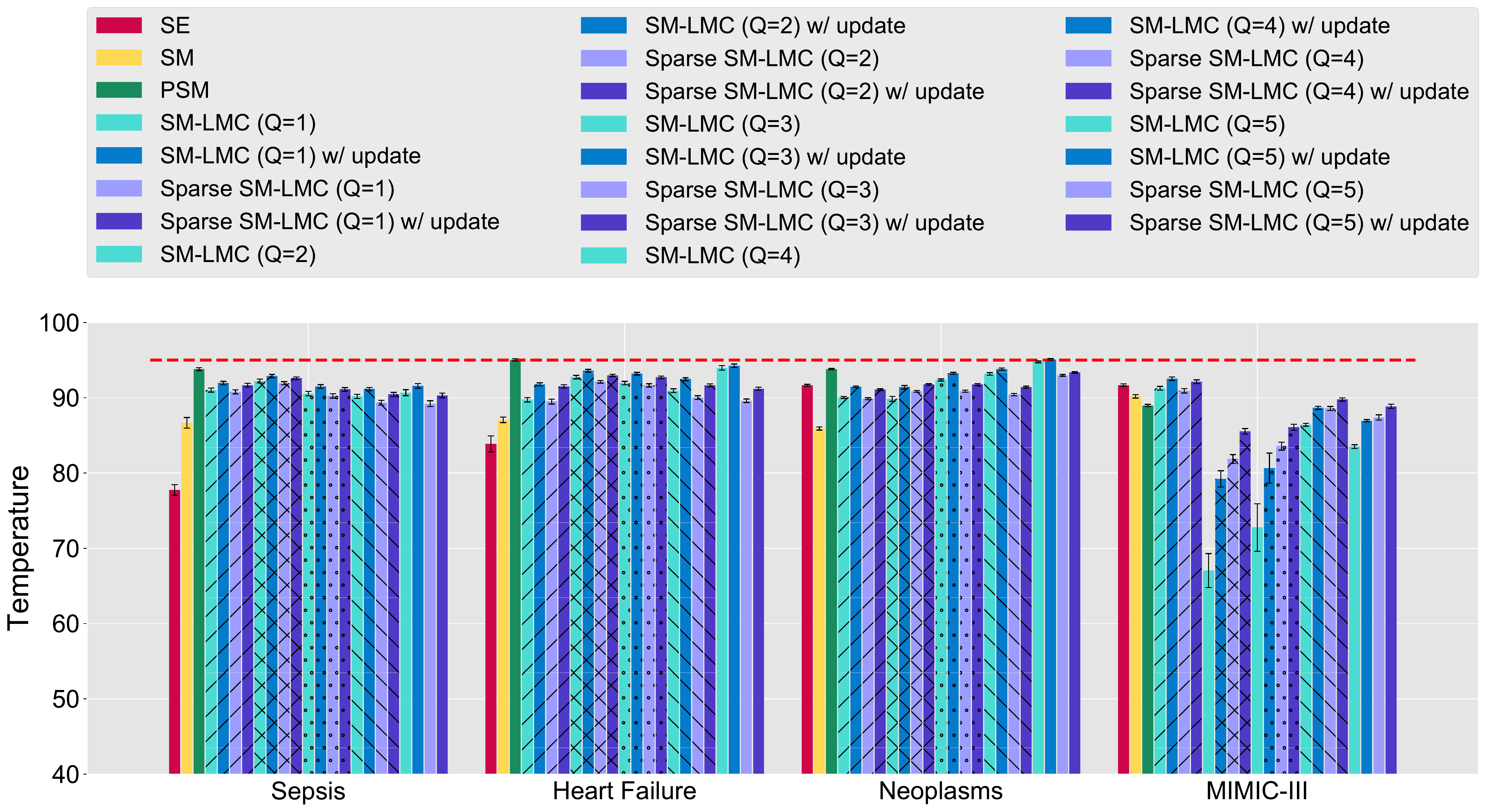}
}
\subfigure{
\includegraphics[width=16.0cm]{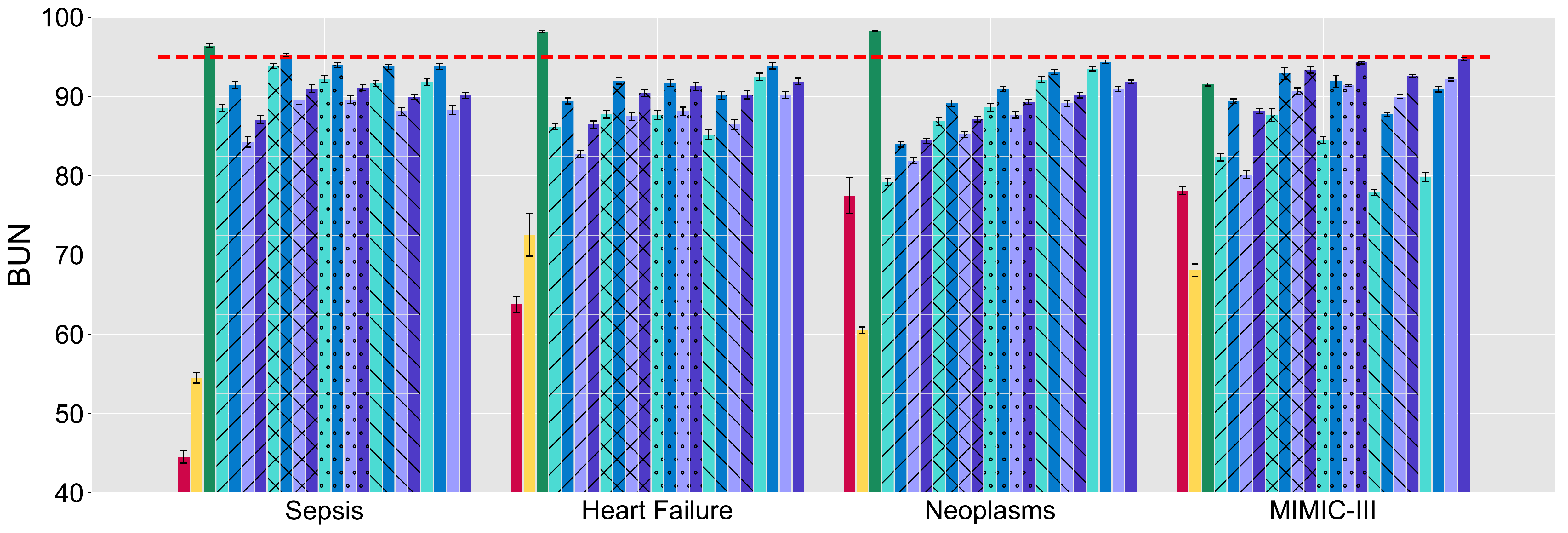}
}
\subfigure{
\includegraphics[width=16.0cm]{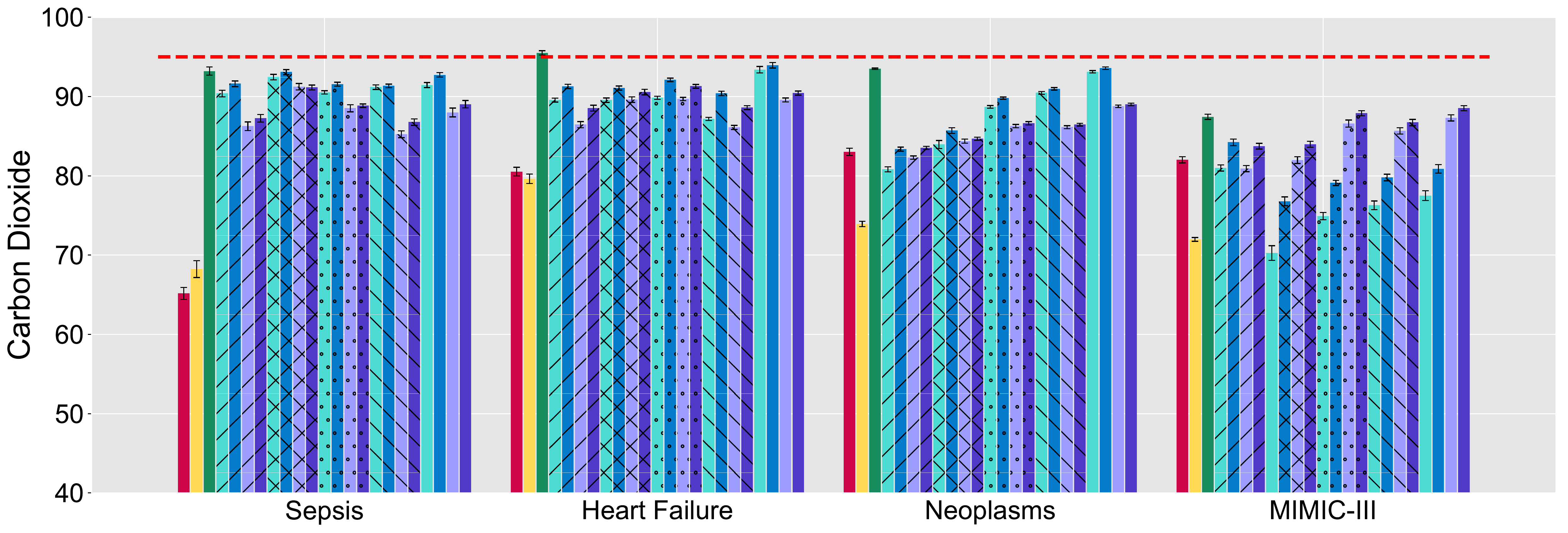}
}
\caption{\textbf{The 95\% coverage (in percentage) of online imputation under different $Q$ for all cohorts.} The error bars denote $\pm 1$ standard error. The red dashed line indicates 95\%.}
\label{fig:new_ci_p2}
\end{figure*}

\begin{figure*}
\centering
\subfigure{
\includegraphics[width=16.0cm]{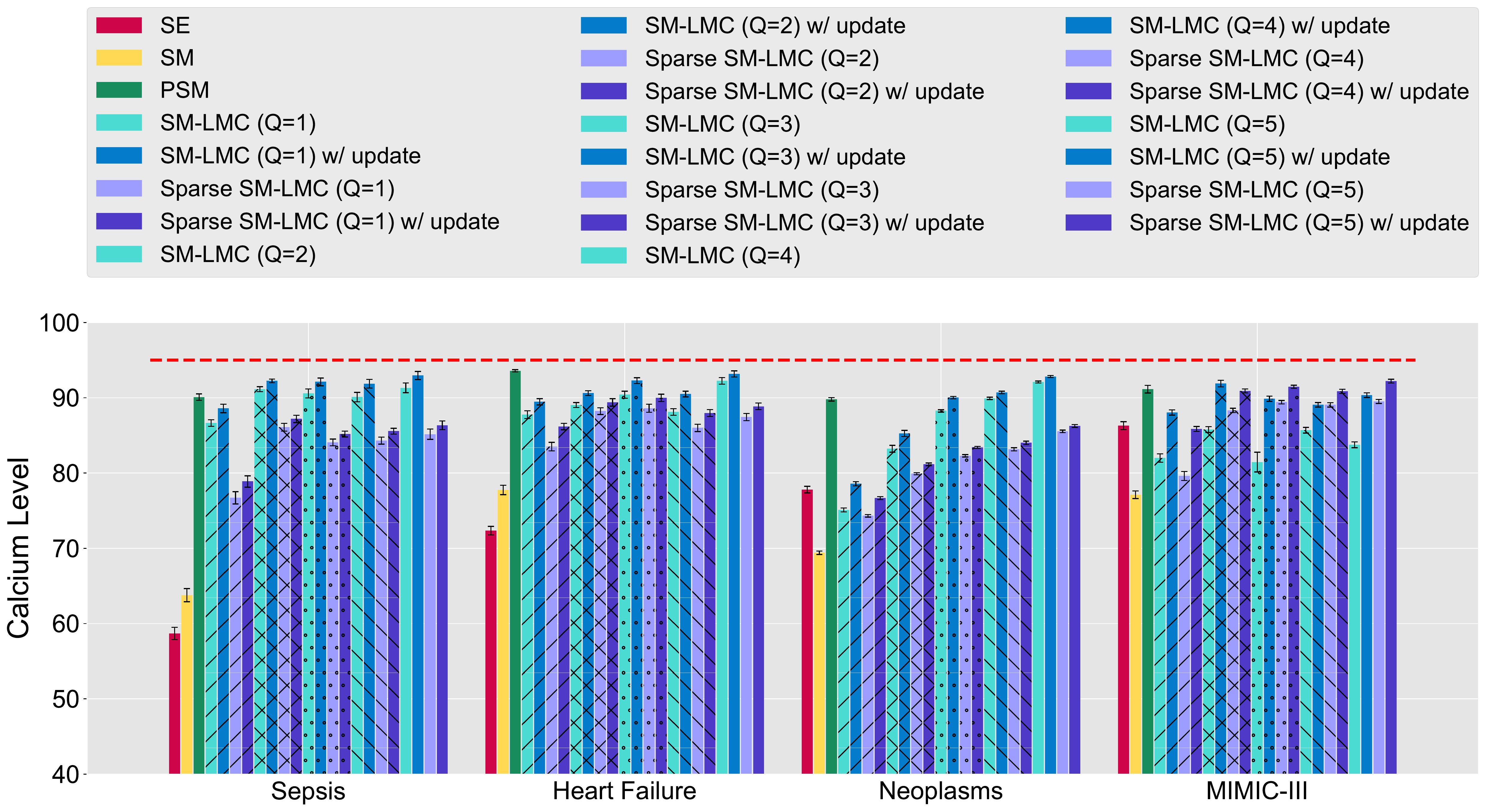}
}
\subfigure{
\includegraphics[width=16.0cm]{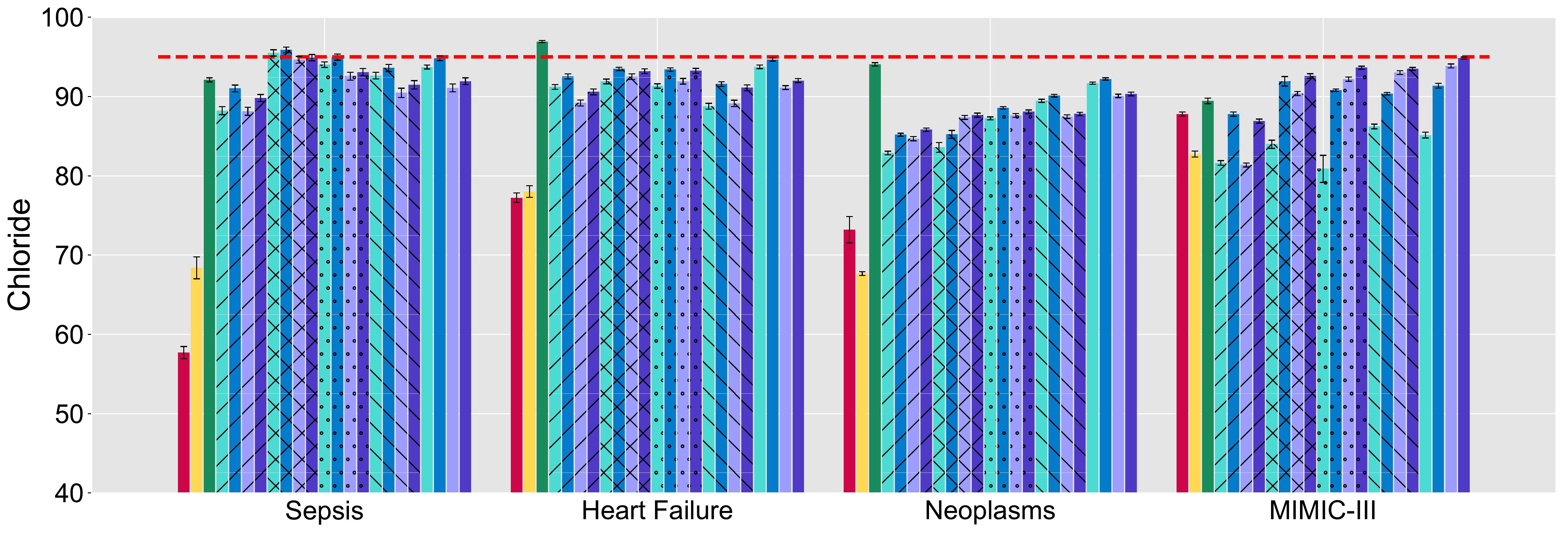}
}
\subfigure{
\includegraphics[width=16.0cm]{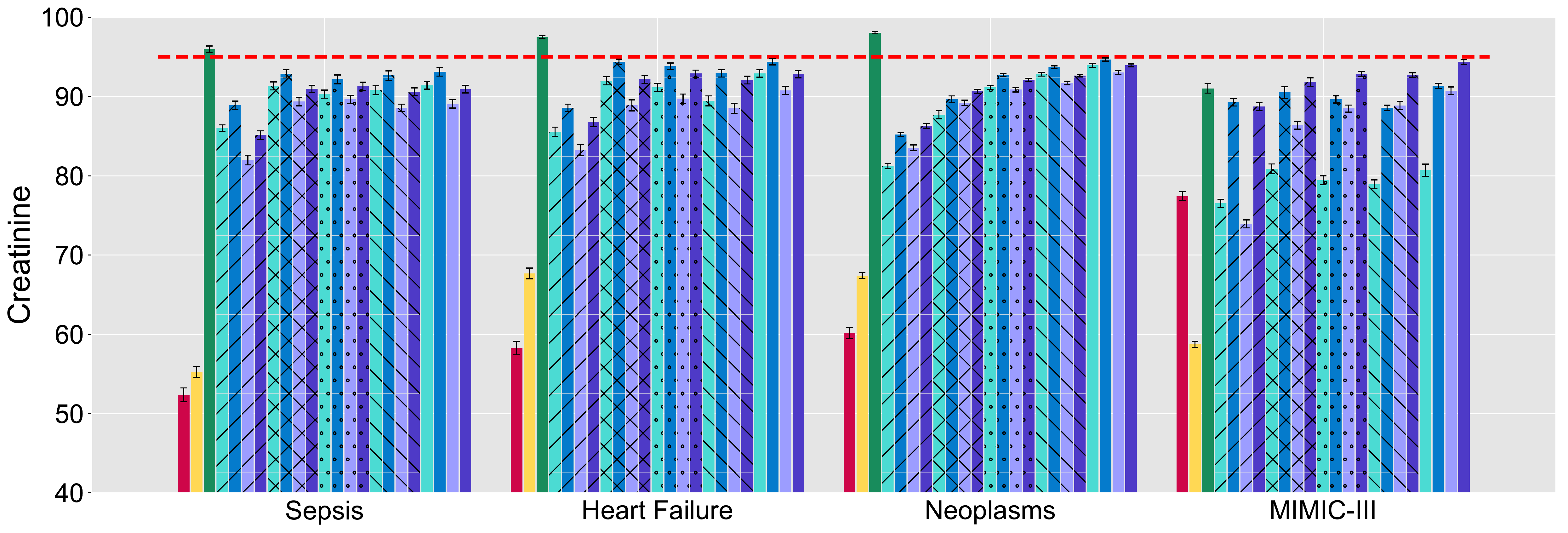}
}
\caption{\textbf{The 95\% coverage (in percentage) of online imputation under different $Q$ for all cohorts.} The error bars denote $\pm 1$ standard error. The red dashed line indicates 95\%.}
\label{fig:new_ci_p3}
\end{figure*}

\begin{figure*}
\centering
\subfigure{
\includegraphics[width=16.0cm]{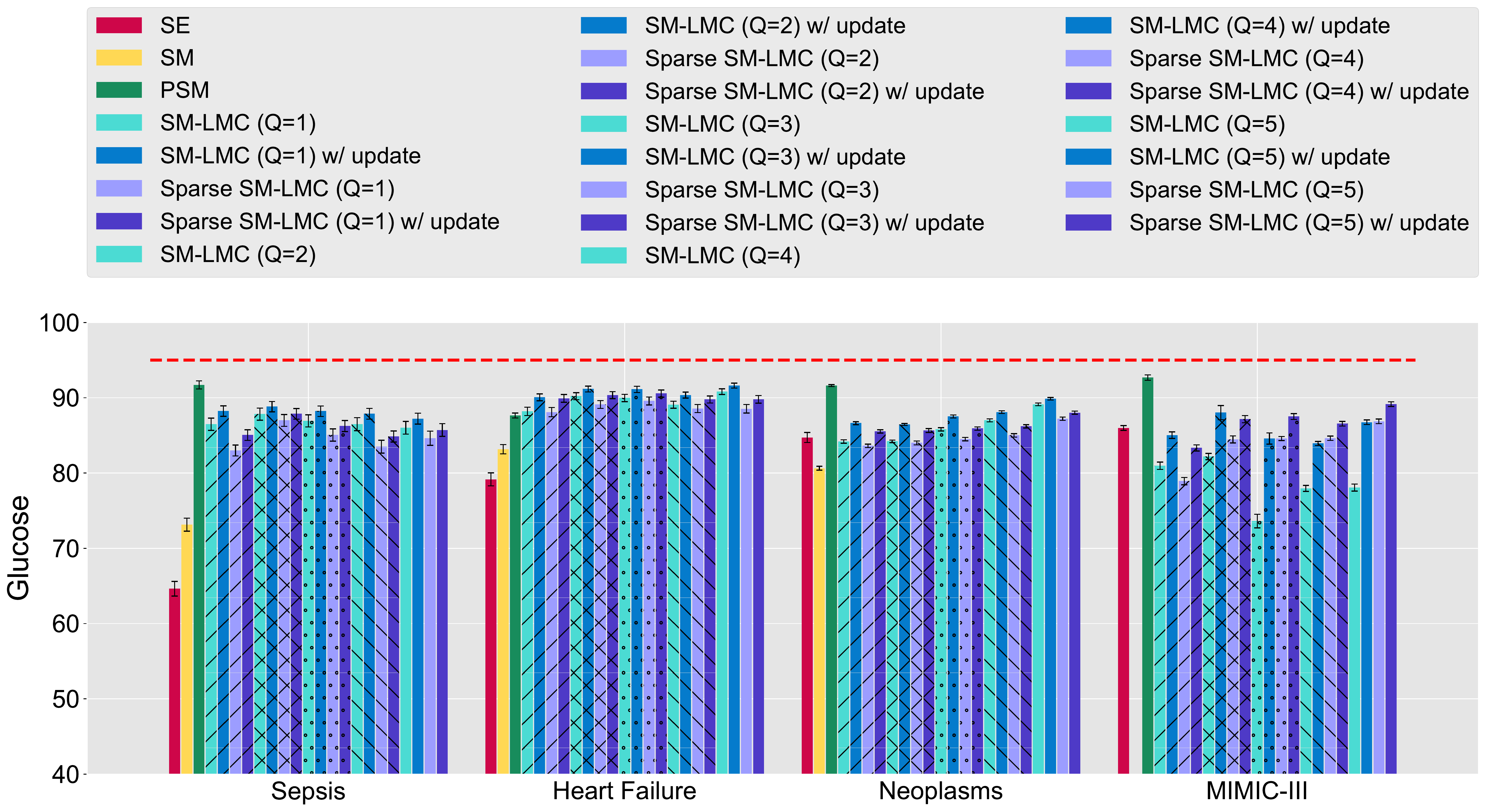}
}
\subfigure{
\includegraphics[width=16.0cm]{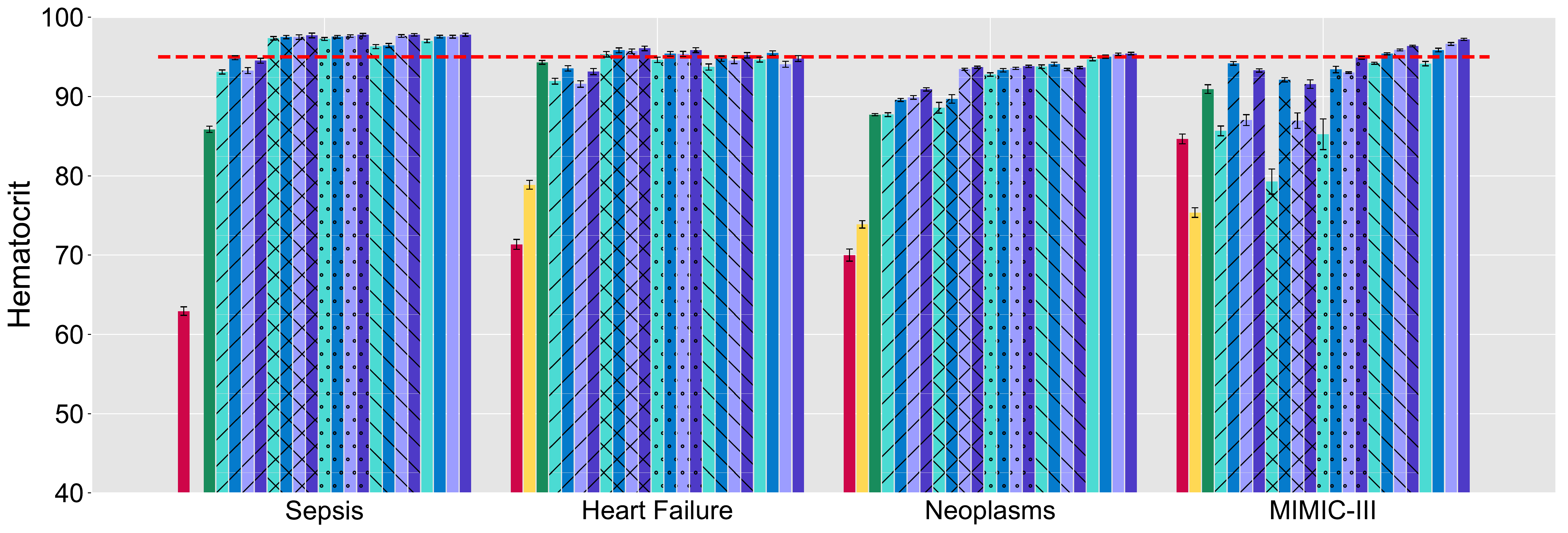}
}
\subfigure{
\includegraphics[width=16.0cm]{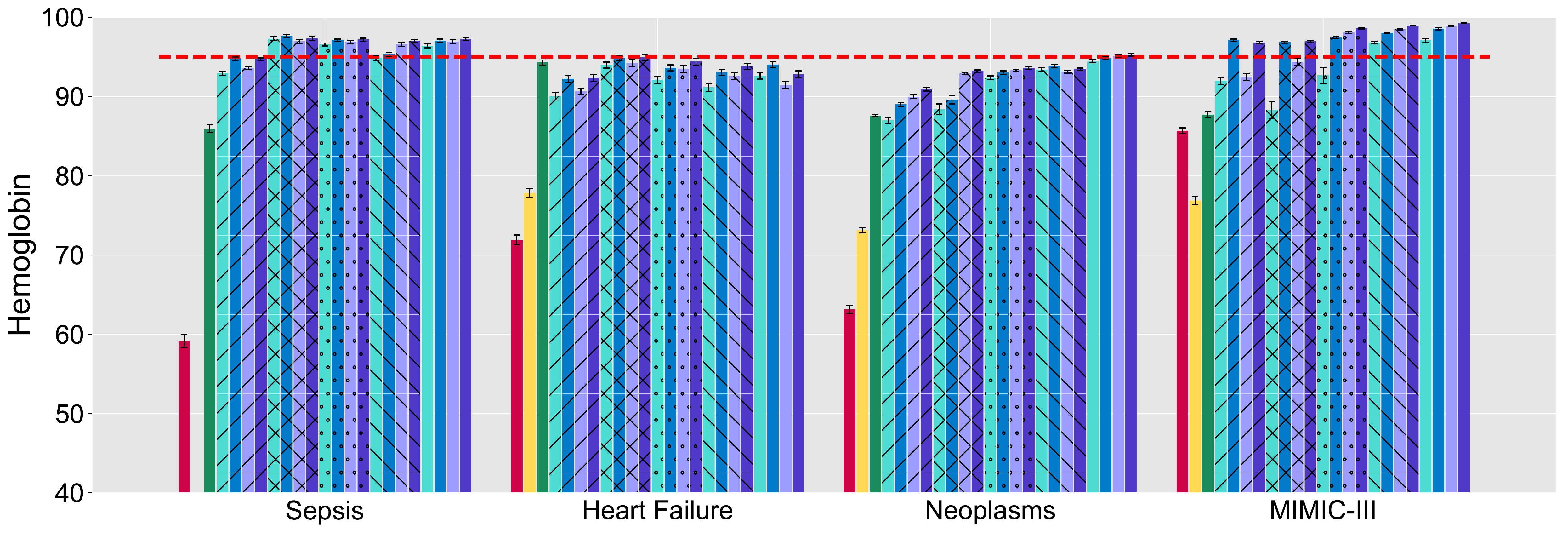}
}
\caption{\textbf{The 95\% coverage (in percentage) of online imputation under different $Q$ for all cohorts.} The error bars denote $\pm 1$ standard error. The red dashed line indicates 95\%.}
\label{fig:new_ci_p4}
\end{figure*}

\begin{figure*}
\centering
\subfigure{
\includegraphics[width=16.0cm]{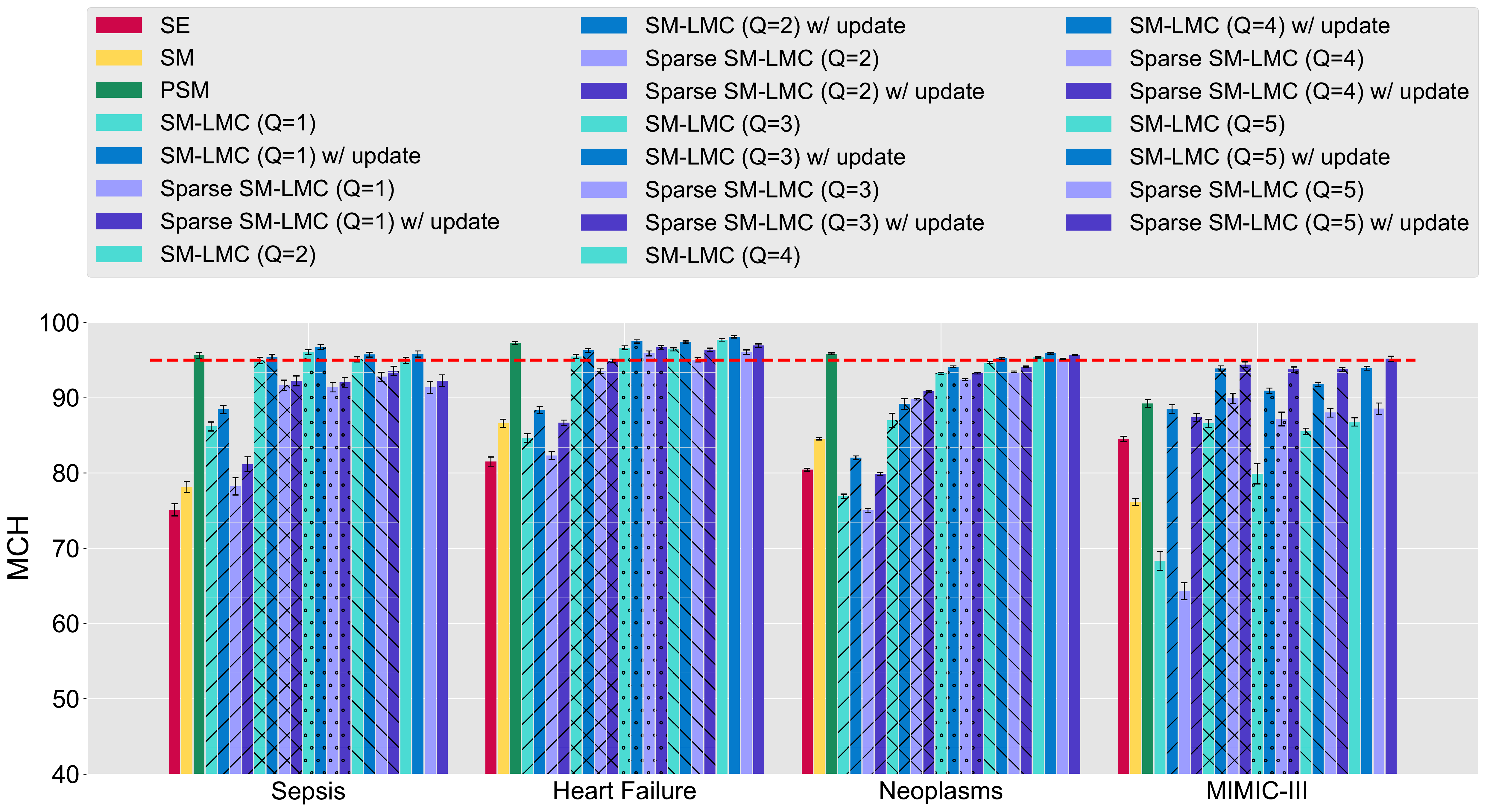}
}
\subfigure{
\includegraphics[width=16.0cm]{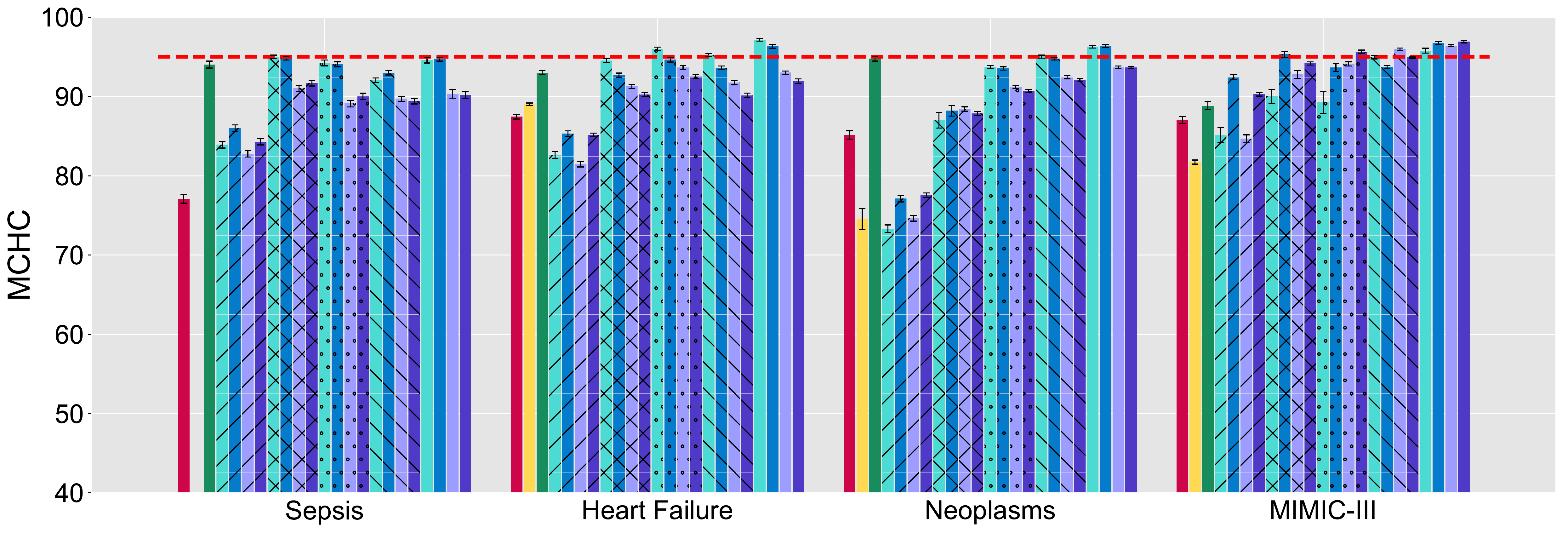}
}
\subfigure{
\includegraphics[width=16.0cm]{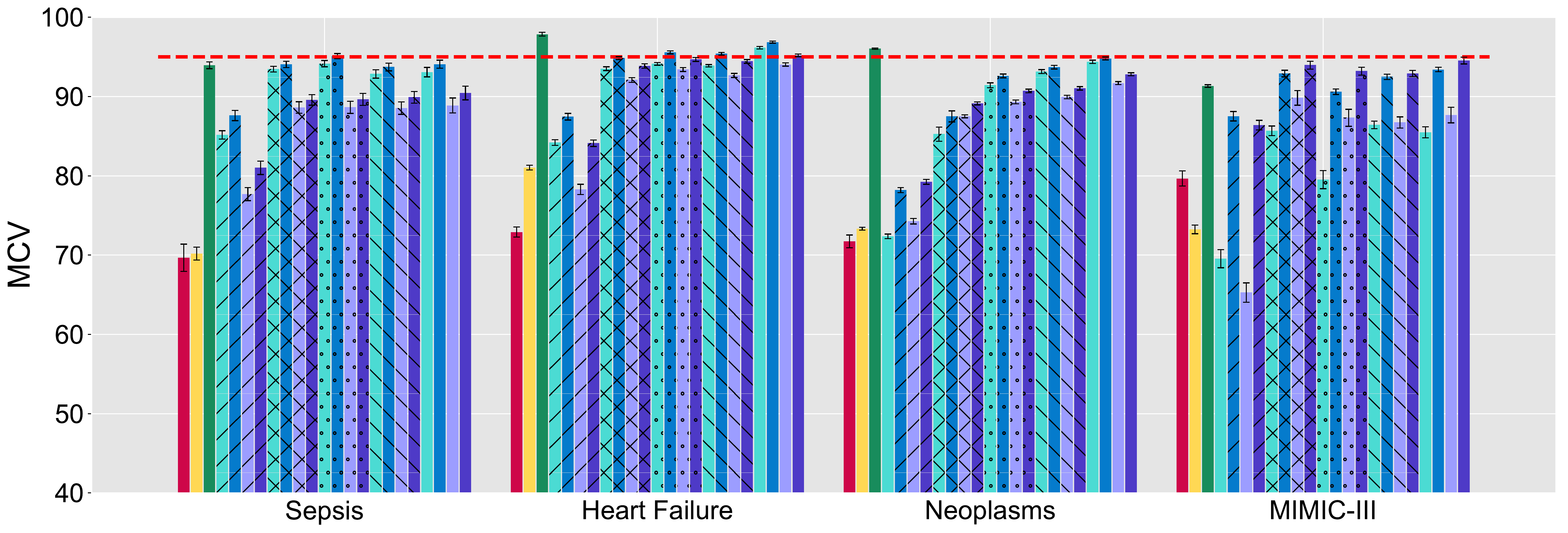}
}
\caption{\textbf{The 95\% coverage (in percentage) of online imputation under different $Q$ for all cohorts.} The error bars denote $\pm 1$ standard error. The red dashed line indicates 95\%.}
\label{fig:new_ci_p5}
\end{figure*}

\begin{figure*}
\centering
\subfigure{
\includegraphics[width=16.0cm]{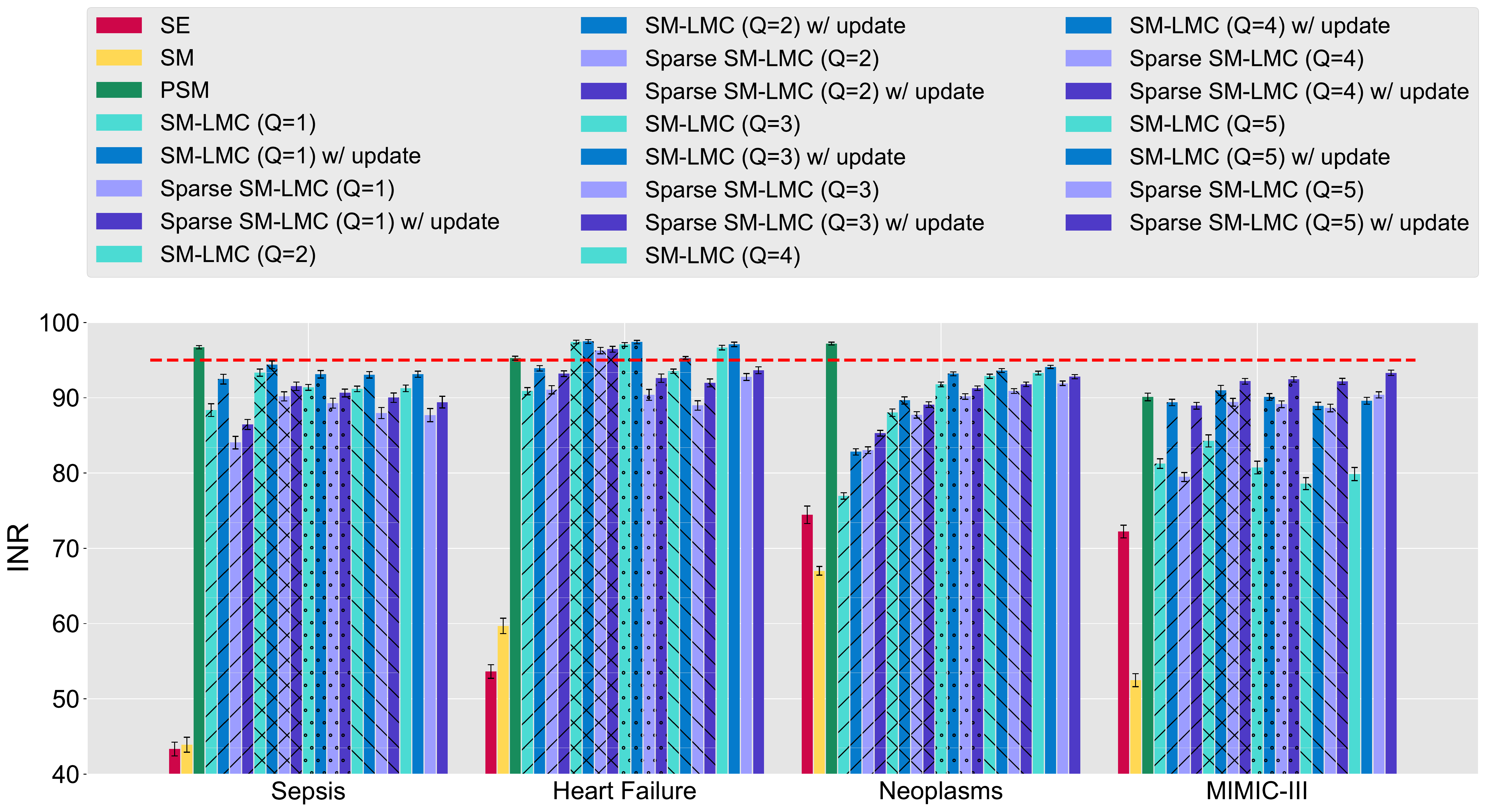}
}
\subfigure{
\includegraphics[width=16.0cm]{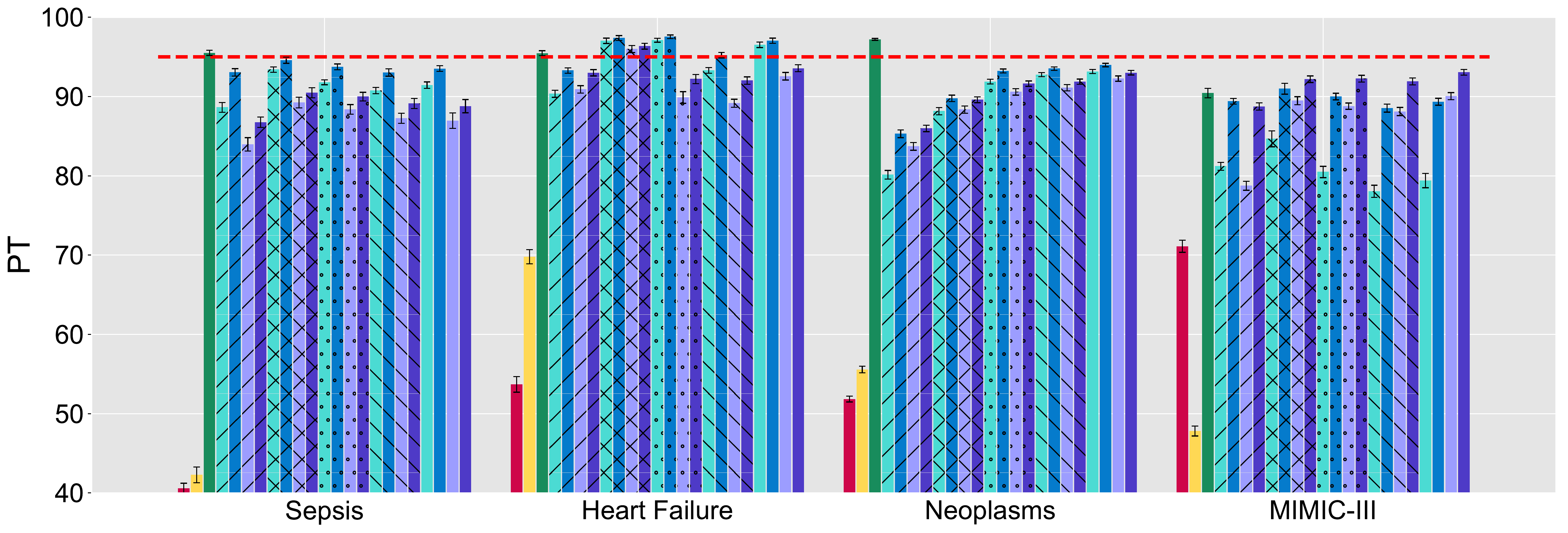}
}
\subfigure{
\includegraphics[width=16.0cm]{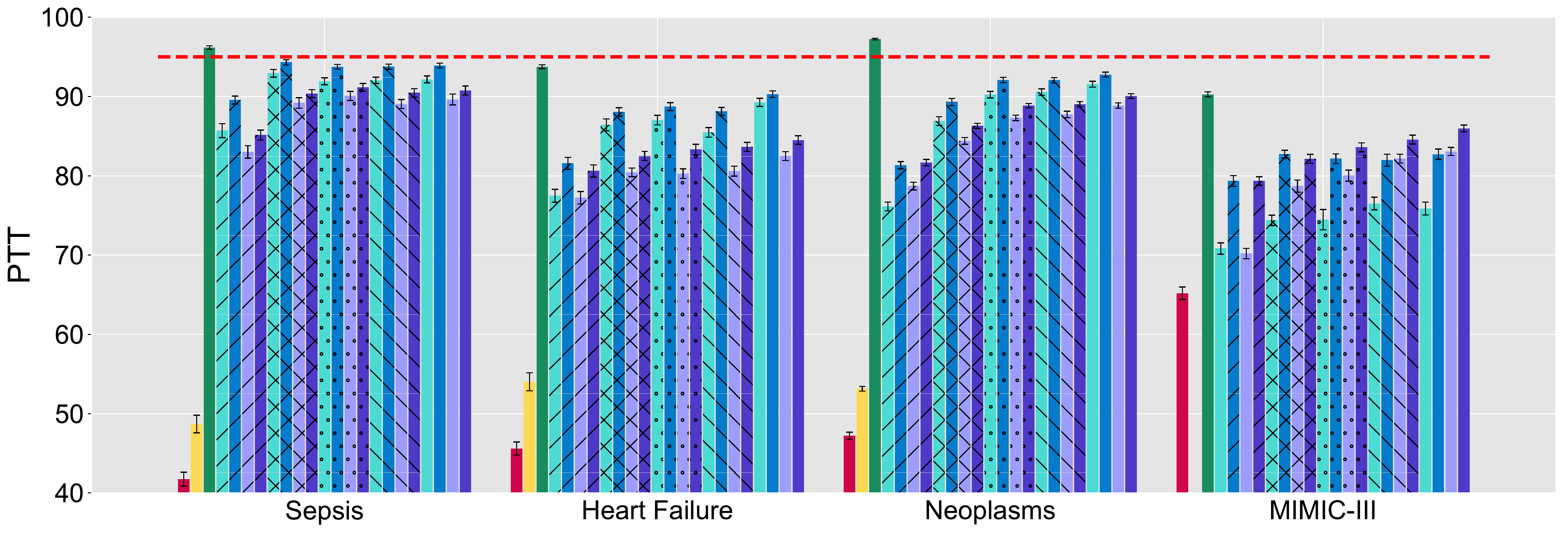}
}
\caption{\textbf{The 95\% coverage (in percentage) of online imputation under different $Q$ for all cohorts.} The error bars denote $\pm 1$ standard error. The red dashed line indicates 95\%.}
\label{fig:new_ci_p6}
\end{figure*}

\begin{figure*}
\centering
\subfigure{
\includegraphics[width=16.0cm]{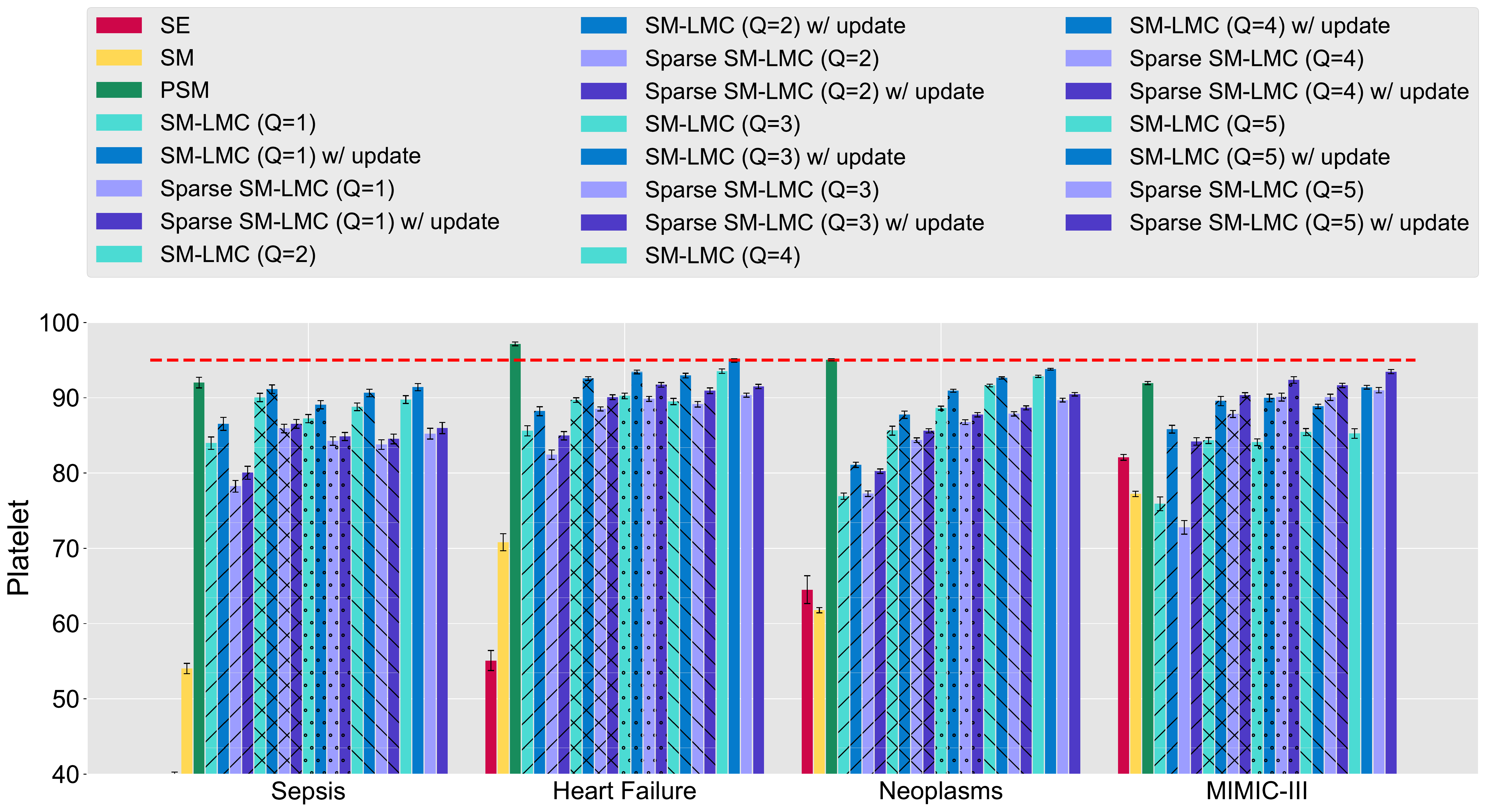}
}
\subfigure{
\includegraphics[width=16.0cm]{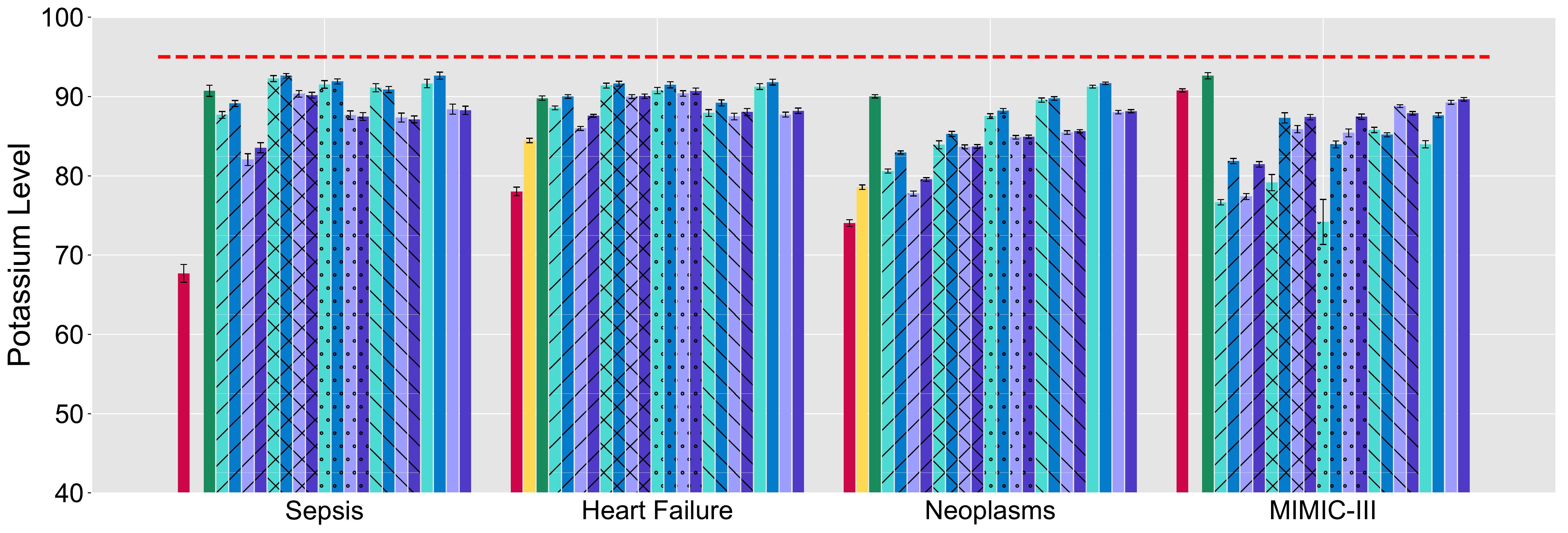}
}
\subfigure{
\includegraphics[width=16.0cm]{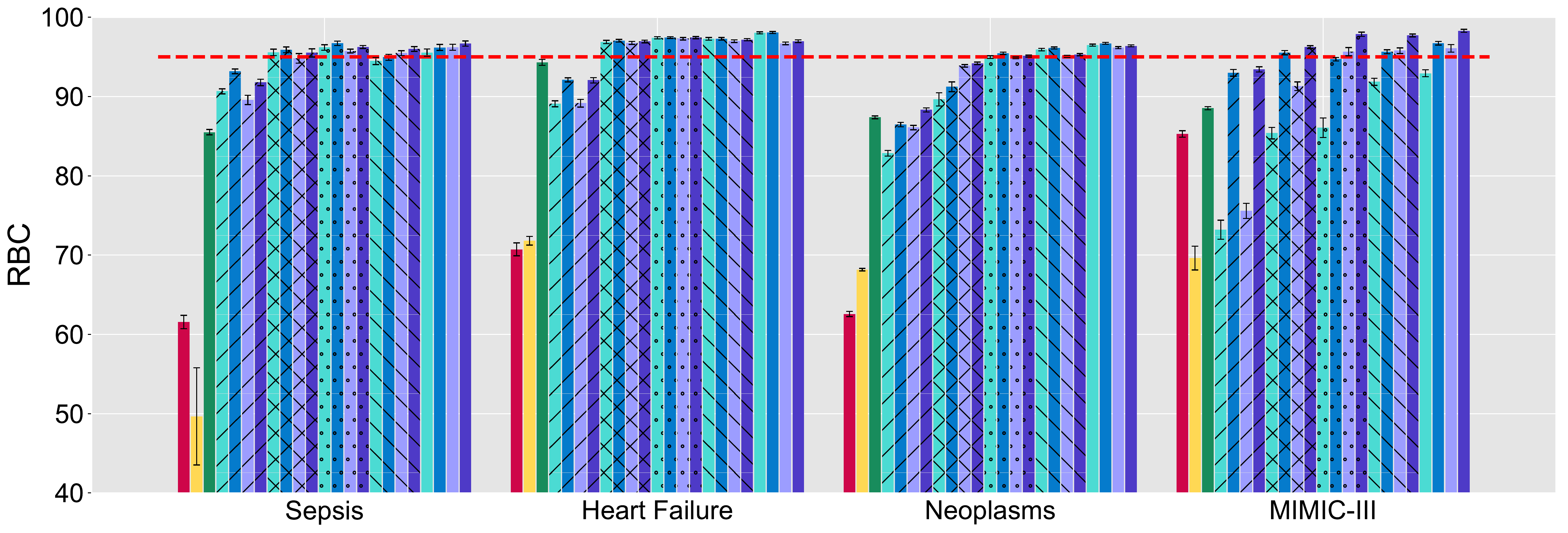}
}
\caption{\textbf{The 95\% coverage (in percentage) of online imputation under different $Q$ for all cohorts.} The error bars denote $\pm 1$ standard error. The red dashed line indicates 95\%.}
\label{fig:new_ci_p7}
\end{figure*}

\begin{figure*}
\centering
\subfigure{
\includegraphics[width=16.0cm]{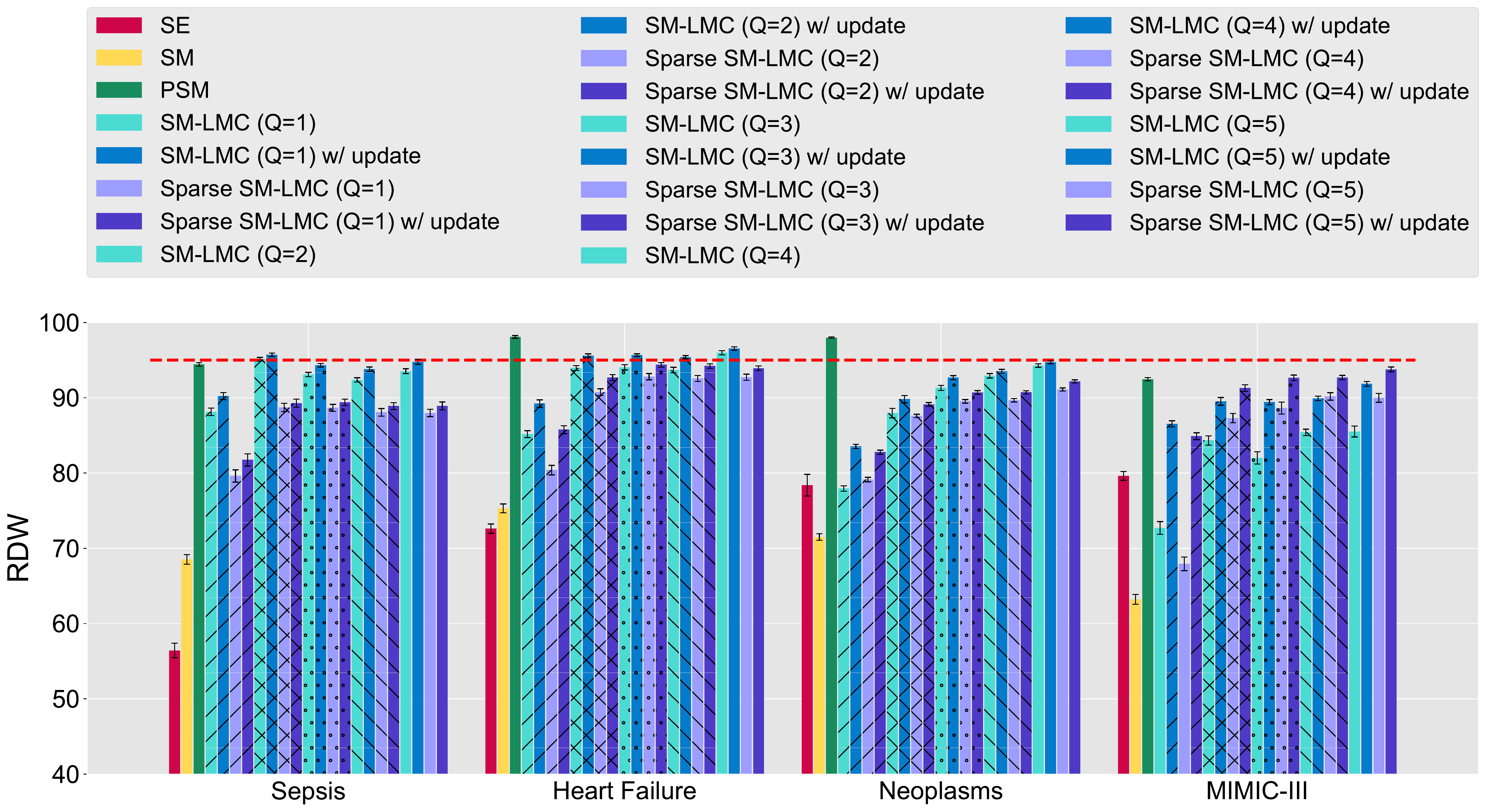}
}
\subfigure{
\includegraphics[width=16.0cm]{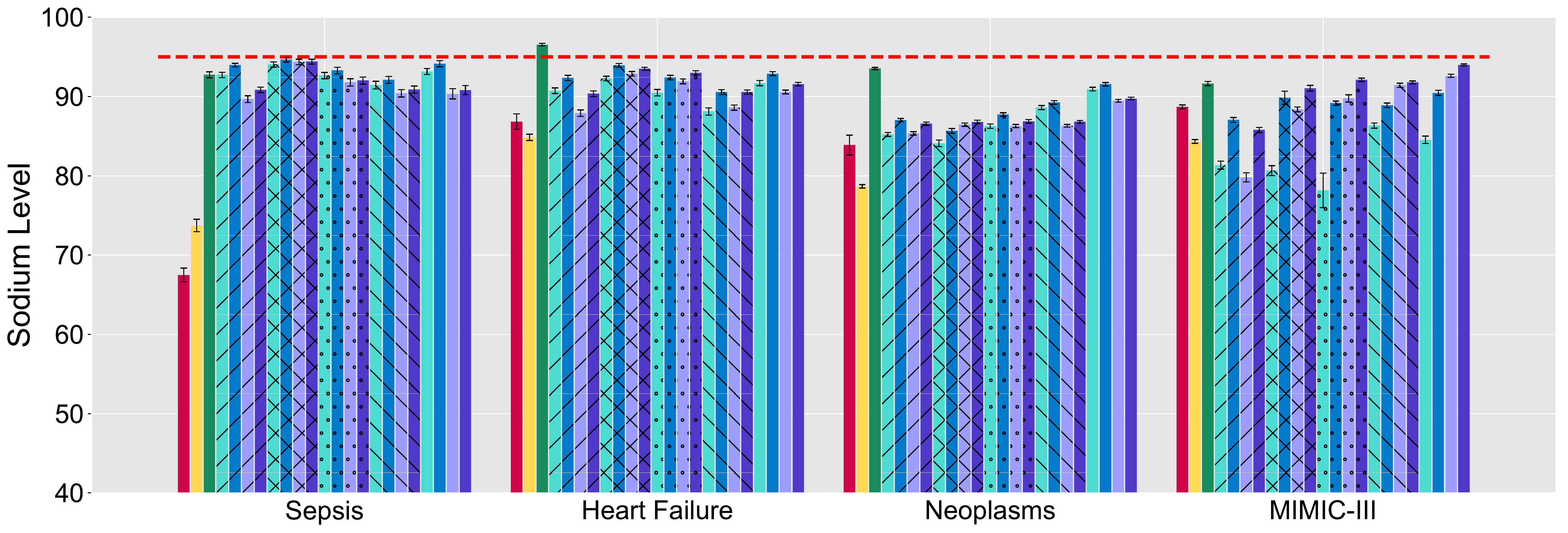}
}
\subfigure{
\includegraphics[width=16.0cm]{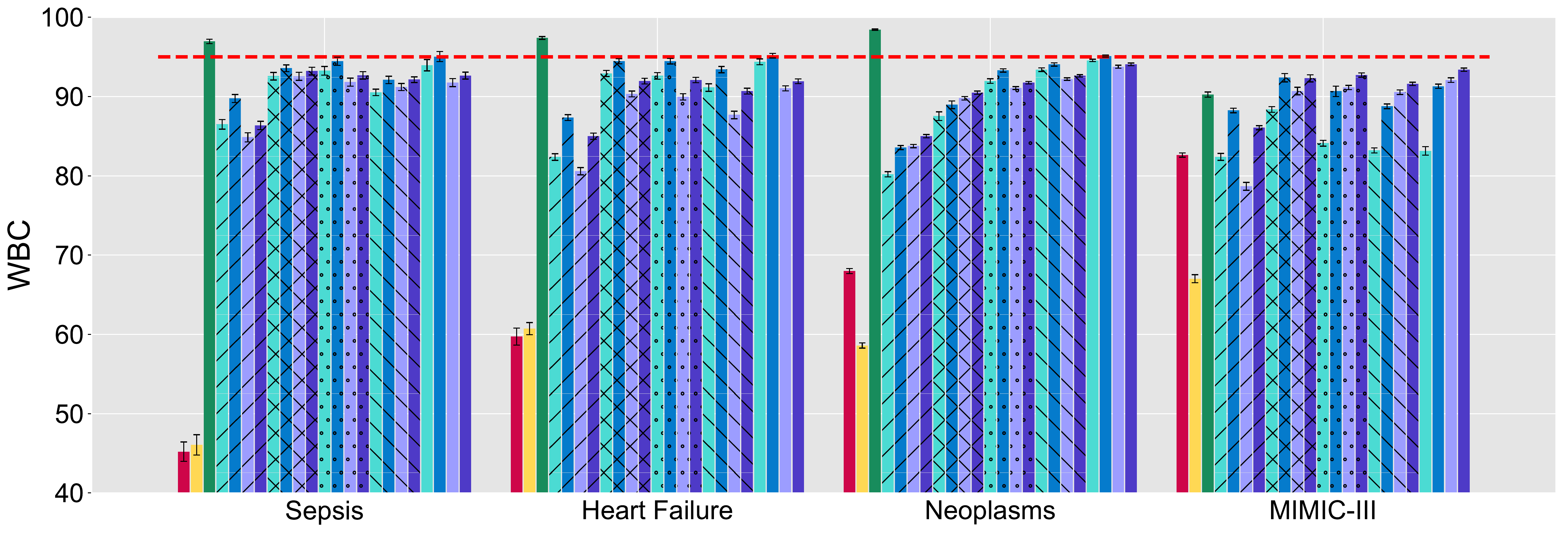}
}
\caption{\textbf{The 95\% coverage (in percentage) of online imputation under different $Q$ for all cohorts.} The error bars denote $\pm 1$ standard error. The red dashed line indicates 95\%.}
\label{fig:new_ci_p8}
\end{figure*}

In this appendix, we showed more detailed results of the experiments using different number of basis kernels. We ran experiments with for $Q=1,\cdots,5$ on all four subsets. The results include all three subgroups in the HUP data set and the MIMIC-III heart failure subset. We visualized the results in Figure~\ref{fig:new_mae_p1}--\ref{fig:new_ci_p8}. We noticed that for most of the covariates, the imputation performance (both MAE and 95\% coverage) improves as the number of $Q$ increases. We also observed that the best number of $Q$ varies across covariates under different metrics. For instance, for lab covariates INR and PT, we observed that setting $Q=1$ or $Q=2$ reduces MAE compared with $Q=5$, but the coverage still improves after $Q=2$. Allowing more numbers of basis kernels increases the flexibility for customization, but also increases complexity and thus the risk of overfitting for some covariates or patients. Overall $Q=5$ for HUP subsets and $Q=4$ for the MIMIC-III subset reached the largest number of covariates improved over the best of baselines using imputation error as the performance metric. How to improve the performance for a specific clinical covariate at patient-level would be one future direction of interest.

\begin{figure*}
\centering
\subfigure{
\includegraphics[width=16.0cm]{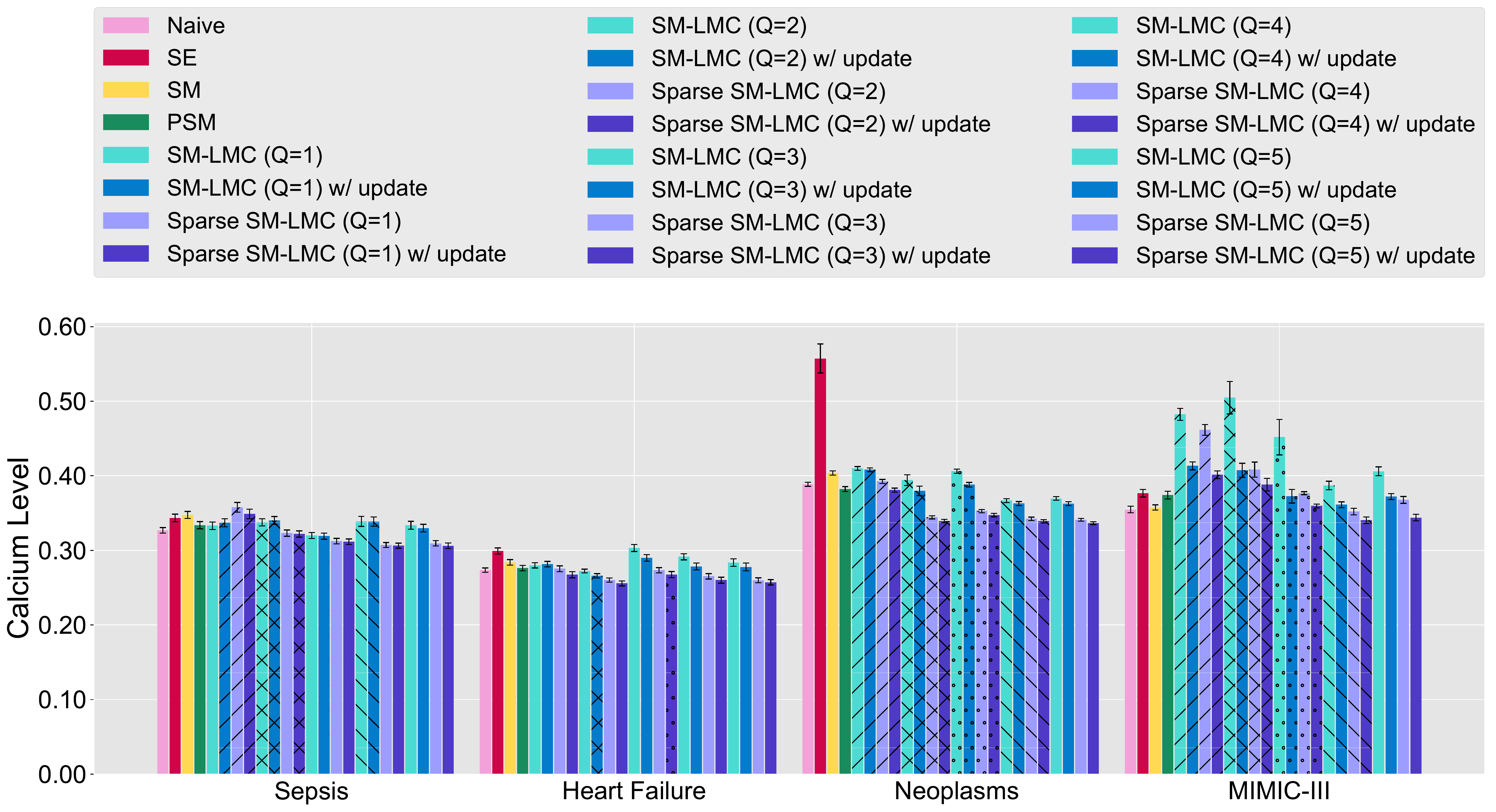}
}
\subfigure{
\includegraphics[width=16.0cm]{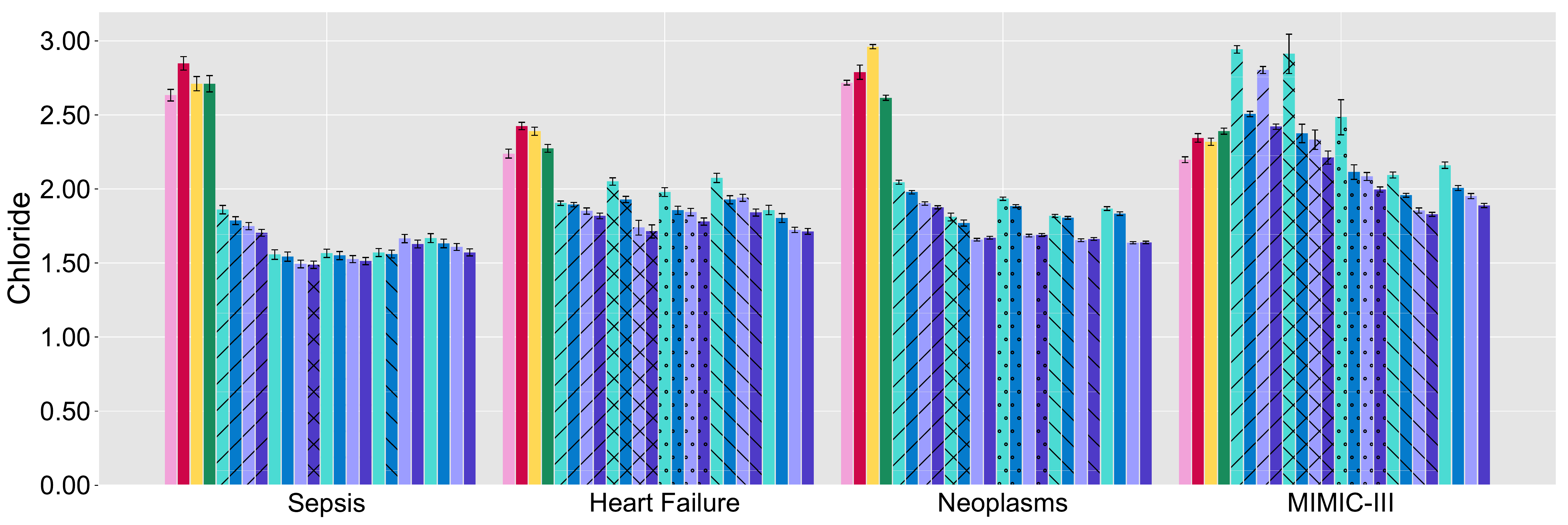}
}
\subfigure{
\includegraphics[width=16.0cm]{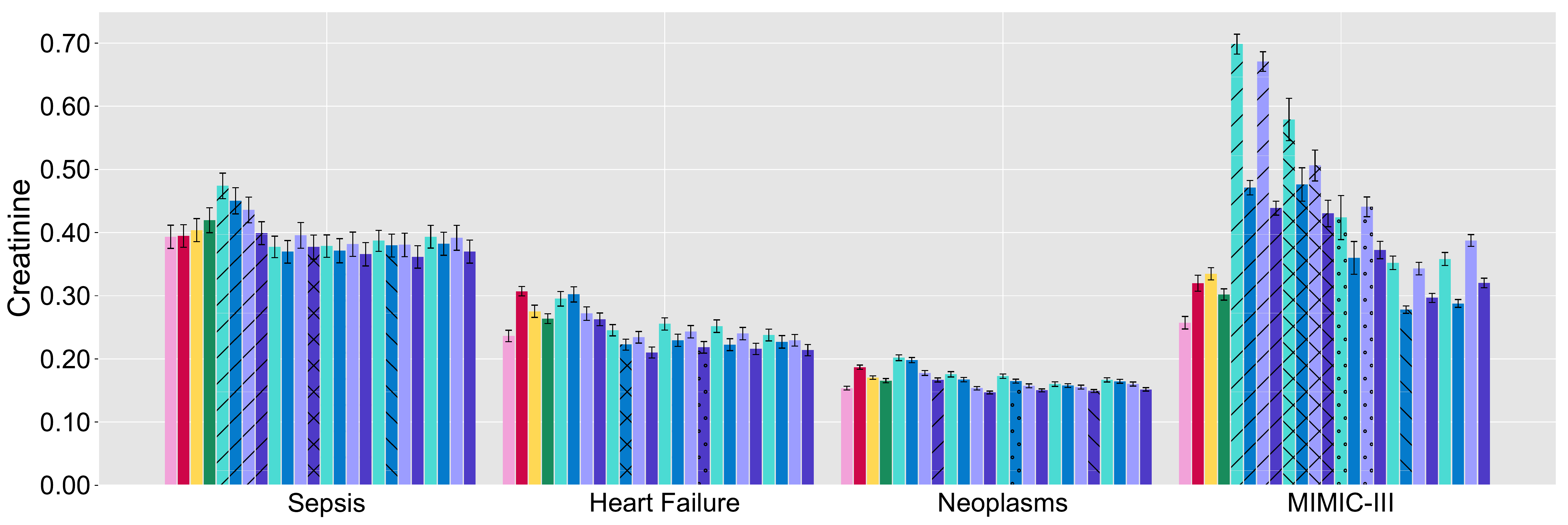}
}
\caption{\textbf{The mean absolute error (MAE) of online imputation under different $Q$ for all cohorts.} The error bars denote $\pm 1$ standard error.}
\label{fig:new_mae_p3}
\end{figure*}

\begin{figure*}
\centering
\subfigure{
\includegraphics[width=16.0cm]{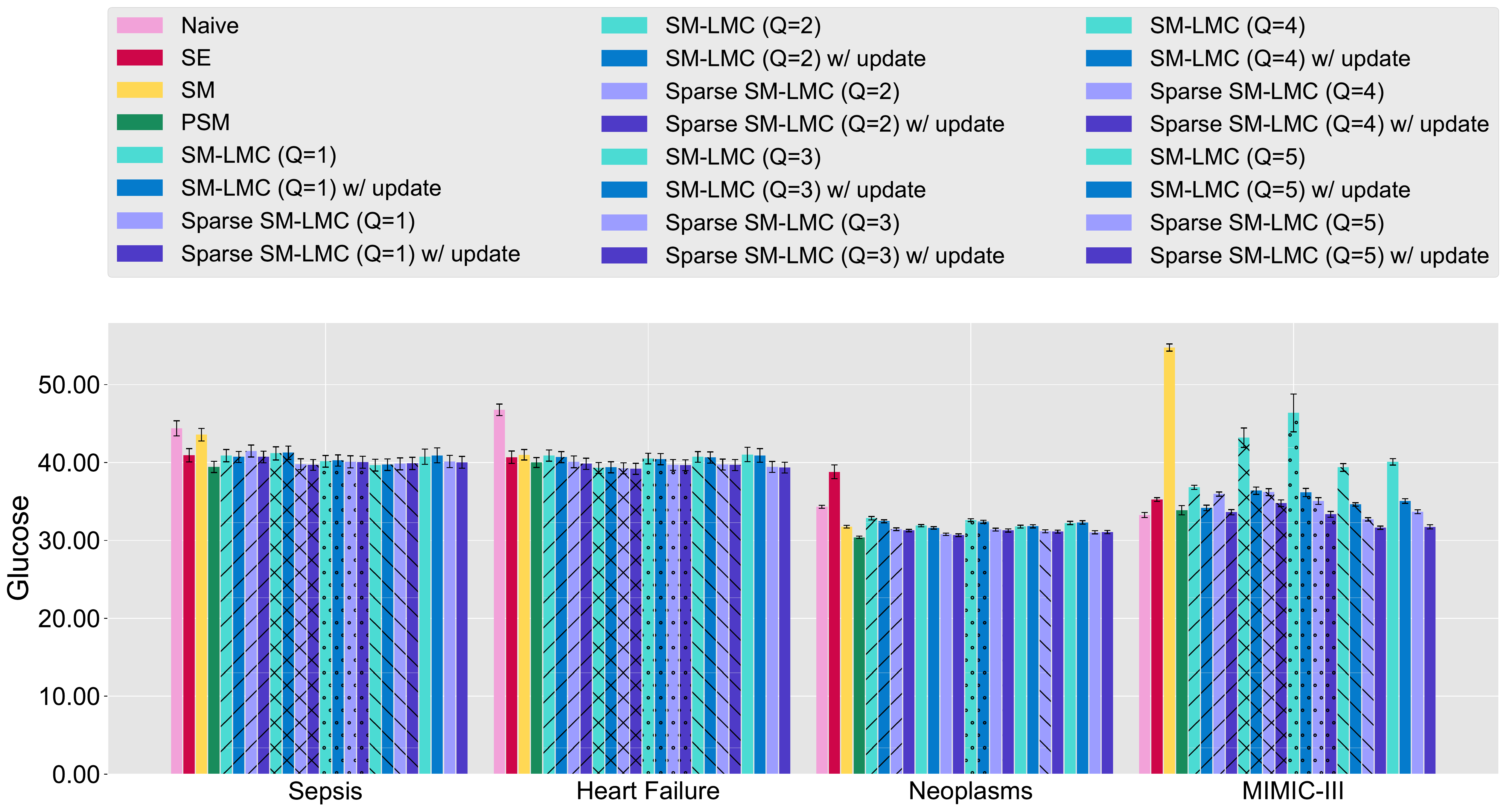}
}
\subfigure{
\includegraphics[width=16.0cm]{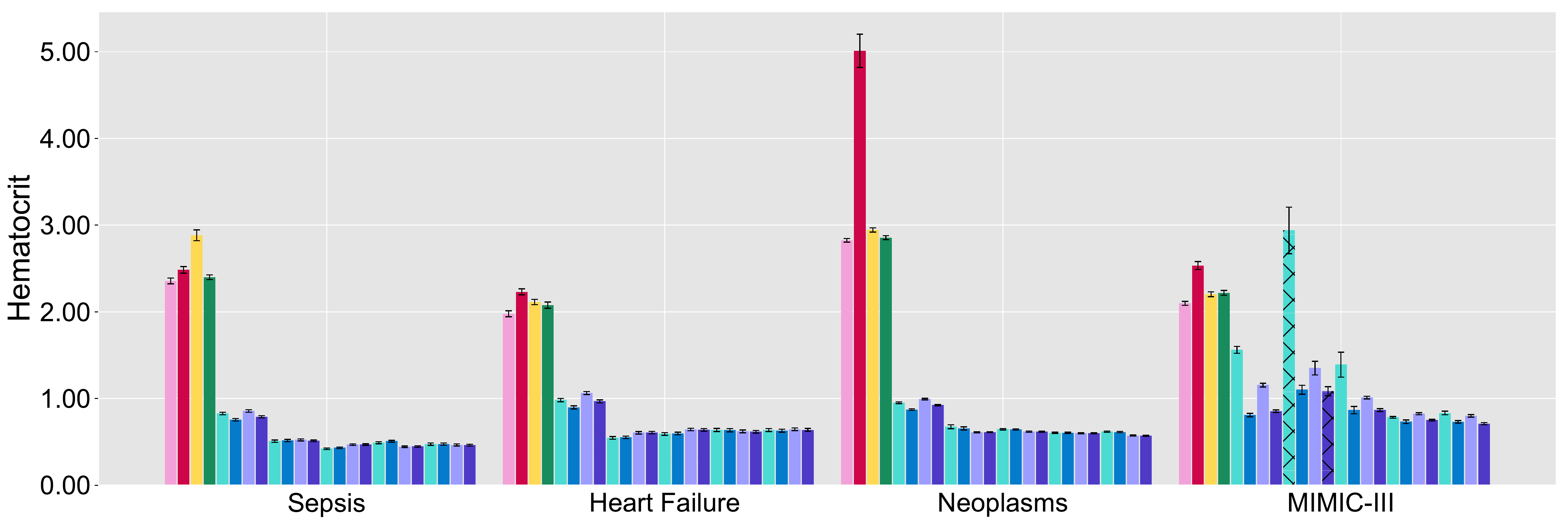}
}
\subfigure{
\includegraphics[width=16.0cm]{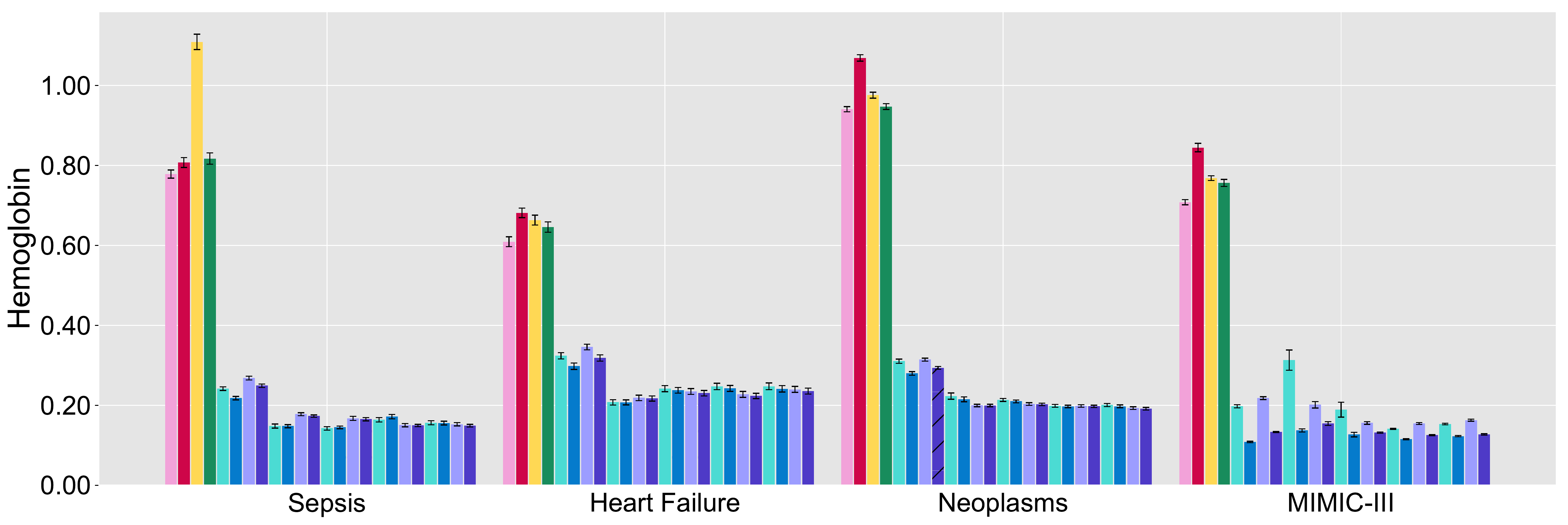}
}
\caption{\textbf{The mean absolute error (MAE) of online imputation under different $Q$ for all cohorts.} The error bars denote $\pm 1$ standard error.}
\label{fig:new_mae_p3}
\end{figure*}

%% file: AppendixE.tex
\newpage
\appendix
\section*{Appendix E. Improvements in Empirical Runtime}

\begin{figure*}[h]
 \centering
    \includegraphics[width=11cm]{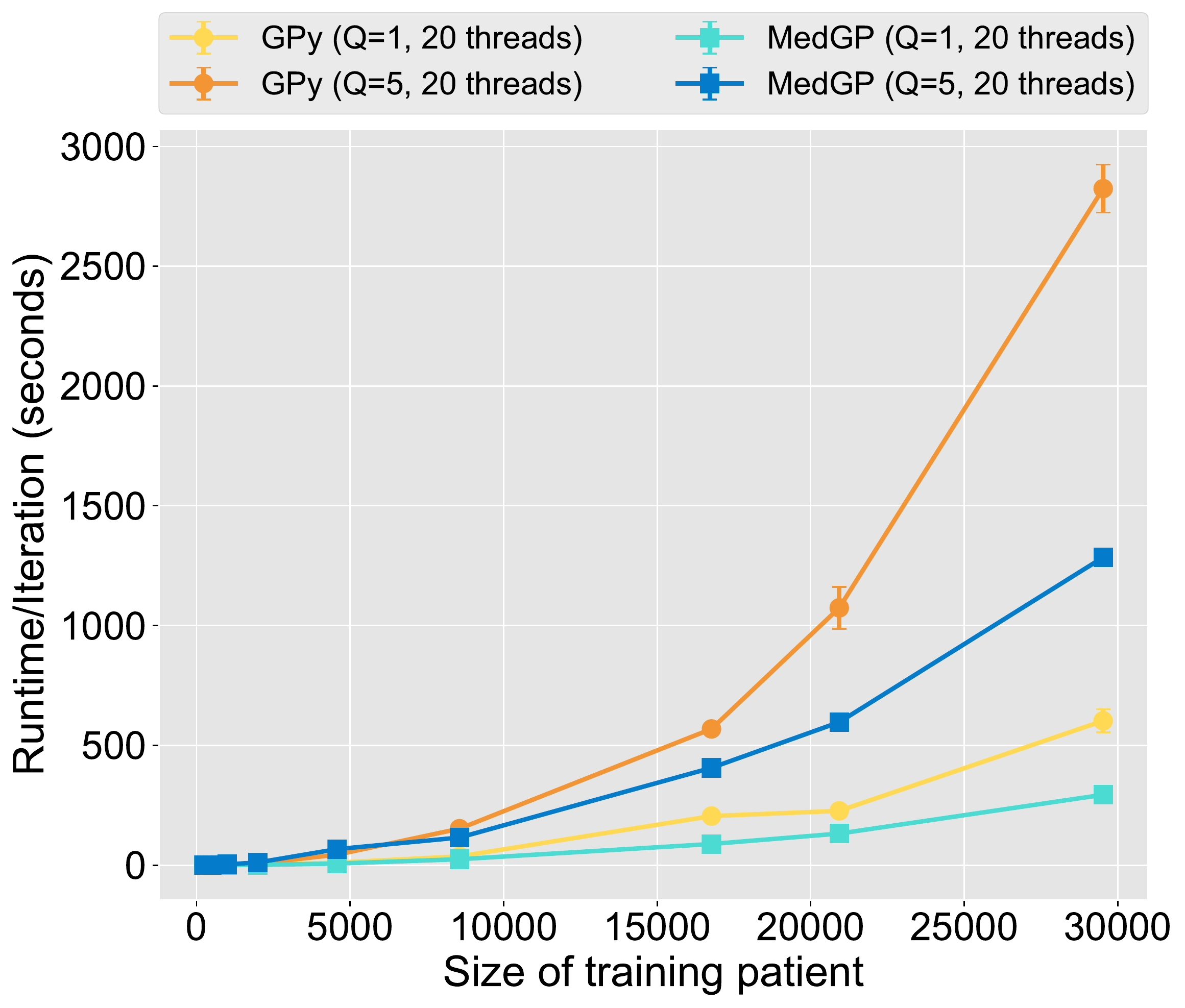}
    \caption{{\bf The empirical runtime of our implementation.} A comparison of the average runtimes for one iteration (including computation of gradients) for MedGP and optimized baseline GPy.}
    \label{fig:runtime}
\end{figure*}

 In this appendix, we provide the comparisons in runtime with GPy~\citep{gpy2014}, a state-of-the-art optimized Python library for GPs. We selected few benchmark cases from the MIMIC-III subset, and profiled the runtime for performing one iteration when using gradient-based optimizers. That is, the runtime for computing the gram matrix, log marginal likelihood, and gradients of all parameters. The experiments were performed on the machine with 20 Intel(R) Xeon(R) CPUs running at 2.50GHz (no GPUs were used). For GPy implementation, we also allowed multithreading and the access to MKL optimization for matrix operations, provided by Anaconda with academic license. In Figure~\ref{fig:runtime}, we show the average runtime for a single iteration under different number of basis kernels: $Q=1$ and $Q=5$, corresponding to 242 and 1114 parameters ($D=24, R=8$). We found that for training cases smaller than $10^4$ observations, GPy with multithreading is comparable to our implementation. However, for the cases larger than $10^4$ observations, our implementation speeds up by up to 2.5 times. The largest case we tested here includes 29,525 observations. 

